\documentclass{article} []
\pdfoutput=1
\usepackage[numbers,sort&compress]{natbib}
\setcitestyle{numbers,sort&compress,square,comma}
\usepackage{authblk}
\usepackage{amssymb}
\usepackage[preprint]{neurips_2025}
\pdfoutput=1
\usepackage[utf8]{inputenc} 
\usepackage[T1]{fontenc}    
\usepackage{url}            
\usepackage{booktabs}       
\usepackage{amsfonts}       
\usepackage{nicefrac}       
\usepackage{microtype}      
\usepackage{xcolor}         
\usepackage{float}          
\usepackage{graphicx}       
\usepackage{amsmath}
\usepackage{appendix}
\usepackage{booktabs}
\usepackage{multirow}
\usepackage{epsfig}
\usepackage{colortbl}
\usepackage{xcolor}
\usepackage{makecell}
\usepackage{hyperref} 

\usepackage{pifont}
\usepackage{graphicx}
\usepackage{wrapfig}
\usepackage{graphicx}
\usepackage{booktabs}
\usepackage{svg} 
\usepackage{colortbl}  
\usepackage{amsmath}
\usepackage{amssymb}
\usepackage{indentfirst}
\usepackage{threeparttable}
\usepackage{pifont}
\usepackage{makecell}
\usepackage{arydshln}
\usepackage{arydshln}
\usepackage{tabularx}
\newcommand{\blue}[1]{\textbf{\textcolor{mblue}{#1}}}
\newcommand{\red}[1]{\textbf{\textcolor{mred}{#1}}}
\usepackage{makecell}

\usepackage{pifont}
\usepackage{minitoc}
\definecolor{mblue}{RGB}{0, 77, 128}

\definecolor{mblue}{RGB}{0, 77, 128}
\definecolor{mred}{RGB}{192,0, 0}
\definecolor{darkgreen}{rgb}{0.0, 0.5, 0.0}
\definecolor{mycolor_blue}{HTML}{E7EFFA}
\definecolor{mycolor_green}{HTML}{E6F8E0}
\definecolor{mycolor_gray}{HTML}{ECECEC}
\newcommand{\cmark}{\textcolor{darkgreen}{\ding{51}}}  
\newcommand{\xmark}{\textcolor{red}{\ding{55}}}        

\hypersetup{
    colorlinks=true,        
    linkcolor=blue,        
    filecolor=magenta,     
    urlcolor=magenta,          
}
\title{LOVE: Benchmarking and Evaluating Text-to-Video Generation and Video-to-Text Interpretation}

\author{%
Jiarui Wang$^{1}$,    
Huiyu Duan$^{1,2}$, 
Ziheng Jia$^{1}$,
Yu Zhao$^{1}$,
Woo Yi Yang$^{1}$,
Zicheng Zhang$^{1}$,\\
Zijian Chen$^{1}$,
Juntong Wang$^{1}$, 
Yuke Xing$^{1}$,
Guangtao Zhai$^{1,2}$,
Xiongkuo Min$^{1}$\thanks{Corresponding author} \\
 $^{1}$Institute of Image Communication and Network Engineering, \\
$^{2}$ MoE Key Lab of Artificial Intelligence, AI Institute,\\ Shanghai Jiao Tong University, Shanghai, China\\
} 

\begin{document}
\doparttoc

\faketableofcontents
\maketitle

\begin{abstract}
Recent advancements in large multimodal models (LMMs) have driven substantial progress in both text-to-video (T2V) generation and video-to-text (V2T) interpretation tasks. 
However, current AI-generated videos (AIGVs) still exhibit limitations in terms of perceptual quality and text-video alignment. Therefore, a reliable and scalable automatic model for AIGV evaluation is desirable, which heavily relies on the scale and quality of human annotations. 
To this end, we present \textbf{AIGVE-60K}, a comprehensive dataset and benchmark for \underline{AI}-\underline{G}enerated \underline{V}ideo \underline{E}valuation,
which features \textbf{(i) comprehensive tasks}, encompassing 3,050 extensive prompts across 20 fine-grained task dimensions, \textbf{(ii) the largest human annotations}, including 120K mean-opinion scores (MOSs) and 60K question-answering (QA) pairs annotated on 58,500 videos generated from 30 T2V models, and \textbf{(iii)
bidirectional benchmarking and evaluating} for both T2V generation and V2T interpretation capabilities.
Based on AIGVE-60K, we propose \textbf{LOVE}, a \underline{L}MM-based metric f\underline{o}r AIG\underline{V} \underline{E}valuation from multiple dimensions including perceptual preference, text-video correspondence, and task-specific accuracy in terms of both instance level and model level.
Comprehensive experiments demonstrate that LOVE not only achieves state-of-the-art performance on the AIGVE-60K dataset, but also generalizes effectively to a wide range of other AIGV evaluation benchmarks. 
These findings highlight the significance of the AIGVE-60K dataset. Database and codes are available at \url{https://github.com/IntMeGroup/LOVE}.
\end{abstract}

\section{Introduction}
\label{sec:intro}
\vspace{-1mm}

The rapid advancement of large multimodal models (LMMs) has revolutionized the fields of both text-to-video (T2V) generation \cite{singh2023survey,xing2024survey,liao2024evaluation} and video-to-text (V2T) interpretation \cite{internlmxcomposer,ye2024mplug,liu2024llavanext}, leading to high-quality video generation and comprehensive multimodal video understanding capabilities.
However, state-of-the-art T2V models may still produce videos with degraded \textbf{perceptual quality} and limited \textbf{text-video correspondence}, thus may fail to meet human preferences \cite{wang2024aigv,li2024evaluating,huang2024vbench}.
Given the high cost and inefficiency of human evaluation, it is of great significance to develop a reliable and scalable evaluation metric that aligns well with human preferences for AI-generated videos (AIGVs) and corresponding T2V models. 

To fairly and effectively evaluate T2V models and AIGVs, many T2V model benchmarks and AIGV evaluation datasets~\cite{huang2024vbench,zheng2025vbench,liu2024evalcrafter,liu2024fetv,li2024evaluating,zhang2025q,chivileva2023measuring,zhang2024benchmarking,kou2024subjective,wang2024aigv,NEURIPS2024_46b5405a,yang2025lmme3dhfbenchmarkingevaluatingmultimodal} have been constructed as shown in Table \ref{tab:relate}, and many AIGV evaluation metrics have been proposed~\cite{VSFA,sun2022deep,wu2022fast,wu2023dover}. However, these efforts face the following limitations that may affect their effectiveness in diverse applications. (1) \textbf{Most benchmarks or datasets only consider either perception, correspondence, or task-specific accuracy dimensions, while comprehensive subjective evaluation works are still lacking.}
Since high-quality AIGVs may exhibit poor text-video alignment, well-aligned AIGVs may suffer from low perceptual quality \cite{wang2024aigv}, and we also need a binary true-or-false metric in some scenarios (such as number-based generation scenes) \cite{ghosh2023geneval,zheng2025vbench}, an extensive evaluation is important.
However, some existing metrics only focus on one dimension \cite{li2024evaluating} or use the fused overall evaluation \cite{liu2024fetv,kou2024subjective}.
(2) \textbf{The scale of the datasets remains small and annotations remain coarse.}
Some datasets or benchmarks include only a limited number of T2V models \cite{huang2024vbench,zheng2025vbench,liu2024fetv,li2024evaluating} or AIGVs \cite{liu2024evalcrafter,chivileva2023measuring,zhang2024benchmarking,kou2024subjective}, which constrains the ability to validate the effectiveness and scalability of evaluation methods across diverse models and outputs, and some works use coarse-MOSs \cite{liu2024evalcrafter,zhang2025q}, which do not meet ITU standards \cite{series2012methodology} and may produce invalid MOSs.
(3) \textbf{Current evaluation metrics for T2V generation consider from either the instance level or the model level, while a comprehensive study integrates both perspectives remains lacking.}
For example, Inception Score (IS) \cite{gulrajani2017improved}, Fréchet Video Distance (FVD) \cite{unterthiner2018towards}, VBench \cite{huang2024vbench}, \textit{etc.}, are mainly designed to evaluate T2V models, \textit{i.e.}, from a model level perspective, while LGVQ \cite{zhang2024benchmarking}, AIGV-Assessor \cite{wang2024aigv}, \textit{etc.}, are mainly designed for the individual AIGV evaluation or multiple AIGV comparison, \textit{i.e.}, from a instance level perspective. To the best of our knowledge, no existing work has jointly considered both perspectives towards a comprehensive evaluation.

\pdfoutput=1
\begin{figure}
    \centering
         \vspace{-6mm}
    \includegraphics[width=1\linewidth]{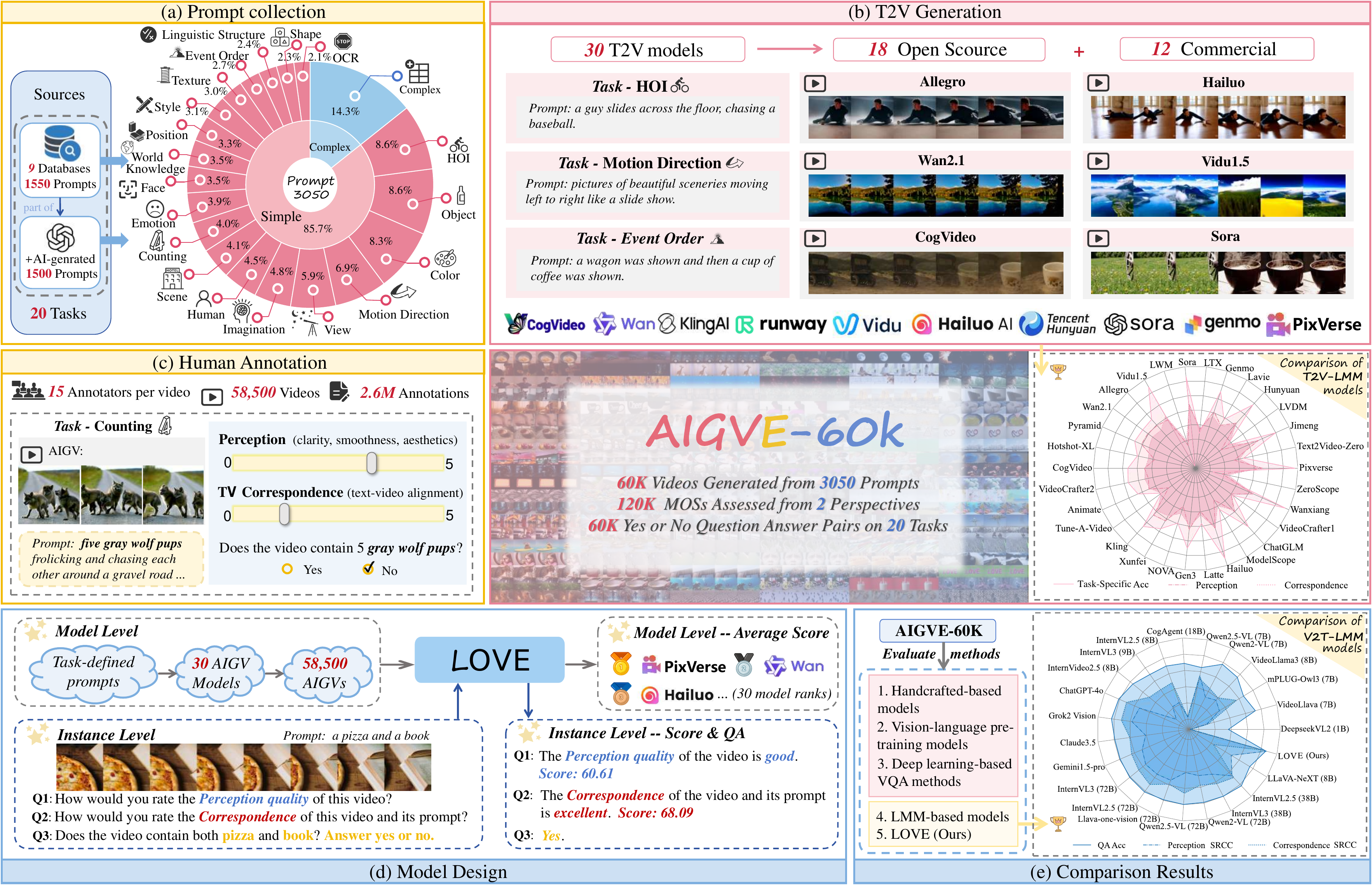}
     \vspace{-7mm}
    \caption{We present the largest AIGV evaluation database (\textbf{AIGVE-60K}) and a novel model (\textbf{LOVE}).
(a) We collect 3,050 prompts across 20 fine-grained tasks.
(b) 30 T2V models are applied to generate 60K videos.
(c) 120K MOSs from 2 perspectives and 60K question-answer pairs are acquired from annotators.
(d) We design a model to evaluate \textbf{T2V generation} at both the instance and model levels.
(e) Comparison results of LMM's \textbf{V2T interpretation} ability based on AIGVE-60K.} 
     \vspace{-5mm}
    \label{abstract}
\end{figure}
To address these challenges, we present \textbf{AIGVE-60K}, a large-scale dataset and benchmark for \underline{AI}-\underline{G}enerated \underline{V}ideo \underline{E}valuation, which includes 58,500 videos generated by 30 state-of-the-art T2V models using 3,050 diverse prompts across 20 task-specific challenges.
As shown in Figure \ref{abstract}, we collect \textbf{2.6M} human annotations from the perception, text-video correspondence, and task-specific accuracy, respectively, and finally obtain 120K mean opinion scores (MOSs) and 60K question-answering (QA) pairs.
Based on AIGVE-60K, we propose \textbf{LOVE}, a \underline{L}MM-based metric for \underline{AIGV} \underline{E}valuation from multiple dimensions at both instance level and model level, 
which integrates: (1) dual encoders for vision-temporal feature extraction, (2) a large-language model (LLM) backbone for \textbf{\textit{all-in-one}} video quality assessment, and (3) instruction tuning techniques \cite{liu2023visual} for accurate response generation. 
Through extensive experimental validation, we demonstrate that LOVE achieves state-of-the-art performance on the AIGVE-60K dataset and manifests strong zero-shot generalization ability on other benchmarks.
Our contributions are summarized as follows:

\begin{table*}[tbph]
\centering
\vspace{-9mm}
\caption{Comparison of T2V model evaluation benchmarks and AIGV quality evaluation databases. }
\renewcommand\arraystretch{1.2}
\label{tab:relate}
\resizebox{\textwidth}{!}{\begin{tabular}{c|c|c|c|c|c|cccc}
\hline
 \multirow{2}{*}{\textbf{Database} }& Annotation Type  & \multirow{2}{*}{\textbf{Videos}}  &\multirow{2}{*}{\textbf{ Prompts }} & \multirow{2}{*}{\textbf{Annotations }} & \multirow{2}{*}{\textbf{Models}} &   \multicolumn{4}{c}{\textbf{Evaluation Concern}}\\

 & (People per Sample) &&&&&T2V Tasks&  Perception & T2V Correspondence & QA Acc\\
  \hline
VBench~\cite{huang2024vbench} &Pairs (1)  & 3,200 & 800 &24,000 & 4 &16 & \cmark & \cmark& \xmark\\
VBench2.0~\cite{zheng2025vbench} &Pairs (1) & 100,800& 1,260 & 151,200 &4& 18 & \xmark & \cmark & \cmark \\
\hdashline

EvalCrafter~\cite{liu2024evalcrafter}&Coarse-MOS (7)&2,500& 700 & 70,000 &8& 17 & \cmark & \cmark & \xmark\\
FETV~\cite{liu2024fetv}&Coarse-MOS (3) &2,476& 619 & 7,428 &4&5 & \multicolumn{2}{c}{\textit{Overall} }  & \xmark\\
GenAIBench~\cite{li2024evaluating}&Coarse-MOS (3) & 32,000& 800&9,600&4&8&\xmark & \cmark &\xmark\\
Q-Eval~\cite{zhang2025q} &Coarse-MOS (3) &40,000 & 2,500 & 384,000 &16& \xmark& \cmark& \cmark& \cmark\\
%
\hdashline
MQT~\cite{chivileva2023measuring} &Fine-MOS (24)&1,005& 201 & 48,240&5&\xmark&\cmark & \cmark& \cmark\\
LGVQ~\cite{zhang2024benchmarking}      & Fine-MOS (20)   & 2,808   & 468  &  168,480    & 6  &\xmark  & \cmark    & \cmark     &\xmark\\

T2VQA-DB~\cite{kou2024subjective}         & Fine-MOS (27)   & 10,000   & 1,000  & 270,000      & 9  & \xmark &  \multicolumn{2}{c}{\textit{Overall} }   &\xmark     \\

AIGVQA-DB~\cite{wang2024aigv} & Fine-MOS (20) \& Pairs (3) & 36,576  & 2,048 &  371,520 & 15 & \xmark & \cmark & \cmark & \cmark\\
 \rowcolor{gray!20} \textbf{AIGVE-60K (Ours)}           & \textbf{Fine-MOS (15) }   & \textbf{58,500} & \textbf{3,050}  & \textbf{2,632,500}  & \textbf{30}     & \textbf{20}  &  \textbf{\cmark} &\textbf{\cmark} &\textbf{\cmark}    \\
\hline
\end{tabular}}
\vspace{-4mm}
\end{table*}
\begin{itemize}
    \item We present \textbf{AIGVE-60K}, \textbf{the largest text-to-video evaluation dataset} so far that contains 58,500 generated videos with 2.6M subjective ratings from the perception, text-video correspondence, and task-specific accuracy, respectively.

    \item We introduce \textbf{a bidirectional benchmarking and evaluation strategy}. Based on AIGVE-60K, we can benchmark the \textbf{T2V \textit{generation ability}} of 30 T2V models, and the \textbf{V2T \textit{interpretation ability}} of 23 LMMs and 24 VQA metrics, respectively.

    \item We propose \textbf{LOVE}, \textbf{a novel LMM-based evaluation model} capable of assessing both the perceptual quality and T2V alignment for AIGVs.
     Extensive experimental results on AIGVE-60K and other AIGV benchmarks manifest the state-of-the-art performance and strong generalization ability of LOVE.
\end{itemize}

\section{Related Works}\label{sec:related}

\subsection{Benchmarks for T2V Generation}
\vspace{-2mm}
As shown in Table \ref{tab:relate}, the development of T2V generation has spawned many T2V model evaluation benchmarks and VQA databases, which can be categorized into pairs, coarse MOS, and fine-MOS based on the annotation method and granularity.
VBench~\cite{huang2024vbench} and VBench2.0~\cite{zheng2025vbench} focus on video pairs comparison, but are limited in T2V comparison model number and lack precise quality assessment for each AIGV.
Fine-MOS databases offer more reliable assessments derived from more than 15 annotators, following the guidelines of ITU-R BT.500 \cite{series2012methodology}. 
MQT~\cite{chivileva2023measuring}, LGVQ~\cite{zhang2024benchmarking} collect fine-grained MOSs but the number of AIGVs is limited. AIGVQA-DB~\cite{wang2024aigv} considers both perceptual quality and T2V correspondence, however, it mainly focuses on the pair comparison and lacks task-specific QA pairs, limiting their ability to assess T2V generation across diverse tasks. AIGVE-60K stands out by its largest scale of annotations, providing fine-grained MOSs for both perceptual quality and T2V correspondence, and answer annotations for  task-specific questions.
\subsection{Evaluation Metrics for T2V Generation}
\vspace{-2mm}
Many quality assessment models have been proposed in the literature \cite{mittal2012making,10647885,duan2024finevq,wang2025qualityassessmentaigenerated,xu2025harmonyiqapioneeringbenchmarkmodel,li2022blindly,kirstain2023pick,li2024evaluating}, including handcrafted models (\textit{e.g.}, NIQE~\cite{mittal2012making}, QAC~\cite{xue2013learning}, BRISQUE~\cite{mittal2012no}) and deep learning-based VQA models (\textit{e.g.}, BVQA \cite{li2022blindly}, FAST-VQA \cite{wu2022fast}, DOVER \cite{wu2023dover}). These models characterize
quality-aware information to predict perception quality scores but can not evaluate T2V correspondence, which is crucial for assessing the relationship between the generated video and its corresponding text prompt.
 PickScore~\cite{kirstain2023pick} and VQAScore~\cite{li2024evaluating} improve the evaluation of the T2V correspondence, but they struggle to assess the perception quality of AIGV.
LMMs with visual understanding capabilities perform well in QA tasks, but their ability to assess image perceptual quality remains limited and often fail to give precise quality scores. 
VBench~\cite{huang2024vbench} employs various detection models for task-specific accuracy, but this approach is quite complex. 
To address this gap, our proposed LMM-based model complies with an \textit{\textbf{all-in-one}} framework, which can evaluate quality scores and task-specific accuracies in one model.

\section{AIGVE-60K Dataset \& Benchmark}
\vspace{-2mm}
In this section, we introduce the construction process of AIGVE-60K and \textbf{benchmark T2V models} based on the dataset.
\subsection{Data Collection}

\vspace{-2mm}
Prompts of the AIGVE-60K are primarily sourced from 9 existing open-domain text-video pair datasets and some are refined using DeepSeek R1~\cite{guo2025deepseek} to expand and modify them, ensuring clarity and diversity.
Our prompt design focuses on 20 different tasks as shown in Figure \ref{abstract}(a). 
The complex tasks are designed by combining simpler task components, such as motion direction, event order, and counting, into more complex challenges. 
In total, we collect 3,050 prompts, each corresponding to a specific task. 
To generate the AIGVs, we utilize 30 of the latest T2V models, as shown in Figure \ref{abstract}(b).
We leverage open-source website APIs or the default weights of these models to generate videos. 
For the training set, we employ 2,750 distinct prompts, each processed by 18 open-source models. The test set consists of 300 unique prompts generated using all 30 models.  With 3,050 distinct prompts, this process results in a total of 58,500 videos (2,750 prompts × 18 open-source models + 300 prompts × 30 open-source and close-source models). The imbalance in the number of T2V models used between the training and testing sets is attributed to two factors: (1) generating videos using close-source tools is costly, and (2) we aim to evaluate the scalability of evaluation metrics on training-set unseen generation models.
More details in \textit{Appendix} Sections \ref{tasks} and \ref{modelss}.
\pdfoutput=1
\begin{figure}[!t]
    \centering
         \vspace{-9mm}
    \includegraphics[width=1\linewidth]{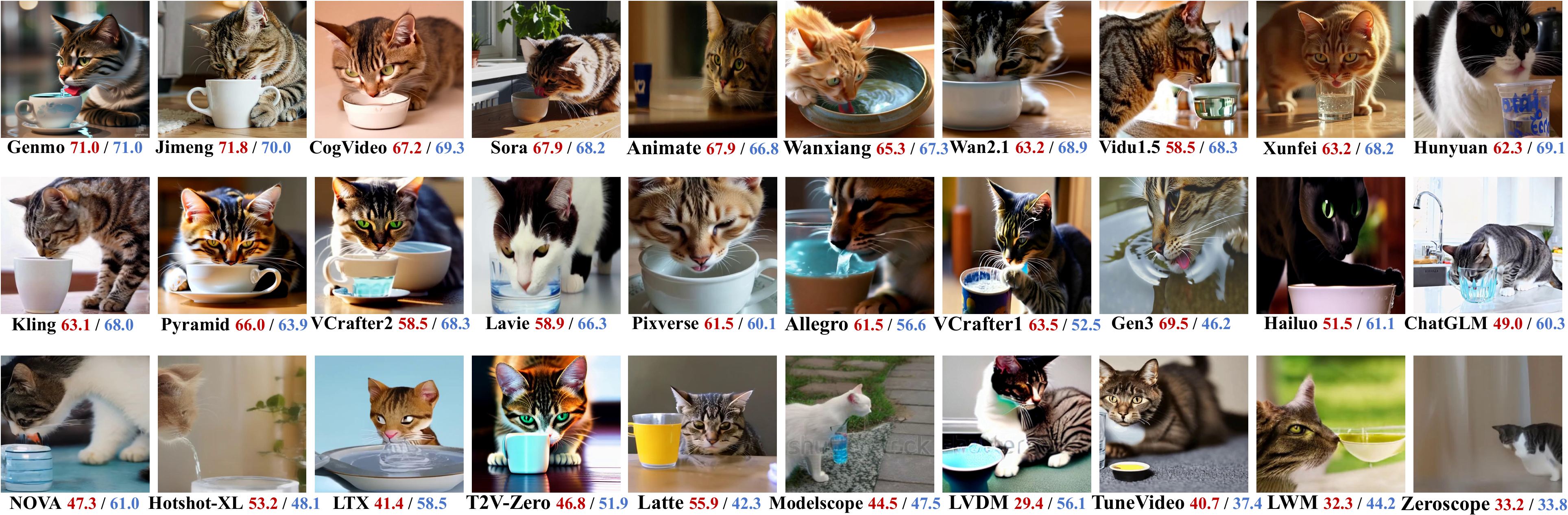}
     \vspace{-6mm}
    \caption{Video examples generated by 30 T2V models using prompt: “\textit{a cat is drinking a cup of water}”, annotated with MOSs from 2 dimensions: \red{perceptual quality} and \blue{text-video correpondence}.} 
     \vspace{-3mm}
    \label{myexample}
\end{figure}

\subsection{Subjective Experiment Setup and Procedure}
\vspace{-2mm}
Due to the unique distortions in AIGVs and varying elements determined by different text prompts, relying solely on an overall score for evaluation is inadequate. In this paper, we propose to evaluate AIGVs across two dimensions, as shown in Figure \ref{myexample}.
(1) \textbf{Perceptual quality} focuses on visual perception, evaluating factors such as detail richness, motion smoothness, color vibrancy, and distortion levels.
(2) \textbf{Text-video correspondence} evaluates how accurately the generated video reflects the objects, scenes, styles, and details described in the text prompt.
We use a 1-5 Likert scale to score the videos based on the perception and T2V correspondence. For the correspondence evaluation, in addition to the rating, annotators are instructed to answer task-specific yes/no questions to determine whether the video consistently aligns with the prompt.
 Finally, we obtain a total of 2,632,500 human annotations including 
 1,755,000 reliable score ratings (15 annotators $\times$ 2 dimensions $\times$ 58,500 videos), and 877,500 task-specific QA pairs (15 annotators $\times$ 58,500 videos).
 
\vspace{-1mm}
\subsection{Subjective Data Processing}
\vspace{-2mm}
In order to obtain the MOS for an AIGV, we first convert the raw ratings into Z-scores, and then 
linearly scale them to the range $[0,100]$ as follows:
\begin{equation}
z_i{}_j=\frac{r_i{}_j-\mu_i{}}{\sigma_i},\quad z_{ij}'=\frac{100(z_{ij}+3)}{6},~~\mu_i=\frac{1}{N_i}\sum_{j=1}^{N_i}r_i{}_j, ~~ \sigma_i=\sqrt{\frac{1}{N_i-1}\sum_{j=1}^{N_i}{(r_i{}_j-\mu_i{}_j)^2}},
\end{equation}
where $r_{ij}$ is the raw rating given by the $i$-th subject to the $j$-th video. $N_i$ is the number of videos judged by subject $i$. 
Next, the MOS of the $j$-th video is computed by averaging the rescaled z-scores across all subjects as follows:
\begin{equation}
\text{MOS}_j=\frac{1}{M}\sum_{i=1}^{M}z_{ij}',
\end{equation}
where $\text{MOS}_j$ indicates the MOS for the $j$-th AIGV, $M$ is the number of subjects, and $z'_i{}_j$ are the rescaled z-scores. 
The task-specific yes/no answer is determined by the most votes.
Therefore, a total of 117,000 MOSs (2 dimensions $\times$ 58,500 videos) and 58,500 question answering pairs are obtained.


\subsection{Subjective Data Analysis \& T2V Model Benchmark}
\begin{figure}
\vspace{-10mm}
    \centering
    \includegraphics[width=1\linewidth]{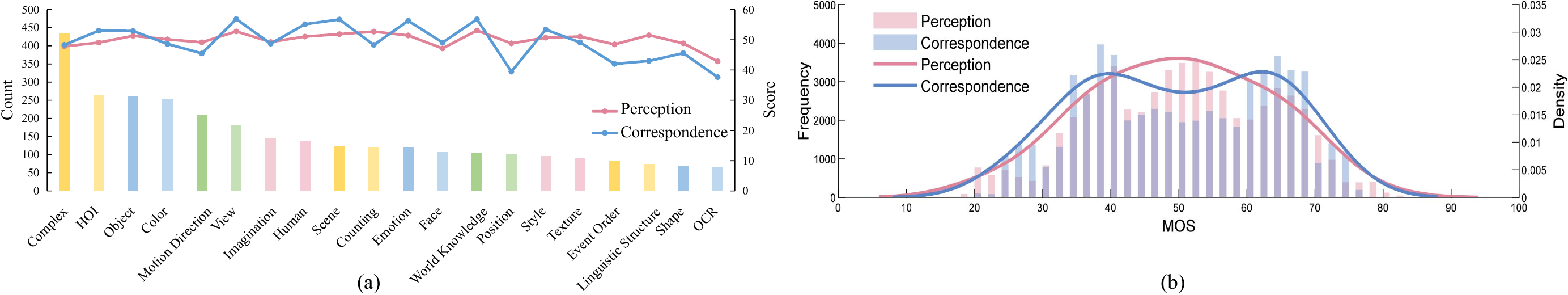}
     \vspace{-7mm}
    \caption{(a) Distribution of task counts and averaged scores across different tasks. (b) Distribution of perception and correspondence MOSs. } 
     \vspace{-2mm}
    \label{distribution}
\end{figure}

\begin{figure}[!t]
    \centering
         \vspace{-1mm}
    \includegraphics[width=1\linewidth]{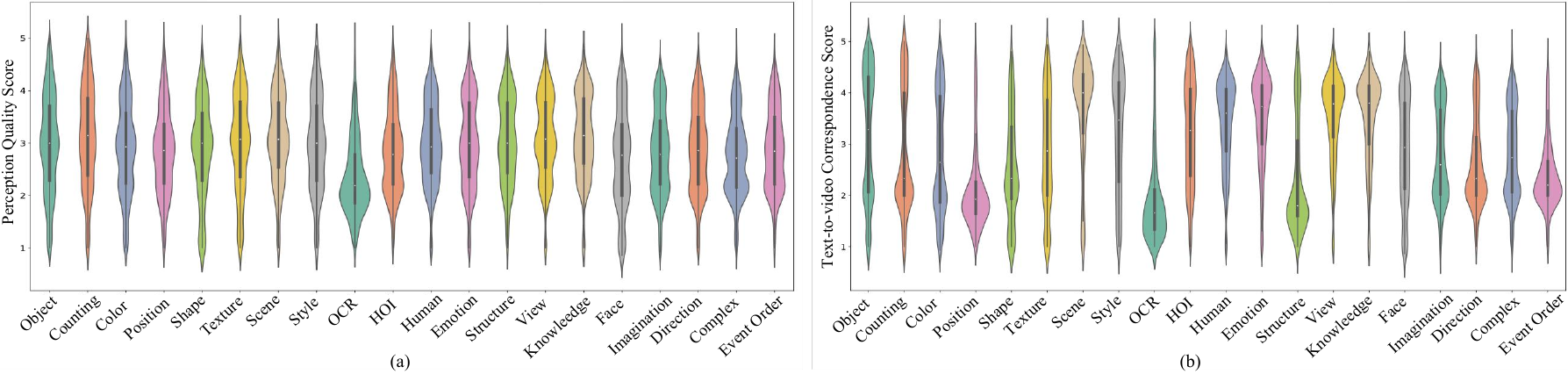}
     \vspace{-7mm}
    \caption{The MOS distribution in terms of different task contents. (a) MOS distribution of perception quality (b) MOS distribution of T2V correspondence. } 
     \vspace{-2mm}
    \label{xtqt}
\end{figure}

\begin{figure}[!t]
    \centering
         \vspace{-1mm}
    \includegraphics[width=1\linewidth]{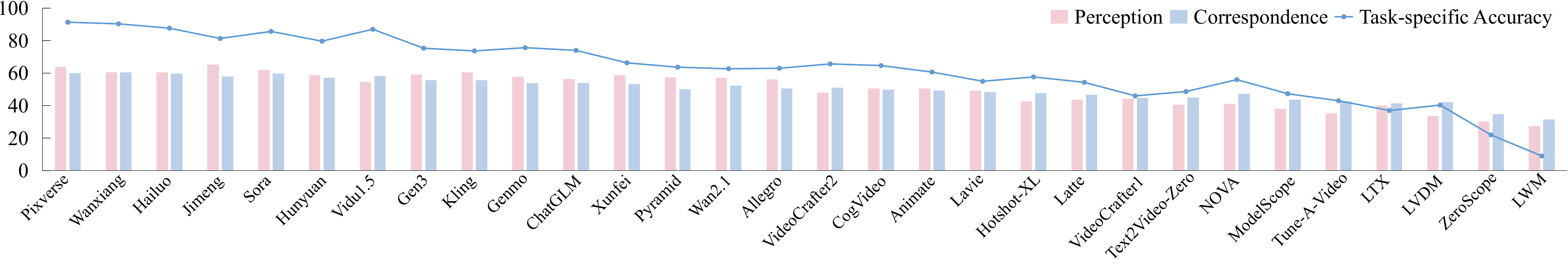}
     \vspace{-7mm}
    \caption{Comparison of T2V generation models regarding the perception MOSs, correspondence MOSs, and task-specific accuracy.} 
     \vspace{-2mm}
    \label{MOS}
\end{figure}

\begin{figure}[!t]
    \centering
     \vspace{-1mm}
    \includegraphics[width=1\linewidth]{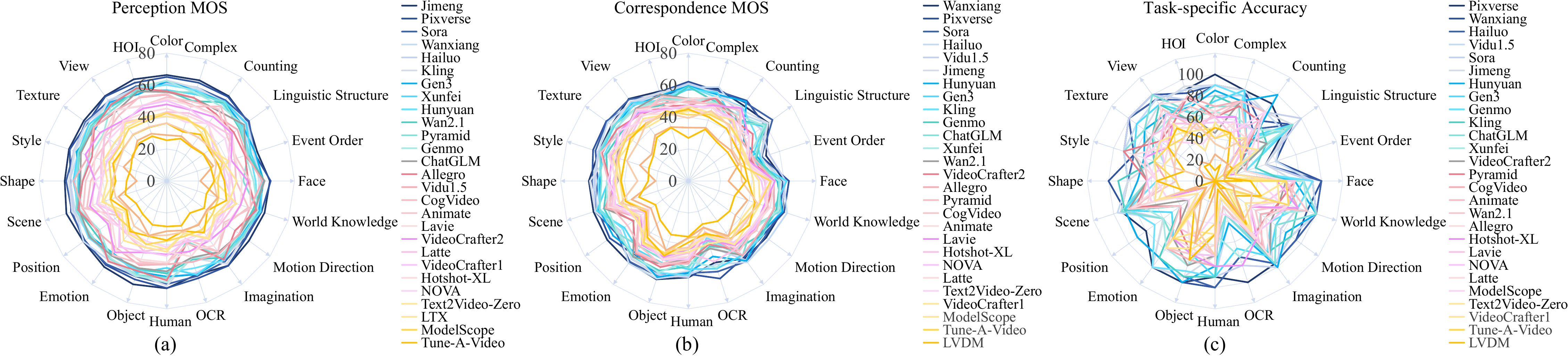}
     \vspace{-6mm}
    \caption{Comparison of MOSs and task-specific accuracy of 30 T2V generation models across 20 tasks with descending order arranged in legend. (a) Results in terms of perception MOSs. (b) Results in terms of correspondence MOSs. (c) Results in terms of task-specific accuracy. } 
     \vspace{-3mm}
    \label{leidat}
\end{figure}

\vspace{-2mm}
The distribution of task counts and averaged scores is shown in Figure \ref{distribution}(a). It can be observed that the correspondence score fluctuates more than the quality score, which means that the text-video alignment is more sensitive to the tasks.
The distribution of MOSs for both T2V correspondence and perceptual quality is shown in Figure \ref{distribution}(b), which approximately follows the Gaussian distributions.  Figure \ref{xtqt} displays the MOS distribution for each task, demonstrating notable differences in model capabilities between tasks, especially for T2V correspondence.
Moreover, we also launch comparisons for T2V generation models across different tasks based on perceptual quality MOSs, T2V correspondence MOSs, and task-specific accuracy, as shown in Figure \ref{MOS}. Jimeng~\cite{jimeng} achieves a higher perceptual quality than correspondence, while Vidu1.5~\cite{vidu_ai} shows the inverse trend. 
We further analyze the MOSs and task-specific accuracies across 20 different tasks. 
As shown in Figure
\ref{leidat}(a), perception MOS shows particular sensitivity to OCR tasks, where video clarity directly impacts character recognition accuracy. Figure \ref{leidat}(b) and (c) display similar trends, with task-specific accuracy results exhibiting sharper distinctions, which manifests the higher discriminative of the task-specific accuracy perspective.
While task-specific accuracy delivers binary (0/1) evaluations, MOS offers continuous scoring, enabling more granular evaluation of T2V correspondence.
Moreover, for the task of event order, most models show poor performance, indicating fundamental limitations in current models' ability to generate temporally coherent narratives.
Finally, the top-performing models vary across perceptual quality, text-video correspondence, and task-specific accuracy, underscoring the importance of evaluating AIGVs from these three perspectives.

\pdfoutput=1
\pdfoutput=1
\pdfoutput=1
\pdfoutput=1

\section{The LOVE Method}
\vspace{
-2mm
}
In this section, we present an \textit{\textbf{all-in-one}} video quality assessment method \textbf{LOVE} to identify quality degradation levels, predict perception and T2V correspondence scores, and deliver visual question answers within a unified model.
\pdfoutput=1
\begin{figure}
    \centering
    \vspace{-10mm}
    \includegraphics[width=1\linewidth]{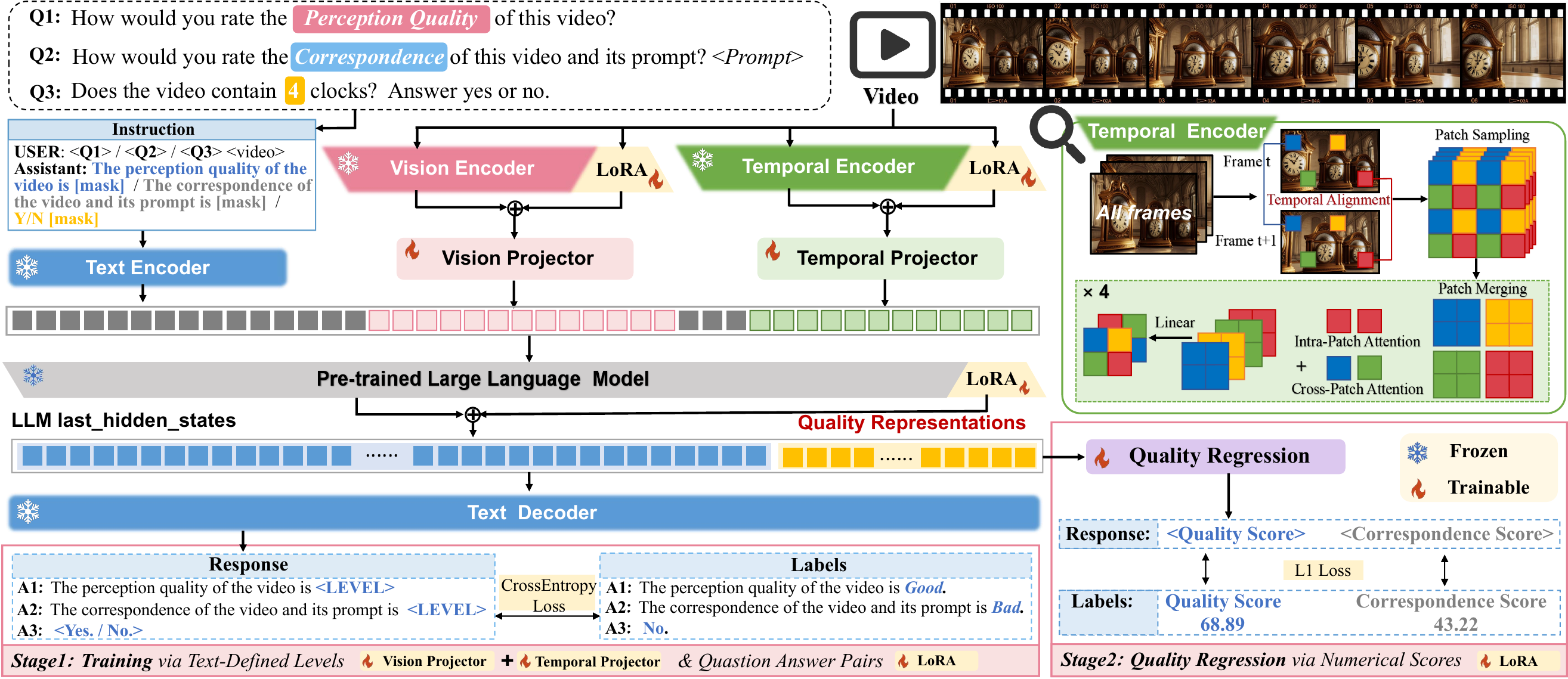}
    \vspace{-6mm}
    \caption{Overview of the LOVE architecture. The model includes two functions: (1) text-defined quality level and score prediction, (2) task-specific visual question answering. The training process consists of two stages: training via text-defined levels, and quality regression via numerical scores. The model incorporates a vision encoder, a temporal encoder, and a text encoder for extracting visual and
textual features, which are fed into a pre-trained LLM to generate results. LoRA  \cite{hulora} weights are introduced to the pre-trained image encoder and the LLM to adapt the models to perception quality and T2V correspondence evaluation, and 20 task-specific visual question-answering tasks.} 
     \vspace{-3mm}
    \label{modell}
\end{figure}
\subsection{Model Structure}
\vspace{-1mm}
\paragraph{Visual and Temporal Encoding.}
Figure \ref{modell} illustrates that the visual encoding component comprises a vision encoder and a temporal encoder for feature extraction, along with two projectors for aligning the video features with the input of the large language model (LLM).
 The vision encoder is constructed upon a pre-trained vision transformer (ViT), \textit{i.e.}, InternViT \cite{chen2024expanding}.
 To improve the scalability of processing high-resolution videos, we use a pixel unshuffle method that decreases the number of visual tokens to a quarter of their initial size.
For the temporal encoder,  we decompose all video frames $\{ F^{(n)}\}_{n=1}^{N_v}$ into temporally aligned mini-patch map through grid mini-patch sampling, following \cite{wu2022fast}. 
For each video frame $F^{(n)}$, we first split it into uniform $L\times L$ grids, the set of grids $G^{(n)}$ can be described as:
\begin{equation}
    G^{(n)}= \{ g_{0,0}^{(n)},\cdots, g_{i,j}^{(n)},\cdots,g_{L,L}^{(n)} \}, \quad g_{i,j}^{(n)} = F^{(n)} [ \frac{i\times H}{L} : \frac{(i+1)\times H}{L}, \frac{j\times W}{L}:\frac{(j+1)\times W}{L} ]
\end{equation}
where $g_{i,j}^{(n)} \in \mathbb R^{\frac{H}{L} \times \frac{W}{L} \times 3}$ denotes the grid in the $i$-th row and $j$-th column of $F^{(n)}$. Then we sample the mini-patches from each $g_{i,j}^{(n)}$  and splice all the selected mini-patches to get the mini-patch map $M \in \mathbb R^{H \times W \times 3}$,
which are then fed into a Swin-T~\cite{liu2021swin} with four hierarchical self-attention layers as the backbone.
To align the extracted features with the input space of the LLM, a vision projector and a temporal projector with two multilayer perceptron (MLP) layers are applied.
\vspace{-2mm}
\paragraph{Feature Fusion and Quality Regression.}
We utilize the InternLM3-9B-instruct~\cite{wang2024mpo} to integrate the visual tokens and text instruction tokens to perform the following two tasks.
(1) Prediction of text-defined quality level: the model produces an evaluation of the input video's quality level, such as “\textit{The perception quality of the video is (bad, poor, fair, good, excellent)}.”
A preliminary sense of the video quality is provided by text-defined categorization, which is useful for directing later quality regression tasks because LLMs comprehend textual data better than numerical data. 
(2) Regression score output: the model takes the quality representations from the last hidden states of the LLM to perform regression through a quality regression module, outputting numerical quality scores.
\vspace{-1mm}
\subsection{Training and Fine-tuning Strategy}
\vspace{-2mm}
LOVE is trained using a two-stage methodology. To enable precise score prediction, we first perform instruction tuning via text-defined quality levels and then train the quality regression module via numerical scores with LoRA \cite{hulora}.
Since LMMs have better understanding of textual data than numerical data, we first convert the continuous scores into categorical text-based quality levels. 
Specifically, we uniformly divide the range between the highest score ($\mathrm{M}$) and the lowest score ($\mathrm{m}$) into five distinct intervals, assigning the scores in each interval to respective levels:

\begin{equation}
    {L(s)} = l_V \text{  if } \text{m} + \frac{i-1}{5}  \times \mathrm{(M-m)} < s \leq \mathrm{m} + \frac{i}{5} \times \mathrm{(M-m)},
\end{equation}
where \{$l_V|_{i=1}^{5}\}=\{\textit{bad, poor, fair, good, excellent}\}$ are the standard text rating levels as defined by ITU~\cite{series2012methodology}. This step provides the LMM with a preliminary concept of video quality in terms of text-defined quality levels. We then take the last-hidden-state features from the LMM to a quality regression module to generate more accurate quality scores.
\begin{table}
\setlength{\belowcaptionskip}{-0.01cm}
\centering
\belowrulesep=0pt
\aboverulesep=0pt
\renewcommand\arraystretch{1.1}
\vspace{-6mm}
\caption{{\bf Performance benchmark on AIGVE-60K.} {\it Overall-Averaged} indicates that we average the MOS of the two dimensions as the overall video quality score. $\spadesuit$Conventional handcrafted metrics, $\diamondsuit$VBench metrics, $\clubsuit$deep learning-based VQA models, $\heartsuit$vision-language pre-training models, $\bigstar$open-source LMM-based models, and $\triangle$close-source LMM-based models. $\blacklozenge$*Refers to fine-tuned models. The best results are marked in {\red{RED}} and the second-best in {\blue{BLUE}}. 
}\vspace{1mm}
   \resizebox{\linewidth}{!}{\begin{tabular}{l|cc:cc|cc:cc|cc:cc|c:c}
    \toprule[1pt]
   {\bf Dimension}  & \multicolumn{4}{c|}{\textbf{Perception Quality}}&\multicolumn{4}{c|}{\textbf{T2V Correspondence}}&\multicolumn{4}{c|}{\it \textbf{Overall-Averaged}}&\multicolumn{2}{c}{\textbf{Question Answer}}\\
   \cdashline{2-15}
   \textit{Evaluation Level} & \multicolumn{2}{c:}{\textit{Instance-level}} & \multicolumn{2}{c|}{\textit{Model-level}}& \multicolumn{2}{c:}{\textit{Instance-level}} & \multicolumn{2}{c|}{\textit{Model-level}}& \multicolumn{2}{c:}{\textit{Instance-level}} & \multicolumn{2}{c|}{\textit{Model-level}}&\textit{Instance}&\textit{Model}\\
   \cdashline{2-15}
{\bf Methods / Metrics}&SRCC$\uparrow$&PLCC$\uparrow$&SRCC$\uparrow$&PLCC$\uparrow$&SRCC$\uparrow$&PLCC$\uparrow$&SRCC$\uparrow$&PLCC$\uparrow$&SRCC$\uparrow$&PLCC$\uparrow$&SRCC$\uparrow$&PLCC$\uparrow$&Acc (\%)$\uparrow$&SRCC$\uparrow$\\
    \midrule
    $\spadesuit$BMPRI \cite{quality:BMPRI} &0.5741&0.4976&0.7878&0.7312& 0.3618 &0.3214 &0.7321 &0.6582&0.5106 &0.4452 &0.7709&0.7112 &64.00 &0.7023\\
    $\spadesuit$BPRI  \cite{min2017blind}&0.3558 &0.3403 &0.6356 &0.7122 &0.2018& 0.2012 &0.5324 & 0.6256&0.3044&0.2945&0.5800&0.6863&63.56&0.4701\\
    $\spadesuit$BRISQUE \cite{mittal2012no} &0.5843&0.4881&0.8131 &0.7196 &0.3806 &0.3325 & 0.7615 & 0.6598&0.5239&0.4459&0.8011&0.7048&64.67&0.7366\\
   $\spadesuit$HOSA \cite{xu2016blind} &0.6474&0.5642&0.8456&0.7397 &0.4153 & 0.3682 & 0.7780 & 0.6735&0.5773&0.5068&0.8269&0.7226&64.34&0.7571\\
   $\spadesuit$NIQE \cite{mittal2012making}&0.6536&0.3517&0.8412&0.7500& 0.4345 &0.2544 & 0.7838 & 0.7037 &0.5917&0.3294&0.8251&0.7412&62.21&0.7602\\
   $\spadesuit$QAC \cite{xue2013learning}&0.5958&0.5717&0.8100&0.7447& 0.3948 & 0.3430 & 0.7717 & 0.6557&0.5421&0.4974&0.8029&0.7182&64.40&0.7495\\
    \hdashline
    $\diamondsuit$V-Aesthetic Quality~\cite{huang2024vbench}&0.5031 &0.5315 &0.7740 &0.8309 &0.4033 &0.4358 &0.7273 &0.7986 &0.4877 &0.5254 &0.7526& 0.8291&64.54&0.7046\\
     $\diamondsuit$V-Imaging Quality~\cite{huang2024vbench}&0.2810 &0.3093 &0.5426 &0.4766 &0.1952 &0.2244 &0.4986 &0.5101 &0.2531 &0.2900 &0.5186 &0.4970&60.60&0.4389\\
    $\diamondsuit$V-Overall Consistency~\cite{huang2024vbench}&0.1559 &0.1939 &0.1742 &0.3319 & 0.3076 &0.3685 &0.3201 &0.5419 &0.2510 &0.3046 &0.2338 &0.4233 &61.96&0.3353\\
    $\diamondsuit$V-Subject Consistency~\cite{huang2024vbench}& 0.3443 & 0.3968 &0.4839 &0.6065 &0.1647 &0.2425 &0.4416 &0.5744 &0.2739 &0.3476 &0.4679 &0.6016&62.52&0.4545\\
   $\diamondsuit$V-Temporal Flickering~\cite{huang2024vbench}& 0.4076 &0.4491 &0.6396 &0.6437 &0.1958 &0.2432 &0.5778 &0.5339 &0.3272 &0.3765 &0.6018 &0.6072&63.69&0.5764\\
       \hdashline
    $\clubsuit$VSFA \cite{VSFA}&0.3750 &0.3898 & 0.6227 & 0.5987 &0.2438 &0.2645 &0.5858 &0.5777 &0.3320 &0.3556 &0.6102 &0.5983&57.09&0.5204\\
    $\clubsuit$BVQA \cite{li2022blindly}&0.3089 & 0.3512 &0.5030 &0.6506 & 0.2379 & 0.2675 &0.4674 &0.6009 &0.3023 &0.3361 &0.4897 &0.6390&58.47&0.4781\\
    $\clubsuit$SimpleVQA \cite{sun2022deep}&0.5631 & 0.5670 & 0.8038 & 0.7873 &0.3474 &0.3487 &0.7273 &0.7327 &0.4884 &0.4979 &0.7620 &0.7756&60.78&0.6894\\
    $\clubsuit$FAST-VQA \cite{wu2022fast}&0.6391 & 0.6394 & 0.8945 & 0.9063 & 0.3919 &0.4162 &0.8376 &0.8492 &0.5563 & 0.5738 &0.8683 &0.8952&66.27&0.8122\\ 
    $\clubsuit$DOVER \cite{wu2023dover}& 0.6414 & 0.6552 & 0.8874 & 0.9244 &0.3759 &0.3936 &0.8038 &0
    .8448 & 0.5480 &0.5702 &0.8496 &0.9043&62.61&0.7662\\
     \hdashline
     $\heartsuit$CLIPScore~\cite{hessel2021clipscore} &0.0947 &0.1348 &0.0300 &0.1732 &0.2290 & 0.2818 & 0.1408 & 0.3835 &0.1738 &0.2256 &0.0848 &0.2625&58.27&0.1695\\
     $\heartsuit$BLIPScore~\cite{li2022blip}&0.1884 &0.2341 &0.2111 &0.3436 &0.3163 &0.3813 & 0.3451 & 0.5354&0.2735 &0.3335 &0.2814 &0.4277&63.93&0.3775\\
     $\heartsuit$AestheticScore~\cite{schuhmann2022laion}&0.5524&0.5613 &0.7566 &0.8053 &0.3931 & 0.4149 & 0.7001 & 0.7634 &0.5107 &0.5304 &0.7206 &0.7991&64.87&0.6734\\
     $\heartsuit$ImageReward~\cite{xu2023imagereward}&0.4180 &0.4472 &0.8016 &0.8097 &0.5076 & 0.5419 & 0.8549 & 0.8790&0.4992&0.5365&0.8331&0.8495&68.33&0.8586\\
     $\heartsuit$PickScore~\cite{kirstain2023pick}&0.4026 &0.4193 &0.8198 &0.8344 &0.4135 & 0.4246 & 0.7775 & 0.8497&0.4395&0.4580&0.8033&0.8522&62.29&0.7844\\
    $\heartsuit$HPSv2~\cite{wu2023human}&0.5415 &0.5690&0.7504&0.8096& 0.4989 & 0.5325 & 0.7522 & 0.8379&0.5605&0.5980&0.7468&0.8325&67.68&0.7789\\
    $\heartsuit$VQAScore~\cite{li2024evaluating}&0.1677&0.1961&0.3437&0.3365& 0.1763 & 0.2049 & 0.3922 & 0.4173&0.1880&0.2176&0.3811&0.3746&52.97&0.3280\\
    $\heartsuit$FGA-BLIP2~\cite{han2024evalmuse40kreliablefinegrainedbenchmark}&0.5181&0.5340&0.8954&0.8962& 0.5962 & 0.6081 & 0.9502 & 0.9543 &0.6035&0.6196&0.9297&0.9326&67.06&0.9372\\
    \hdashline
    $\bigstar$DeepseekVL2 (1B)~\cite{wu2024deepseekvl2mixtureofexpertsvisionlanguagemodels}& 0.0121 & 0.0064 & 0.0607 & 0.0427 & 0.0173 & 0.0252& 0.0785 &0.0288&0.0012&0.0137&0.0389&0.0554 & 39.29&0.0965\\
    $\bigstar$VideoLlava (7B)~\cite{lin2023video} & 0.1809 & 0.1938 & 0.6125 & 0.6817 &0.2005&0.2127 &0.6406 &0.6700&0.2026&0.2267&0.6764&0.7505& 68.46&0.7548\\
    $\bigstar$mPLUG-Owl3 (7B)~\cite{ye2024mplug} &0.3532 & 0.3542 & 0.7962 & 0.7789 &0.5478 &0.5551&0.9310 &0.9273&0.5540&0.5626&0.9008&0.8886& 63.02&0.8897\\ 
   $\bigstar$VideoLlama3 (8B)~\cite{damonlpsg2025videollama3} & 0.3922 & 0.4196 & 0.9073 &0.9144&0.4228&0.4817&0.9075&0.8569 &0.5063&0.5432&0.9576&0.9550& 70.16&0.8244\\
   $\bigstar$Qwen2-VL (7B) \cite{wang2024qwen2}&0.3568 &0.3615 &0.7085 &0.7588 &0.4498 &0.4789 &0.7953 &0.8464 &0.4105 &0.4313 &0.7775 &0.8127 &71.56&0.8681\\
 $\bigstar$Qwen2.5-VL (7B)~\cite{Qwen2.5-VL} &0.5410 & 0.5410 & 0.8652 & 0.8683 &0.5110&0.5223&0.8167&0.8692&0.5578&0.5799&0.8888&0.8949&62.34&0.6655\\
   $\bigstar$Llama3.2-Vision (11B)~\cite{meta2024llama} & 0.0940 & 0.0752 & 0.4483 & 0.5023 &0.0804&0.0530&0.2783&0.1784
    &0.1064&0.0792&0.4061&0.4294& 62.19&0.6033\\
  $\bigstar$CogAgent (18B)~\cite{hong2024cogagentvisuallanguagemodel} & 0.1244 & 0.1242 &0.4834 & 0.5906 &0.1190&0.1129&0.8198&0.8363&0.1481&0.1471&0.6311&0.7133& 65.32&0.7763\\
    $\bigstar$InternVL2.5 (8B)~\cite{chen2024expanding} & 0.2799 & 0.3336 & 0.7882 & 0.7978&0.4856&0.5268&0.9390&0.9483&0.4715&0.5164&0.9057&0.9084& 66.30&0.8051\\
    $\bigstar$InternVL3 (9B)~\cite{wang2024mpo} & 0.2731 &0.3519 &0.8300 &0.7576 & 0.4768 &0.5334 &0.9373 &0.9610 &0.4793 & 0.5307 & 0.9270 &0.9224& 65.82&0.7719\\
   $\bigstar$InternVideo2.5 (8B)~\cite{wang2025internvideo} & 0.1563 & 0.5454 &0.3361 &0.6574 &0.4978&0.5538&0.9560&0.9646 &0.4430&0.5366&0.9079&0.8968& 70.64&0.8435\\
   $\bigstar$LLaVA-NeXT (8B)~\cite{liu2024llavanext} &0.4888 & 0.5002 &0.8785 &0.8956 &0.2847&0.3090&0.8042&0.8510&0.4910&0.5194& 0.8630&0.9116& 70.21&0.9201\\
   $\bigstar$InternVL2.5 (38B)~\cite{chen2024expanding} &0.6227 &0.5766 &0.9052 &0.8814 &0.6470 &0.6417 &0.9586 &0.9646 &0.6627 &0.6497 &0.9471 &0.9330 &75.81 &0.9456\\
   $\bigstar$InternVL3 (38B)~\cite{wang2024mpo}
   &0.4950 &0.5150 &0.8118 &0.8082 & 0.5996 &0.5920 &0.9439 &0.9463 &0.6040 &0.5984 &0.9128 &0.8946 &73.89 &0.9386\\
   $\bigstar$Qwen2-VL (72B) \cite{wang2024qwen2}&0.4628 &0.5333 &0.8388 &0.8263 &0.5598 &0.5720 &0.9271 &0.9444 &0.5948 &0.6237 &0.9382 &0.9239 &73.12 &0.9064\\
   $\bigstar$Qwen2.5-VL (72B)~\cite{Qwen2.5-VL}&0.4245&0.4562&0.7762&0.7602& 0.6272 &0.5884 &0.9364 &0.9522 &0.5891 &0.5794 &0.9239 &0.9083 &73.83 &0.9516\\
   $\bigstar$Llava-one-vision (72B)~\cite{xiong2024llavaovchat} &0.5291 &0.5196 &0.7829 &0.7689 &0.5702 &0.5510 &0.8741 &0.9055 &0.6010 &0.5827 &0.8478 & 0.8497 &73.31 &0.9112\\
   $\bigstar$InternVL2.5 (72B)~\cite{chen2024expanding} & 0.5383 &0.5394 &0.8843 &0.8303 &0.6612 &0.6326 &0.9542 &0.9513& 0.6417 &0.6324 &0.9146 &0.9039 &75.18&0.9423\\
   $\bigstar$InternVL3 (72B)~\cite{wang2024mpo} & 0.5441 &0.4973 &0.8923 &0.8131 & 0.6314 & 0.6055 &0.9444 &0.9529 &0.6405 &0.6047 &0.9212 &0.9014 & 74.59&0.9623\\
    \hdashline
     $\triangle$Gemini1.5-pro \cite{Gemini}& 0.4972 &0.5012 &0.8790 &0.7965 &0.6095 &0.5874 &0.9430 &0.9491 & 0.5957 &0.5985 &0.9346 &0.9196& 73.38&0.9512\\
     $\triangle$Claude3.5 \cite{Claude3.5}& 0.4267 &0.4711 &0.7602 &0.7343 &0.5827 &0.5809 &0.8919 &0.9238 &0.5532 &0.5822 &0.8598 &0.8481& 73.20&0.9395\\
     $\triangle$Grok2 Vision \cite{Grok2}& 0.5628 &0.5728 &0.8808 &0.8403 &0.6659 &0.6730 &0.9546 &0.9763 &0.6629 &0.6784&0.9399 &0.9368& 76.51&0.9469\\
     $\triangle$ChatGPT-4o \cite{GPT4}& 0.5263 &0.5233 &0.9048 &0.8905 &0.6639 &0.6343 &0.9458 &0.9555&0.6240 &0.6117 &0.9346 &0.9326& 74.84&0.9317\\
     \hdashline
    $\blacklozenge$InternVL3* (9B)~\cite{wang2024mpo} & 0.6421 & 0.6575 &0.9061 & 0.9426 &0.5965 &0.6115 & \blue{0.9671} &0.9659 & 0.6571 & 0.6926 & 0.9341 &0.9597 & \blue{78.36}&\blue{0.9754}\\
    $\blacklozenge$Qwen2.5-VL* (7B)~\cite{Qwen2.5-VL} &\blue{0.7868} &0.7996 & 0.9265 &0.9510 & \blue{0.7354} &\blue{0.7493} &0.9634 &0.9677 & 0.7798& 0.8046& \blue{0.9579} & 0.9716 &77.35 &0.9698\\
    $\blacklozenge$InternVideo2.5* (8B)~\cite{wang2025internvideo} & 0.7845 & \blue{0.8102} &\blue{0.9308} & \blue{0.9605} &0.6773 &0.6946 & 0.9608 &\blue{0.9679} & \blue{0.7802} & \blue{0.8094} & 0.9542 &\blue{0.9721} & 73.37 &0.9136\\
     \rowcolor{gray!20} $\blacklozenge$LOVE (Ours) &\red{0.7932} &\red{0.8259}&\red{0.9324}&\red{0.9725}&\red{0.7466} &\red{0.7657} &\red{0.9778} &\red{0.9825} &\red{0.8115}&\red{0.8361}&\red{0.9657}&\red{0.9779} & \red{78.69}&\red{0.9769}\\


 
    \bottomrule[1pt]
  \end{tabular}}
  \label{mos}
  \vspace{-2em}
\end{table}
We use standard language loss during the instruction tuning phase and employ L1 loss for the quality regression task to minimize the difference between the predicted scores and the groundtruth values.


\vspace{-2mm}
\section{Experiments}
\vspace{-2mm}
In this section, we conduct extensive experiments to \textbf{benchmark V2T models} and evaluate the performance of our proposed model.
We first present the experimental setups in detail. Then we compare the performance of our model with current state-of-the-art VQA and LMM-based models in score prediction and visual question answering based on AIGVE-60K and five other AIGV benchmarks.
We launch further cross-dataset experiments to verify the generalizability of the proposed model.
Finally, we conduct ablation experiments to validate the model components. 
\vspace{-2mm}
\subsection{Experiment Setup}
\vspace{-2mm}
\label{setup}
To evaluate the correlation between the predicted scores and the ground-truth MOSs, we utilize three evaluation criteria: Spearman Rank Correlation Coefficient (SRCC), Pearson Linear Correlation Coefficient (PLCC), and Kendall’s Rank Correlation Coefficient (KRCC).  We compute correlations at two levels: \textbf{instance-level and model-level}. Instance-level evaluates how well the metric aligns with human ratings for each individual video, while model-level evaluates how well the metric ranks all 30 T2V models based on their average score performance.
For comparison models, we calculate the average score for both traditional handcrafted models and vision-language pre-training models by uniformly sampling eight frames from each video.
We load the pre-trained weights for zero-shot model inference. Using the same training and testing split (49,500:9,000) as our model, we fine-tune three of the LMM-based models for comparison. The training set contains AIGVs from 18 open-source T2V models, while the test set contains AIGVs from all 30 T2V models to test the metric scalability. The prompts used in the training and test sets are non-overlapping to ensure fair evaluation. The overall-averaged scores are computed by averaging the perception and T2V correspondence scores for each video. The QA accuracy of non-LMM-based models is obtained by converting the scores into 0/1 through K-Means clustering. 
The models are implemented with PyTorch and trained on a 40GB NVIDIA RTX A6000 GPU with batch size of 4. The initial learning rate is set to 1e-5 and decreased using the cosine annealing strategy. We employ Adam optimizer with $\beta_1 =0.9$ and $\beta_2 = 0.999$. We set the
number of instruction tuning epoch to 1 and the quality regression epochs to 2.

 \vspace{-2mm}
 \subsection{Evaluation on the AIGVE-60K Database}
 \vspace{-2mm}
\begin{table*}[t]
\vspace{-10mm}
\centering
\renewcommand\arraystretch{0.95}
\caption{Comparisons of the alignment between different metric results and human annotations in evaluating T2V model performance. We report the average scores and QA accuracy at model level. 
$\spadesuit$close-source T2V models unseen in LOVE training, $\heartsuit$open-source T2V models. $\triangle$Evaluation performance of all 30 models, $\bigstar$Zero-shot evaluation performance on 12 close-source T2V models.
}
\vspace{1mm}
   \resizebox{\linewidth}{!}{\begin{tabular}{l||>{\columncolor{mycolor_green}}c:>{\columncolor{mycolor_blue}}ccc|>{\columncolor{mycolor_green}}c:>{\columncolor{mycolor_blue}}ccc|>{\columncolor{mycolor_green}}c:>{\columncolor{mycolor_blue}}ccc|>{\columncolor{mycolor_green}}c:>{\columncolor{mycolor_blue}}c}
  \Xhline{1px}
    Dimension  &\multicolumn{4}{c|}{\textbf{Perception Score}}&\multicolumn{4}{c|}{\textbf{Correspondence Score}}&\multicolumn{4}{c|}{\textbf{Question Answering Accuracy (\%)}}&\multicolumn{2}{c}{\textbf{Overall Rank}}\\
  \cmidrule(lr){2-5}  \cmidrule(lr){6-9} \cmidrule(lr){10-13} \cmidrule(lr){14-15}
 Models
&$\text{Human}$&$\text{Ours}$&$\text{V-Aesthetic}$&$\text{DOVER}$ &$\text{Human}$&$\text{Ours}$&$\text{FGA-BLIP2}$&$\text{Grok2v}$&$\text{Human}$&$\text{Ours}$&$\text{Qwen2.5 (72B)}$&$\text{Gemini}$&$\text{Human}$&$\text{Ours}$\\
    \midrule
    $\spadesuit$Pixverse~\cite{pixverse_ai} & 63.81&62.46 &59.06 &67.42 &59.97 &61.65 &66.34 & 86.23& 91.33 &88.00 &80.67 & 76.67 & 1 & 1\\
    
     $\spadesuit$Wanxiang~\cite{alibaba2024tongyiwanxiang}
    & 60.54&61.54 &57.34 &63.86 &60.37 &60.84 &62.56 & 85.45& 90.33 & 89.33&84.00 & 80.00 &2 & 2\\
    $\spadesuit$Hailuo~\cite{hailuo} & 60.58 &62.40 &56.47 &65.37 &59.74 &60.63 &60.51 & 86.26 & 87.67 &87.33 & 84.00&83.33 & 3 & 4\\
    
    $\spadesuit$Jimeng~\cite{jimeng} & 65.25 & 65.72 &61.97 &74.16 &57.86 &59.12 &62.44 & 79.40 & 81.33 & 79.33& 72.33& 68.67 & 4 & 5\\
    $\spadesuit$Sora~\cite{videoworldsimulators2024}& 62.09& 62.71&58.23 &67.49 &59.68&60.33 &61.21 &84.81 & 85.67 &85.00 &78.33 & 80.00 & 4 & 2\\

     $\spadesuit$Hunyuan~\cite{li2024hunyuan} &58.81 & 61.45&54.48 &66.06 &57.25 & 59.29&60.32 &81.30 & 79.67 & 79.67& 75.00& 72.67& 6 & 6\\

     $\spadesuit$Vidu1.5~\cite{vidu_ai} & 54.56 &  55.19 &55.83 &47.61 &58.25 &58.37& 58.15&82.90 & 87.00 & 84.33 &83.00 & 83.33 & 7 & 8\\
     $\spadesuit$Gen3~\cite{runway2024gen3alpha}
    & 59.22&62.52 &59.21 & 66.17 &55.72 & 57.75&55.91 &74.77 & 75.33 & 79.67& 71.00&65.00 &8 & 6\\
    $\spadesuit$Kling~\cite{kling} & 60.56&61.34 &55.48 &61.58 &55.57 &58.21 &59.70 & 76.37& 73.67 &79.00 &73.67 &65.33 & 9 & 8\\

   $\spadesuit$Genmo~\cite{gemo} & 57.66 &59.56 &59.20 &62.13 &53.78 &56.69 &57.65 &73.39 & 75.67 & 77.00&71.33 & 66.67 &10 & 14\\
   $\spadesuit$ChatGLM~\cite{glm2024chatglm} & 56.39 & 60.96&56.81 &60.87 &53.98 & 57.78&56.11 &79.54 & 74.00 &77.33 & 72.67 & 73.33& 11 &10 \\
    $\spadesuit$Xunfei~\cite{xunfei}& 58.60 &61.79 &59.78 &67.53 &53.46 &56.00 &57.11 & 67.71 & 66.33 &74.67 &63.67 & 58.00 & 12 &13\\
    $\heartsuit$Pyramid~\cite{jin2024pyramidal} & 63.67 & 63.15&60.29 &67.46 &50.17 & 56.35&55.35 & 67.97& 50.17 & 71.33&63.67 & 57.33 &13 & 12\\
    $\heartsuit$Wan2.1~\cite{wan2025} & 57.27 & 63.37&60.29 &67.75 &52.33 &56.40 &53.93 &67.99 & 62.67&70.33 &60.33 & 56.67 &14 &10\\
    
    $\heartsuit$Allegro~\cite{allegro2024} & 56.08 & 59.45&52.20 &55.62 &50.70 &55.04 &53.82 &68.81 & 63.00 & 73.00&66.00 & 62.00 & 15 & 16\\

      $\heartsuit$VideoCrafter2~\cite{chen2024videocrafter2} & 48.11 & 56.99 &55.24 &49.67 &51.07 & 57.16&58.86 & 70.99 &65.67 &73.00 &78.67 &59.67 &16 &15\\
    $\heartsuit$CogVideo X1.5~\cite{yang2024cogvideox} & 50.59 &52.66 &52.16 &47.84 &49.73& 52.53 &50.40 &65.33 &64.67 & 63.67&60.67 & 56.67 &17 &21\\
    $\heartsuit$Animate~\cite{xu2024easyanimatehighperformancelongvideo} & 50.48 &53.38 &50.54 &42.13 &49.30 &53.09 &50.21 &65.36 & 60.67 &62.67 &61.33 & 57.33&18 &20\\
  
    $\heartsuit$Lavie~\cite{wang2023lavie} & 49.30 & 55.11&55.25 &52.71 &48.22&54.51 &51.57 &64.38 & 55.00 &67.33 &59.33 & 55.33 &19 &17\\
    $\heartsuit$Hotshot-XL~\cite{Mullan_Hotshot-XL_2023} &42.66 &49.38 &53.07 &46.52 &47.75 &53.26 &54.74 &68.15 & 57.67 &67.67 &65.00 & 58.00 &20 &19\\
    $\heartsuit$Latte~\cite{ma2025latte} & 43.81 &51.10 &53.29 &46.09 &46.73 & 53.29&51.96 &65.78 & 54.33  & 62.33&60.33 &55.33 &21 &22\\
    $\heartsuit$VideoCrafter1~\cite{chen2023videocrafter1} & 44.12 & 45.04&44.81 &36.64 &44.67 &50.85 &48.58 & 60.92& 46.00 & 54.00& 55.33& 52.67 &22 &24 \\

     $\heartsuit$Text2Video-Zero~\cite{text2video-zero} &40.53 & 46.03&57.55 &48.45 &44.89 & 51.69&51.39 &61.94 & 48.67 &60.67 &55.67 &57.67 &23 &23\\
    $\heartsuit$NOVA~\cite{deng2024nova} & 41.18 &50.47 &50.12 &45.39 &47.18 &53.77 &52.86 &70.03 & 56.00 &65.67 &66.00 & 59.33 &24 &18\\

    $\heartsuit$ModelScope~\cite{wang2023modelscope} &38.00 &41.22 &46.15 & 29.92&43.73 &48.99 &45.54 &59.99 & 47.33 & 51.33&58.00 &53.00 &25 &26\\

    $\heartsuit$Tune-A-Video~\cite{wu2023tune} &35.41 &40.87 &53.75 &42.34 &42.69 &49.81 &49.46 &59.85 & 43.00 & 57.67& 57.33 &56.33 &26 &25\\
    $\heartsuit$LTX~\cite{HaCohen2024LTXVideo} &40.11 &42.70 &47.09 &46.13 &41.28&48.08 &42.79 &55.83 & 37.00& 48.00&57.00 &46.67 &27 &27\\
    
    $\heartsuit$LVDM~\cite{he2022lvdm} &33.84 & 38.29& 40.63&29.40 &42.20 & 48.00&43.43 &56.90 & 40.33 &43.00 & 56.00& 51.67 &28 &28\\
    $\heartsuit$ZeroScope~\cite{Zeroscope} &30.08 & 37.58 &42.92 &34.89 &34.69 &43.76 &38.99 &42.18 & 22.00 & 32.67& 45.33 &37.67 &29 &29\\
    $\heartsuit$LWM~\cite{liu2024world} &27.39 &32.98 &40.72 &24.97 &31.49 &40.96 &35.71 &39.12 & 9.00 &25.00 & 37.00 &31.00 &30 &30\\
   \hline
   $\triangle$SRCC   to  human $\uparrow$& - &\red{0.932} &0.774 &\blue{0.887} & - &\red{0.978} &0.950 &\blue{0.955} & - &\red{0.977} &\blue{0.952} &0.951 & - &\red{0.977}\\
   $\triangle$RMSE to human $\downarrow$& - &\red{4.606} &7.427 &\blue{6.005}& - &\red{5.014} &\blue{4.067} &19.46 & - &\red{7.695} &10.11 &\blue{8.816} & - & \red{1.844}\\
   \hdashline
   $\bigstar$SRCC   to  human $\uparrow$& - &\red{0.825} &0.266 &\blue{0.650} & - &\red{0.944} &\blue{0.811} &0.909 & - &\red{0.904} &\blue{0.828} &0.807 & - &\red{0.942}\\
   $\bigstar$RMSE to human $\downarrow$& - &\red{2.241} &\blue{3.234} &5.954& - &\red{2.047} &\blue{3.274} &22.98 & - &\red{3.575} &\blue{5.681} &8.769 & - & \red{1.225}\\
    \Xhline{1px}
  \end{tabular}}\label{model}

 \vspace{-3mm}
\end{table*}
\pdfoutput=1
\begin{figure}[!t]
    \centering
    \includegraphics[width=1\linewidth]{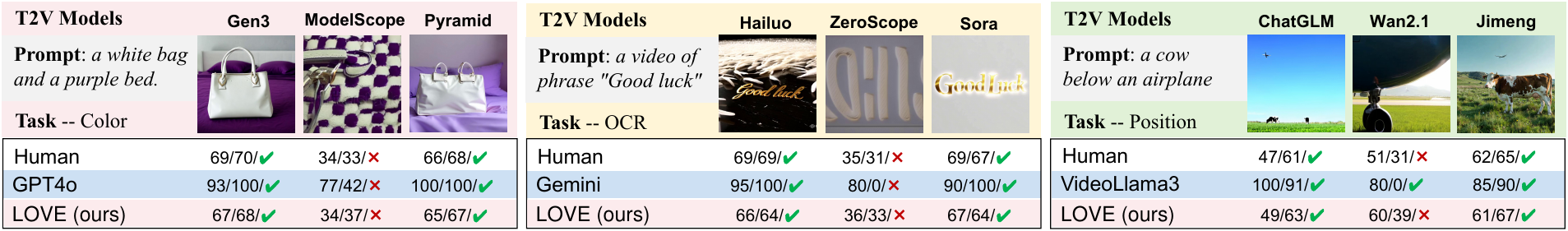}
     \vspace{-7mm}
    \caption{Visualization of the Perception/Correspondence/QA prediction from different V2T interpretation methods compared to human annotation. } 
     \vspace{-5mm}
    \label{example}
\end{figure}

 As shown in Table \ref{mos}, handcrafted metrics such as NIQE~\cite{mittal2012making} and QAC~\cite {xue2013learning}, show poor performance, indicating their features handcrafted mainly for natural scenes are ineffective for evaluating AIGVs. VBench~\cite{huang2024vbench}'s fragmented evaluation framework, which combines multiple detection-based and pretrained metrics in a piecemeal fashion dependent on arbitrary threshold settings, leads to inconsistent cross-dimensional results.
Vision-language pre-training models such as BLIPScore~\cite{li2022blip} and VQAScore~\cite{li2024evaluating} perform poorly in perception dimension due to their focus on T2V correspondence and overlook on perceptual quality.  
Although deep learning-based VQA methods achieve relatively better results, they still fall short in the T2V correspondence dimension.
 In contrast, LMM-based models achieve remarkable zero-shot generalization, especially in handling complex visual question-answering tasks, which highlights LMMs' inherent advantages in VQA tasks: they eliminate the need for VBench~\cite{huang2024vbench}'s fragile threshold tuning while overcoming traditional VQA's narrow specialization through unified multimodal understanding. The consistent performance of LMMs across all evaluation dimensions validates LMM's robustness as a comprehensive solution for AIGV assessment.
Building upon these insights, we propose a LMM-based model that establishes an \textbf{\textit{all-in-one}} evaluation framework and achieves superior performance in both score prediction and visual question answering, making it a more comprehensive method for evaluating AIGVs at both instance-level and model-level. Detailed results and findings are in \textit{Appendix} Sections \ref{analyze} and \ref{comparison}.
 
\begin{table*}[t]
  \centering
  \vspace{-10mm}
    \caption{Zero-shot cross-dataset correspondence performance on multiple benchmarks. *Refers to scores finetuned on the specific dataset. Best results are marked in {\red{RED}} and second-best in {\blue{BLUE}}. } 
  \label{tab:method}%
 
  \renewcommand\arraystretch{0.94}
  \resizebox{\textwidth}{!}{
    \begin{tabular}{l|cc|cc|cc|cc|cc|cc}
    \Xhline{1px}
    \multirow{2}[1]{*}{Method}& \multicolumn{2}{c|}{\textbf{AIGVE-60K}} & \multicolumn{2}{c|}{FETV~\cite{liu2024fetv}} & \multicolumn{2}{c|}{T2VQA-DB~\cite{kou2024subjective}} & \multicolumn{2}{c|}{LGVQ~\cite{zhang2024benchmarking}} &  \multicolumn{2}{c|}{AIGVQA-DB~\cite{wang2024aigv}} & \multicolumn{2}{c|}{GenAI-Bench~\cite{li2024evaluating}} \\
\cline{2-13}     & SRCC  & KRCC & SRCC  & KRCC  & SRCC  & KRCC  & SRCC  & KRCC   & SRCC  & KRCC & SRCC  & KRCC \\
    \hline
    CLIPScore~\cite{hessel2021clipscore} & 0.095 &  0.064 & 0.607 & 0.498 & 0.049 & 0.033 & 0.446 & 0.301  & 0.152 & 0.101 & 0.536 & 0.180\\
    BLIPScore~\cite{li2022blip} &0.188  &  0.127 & 0.616 & 0.505 & 0.174 & 0.118 & 0.455 & 0.319  & 0.181 & 0.122 & 0.546 & 0.201\\
    ImageReward~\cite{xu2023imagereward} & 0.418 & 0.288 & 0.657 & 0.519 & 0.175  & 0.119 & 0.498 & 0.344  & 0.231 & 0.157 & 0.600 & 0.314\\
    PickScore~\cite{kirstain2023pick} & 0.403 & 0.276 & 0.669 & 0.533 & 0.239 & 0.163 & 0.501 & 0.353  & 0.262 & 0.176 & 0.568 & 0.248\\
    HPSv2~\cite{wu2023human} & \blue{0.542} & \blue{0.379} & \blue{0.686} & \blue{0.540} & \blue{0.243} & \blue{0.168} & 0.504 & 0.357  & 0.229 & 0.153 & 0.515 & 0.137 \\
    VQAScore~\cite{li2024evaluating} & 0.168 & 0.110 & 0.565 & 0.414 & 0.177 & 0.121 & \blue{0.553} & \blue{0.394} & \blue{0.444} & \blue{0.307} & \blue{0.632*} & \blue{0.382*} \\
    \hline
    \rowcolor{gray!20} LOVE (Ours)  &\red{0.747*} & \red{0.560*} &\red{0.724} & \red{0.555} &\red{0.363} & \red{0.252}
    &\red{0.602} & \red{0.438}
    &\red{0.552} & \red{0.400} &\red{0.682} & \red{0.517}\\


    \Xhline{1px}
    \end{tabular}%
  }
\end{table*}%
\vspace{-4mm}
\begin{table*}[t]
\vspace{-4mm}
\renewcommand\arraystretch{1.1}
  \caption{Ablation study on the temporal and quality-level features, LoRA strategy of LOVE.}

  \resizebox{1\textwidth}{!}{
  \begin{tabular}{ccccc|ccc:ccc:c|ccc}
    \Xhline{1px}
                                   & \multicolumn{4}{c}{\textbf{Feature \& Strategy}}                                  & \multicolumn{3}{c}{\textbf{Perception  (ours)}}                   & \multicolumn{3}{c}{\textbf{Correspondence (ours)}}                  & \multicolumn{1}{c|}{\textbf{QA (ours)}} &
                                   \multicolumn{3}{c}{\textbf{GenAI-Bench~\cite{li2024evaluating}}} \\
    \cline{2-15}
    \multicolumn{1}{c}{No.}        & Temporal    & Quality level & Vision$_{r=16}$    & LLM$_{r=16}$  & SRCC    & PLCC        &KRCC  & SRCC       & PLCC    & KRCC  &  Acc    &    SRCC     & PLCC   & KRCC              \\
    \hline
    \multicolumn{1}{c}{(1)}          &    &         &       \ding{52}   &      \ding{52}       & 0.642    &  0.658    &  0.458   &  0.597   & 0.612     & 0.397     &   78.4\%   & 0.550   & 0.553   &   0.412  \\
    \multicolumn{1}{c}{(2)}          & \ding{52}   &     \ding{52}     &         &           &  0.756   & 0.781   &  0.618   &    0.723  &  0.735   &  0.584   &  65.8\%    & 0.652    & 0.663        & 0.498   \\
    \multicolumn{1}{c}{(3)}          &   \ding{52}          &  \ding{52} &           &       \ding{52}     & 0.779     & 0.808    & 0.589     & 0.742    & 0.760   & 0.549    & 77.3\%     & 0.676     & 0.689     & 0.509     \\
    \multicolumn{1}{c}{(4)}          & \ding{52}   &  \ding{52} & \ding{52} &     & 0.781     & 0.813    & 0.593     & 0.739    & 0.758    & 0.552    & 76.8\%     & 0.674     & 0.690     & 0.506     \\
     \multicolumn{1}{c}{(5)}          &   & \ding{52} &    \ding{52}    &  \ding{52}       & 0.734     & 0.762     & 0.601    & 0.737     & 0.759     & 0.548     &  78.4\%    & 0.673     & 0.694     & 0.501    \\
    \rowcolor{gray!20} \multicolumn{1}{c}{(6)}          &   \ding{52}        &  \ding{52}         & \ding{52} &      \ding{52}         & \textbf{0.793}     & \textbf{0.826}    & \textbf{0.602}     & \textbf{0.747}    & \textbf{0.766}    & \textbf{0.560}    & \textbf{78.7\%}     & \textbf{0.682}     & \textbf{0.701}     & \textbf{0.517}     \\
   

    \Xhline{1px}
  \end{tabular}\label{ablation}
   }
  \vspace{-4mm}
  \centering
\end{table*}
\vspace{1mm}
 \subsection{Evaluation on Benchmarking T2V Model Performance}
 \vspace{-2mm}
We further conduct comparisons of the alignment between different metric results and human annotations in evaluating T2V model performance, as shown in Table \ref{model}. 
Our model achieves the highest SRCC with human ratings and the lowest relative Root Mean Square Error (RMSE) in score differences. This demonstrates that our model is well-aligned with human judgment at the model level in assessing and ranking the performance of T2V models. Moreover, our model trained on 18 T2V models exhibits strong \textbf{scalability} in 12 unseen T2V model rank predictions.  We also provide examples with model prediction scores at the instance level. As shown in Figure \ref{example}, LOVE generates scores that are more consistent with human annotations and achieves the highest accuracy.

\vspace{-2mm}
\subsection{Zero-shot Cross-dataset Evaluation}
\vspace{-2mm}
 Table \ref{tab:method} reveals LOVE's superior generalization capability. Our proposed method achieves state-of-the-art performance on both AIGVE-60K dataset and other five cross-dataset AIGV benchmarks.
\vspace{-6mm}
\subsection{Ablation Study}
\vspace{-2mm}
We do extensive ablation investigations to verify the contributions of the various modules in LOVE, as shown in Table \ref{ablation}. 
Our analysis reveals three main findings. First, experiments (1) and (5) demonstrate the effectiveness of text-defined quality-level initialization in model performance. Second, we confirm the notable performance improvements brought by LoRA fine-tuning strategy through experiments (2)–(4). Third, the efficacy of the temporal features is validated by experiments (5) and (6).
\vspace{-2mm}
\section{Conclusion}
\label{conclusion}
\vspace{-2mm}
In this paper, we introduce \textbf{AIGVE-60K}, the largest AIGV evaluation dataset to date, consisting of 58,500 videos generated by 30 T2V models using 3,050 prompts across 20 task-specific challenges and 2.6M subjective ratings from the perception, text-video correspondence, and task-specific accuracy, respectively. Based on AIGVE-60K, we benchmark and evaluate both the generation ability of T2V models and the V2T interpretation ability of LMMs. 
We also propose \textbf{LOVE}, a LMM-based evaluation model that leverages instruction tuning and LoRA adaptation to achieve AIGV perceptual quality evaluation and T2V correspondence attribution in terms of both model-level and instance level. Extensive experiments demonstrate that LOVE achieves state-of-the-art performance on the AIGVE-60K dataset and manifests strong zero-shot generalization ability on the other five benchmarks, highlighting the significance of both the AIGVE-60K dataset and the LOVE model.

\noindent
{\bf Limitations and Broader Impact.} 

The current rankings are based on data we obtained from random-selected professional annotators, and we do not intend to offend the developers of these excellent T2V and V2T models. Although our model shows promising scalability in evaluating AIGVs generated by new prompts and previously unseen T2V models, the effectiveness in real-world applications remains an open question. We expect our benchmark and dataset will contribute to the advancement of T2V generation, T2V evaluation, and V2T interpretation.

{
    \small
    \bibliographystyle{ieeetr}
    \bibliography{main}
}


\clearpage
\appendix
\part{Appendix} 
\parttoc

\section{Overview}
In this Appendix, we provide additional details on the Ethical Discussions, data collection, methodology, experiments, and results discussed in the main paper. We detail the broader impact and ethical discussions in Section \ref{ethical}. In data collection, we detail the 20 distinct tasks in Section \ref{tasks} and the overview of the 30 T2V models in Section \ref{modelss}. We then elaborate on the subjective experiments in Section \ref{exp}, including the annotation dimension, criteria, interface and management. In addition, we provide an in-depth analysis of the AIGVE-60K database, including MOS distributions and model performance comparisons across the 20 tasks in Section \ref{analyze}. We outline the loss functions used in the training process for the LOVE model in Section \ref{loss}. Details on the evaluation criteria and algorithms are also included in Section \ref{implementation}. Finally, we provide more performance comparisons between our model and other metrics in Section \ref{comparison}.
\section{Broader Impact and Ethical Discussions}
\label{ethical}
\subsection{Broader Impact of Our Research}
\label{ethical1}
We discuss how our work can be applied to benefit the community through three key contributions. \underline{\bf Firstly}, we resolve the long-standing limitation of evaluation subjectivity and scalability by constructing \textbf{AIGVE-60K}, the \textbf{largest} AIGV evaluation dataset with 58,500 videos and 2.6M human annotations explicitly disentangled into \textit{perceptual quality} (visual artifacts, temporal consistency), \textit{text-video correspondence} (prompt-video alignment), and \textit{task-specific accuracy}. Unlike prior datasets with coarse merged scores \cite{liu2024fetv} or narrow expert judgments (Table~\ref{tab:relate}), our fine-grained annotations eliminate hidden bias in human preference modeling while supporting both instance-level and model-level analysis. 
\underline{\bf Secondly}, we confront the methodological fragmentation in current evaluation ecosystems where isolated metrics like FVD \cite{unterthiner2018towards} lack prompt awareness and VBench's threshold-dependent pipelines \cite{huang2024vbench} introduce inconsistent standards. Our \textbf{LOVE} framework establishes ethical transparency through an \textbf{\textit{all-in-one}} LMM architecture with dual vision-temporal encoders and instruction tuning, unifying perceptual/correspondence/accuracy evaluation under reproducible criteria without arbitrary thresholds. 
\underline{\bf Thirdly}, we pioneer \textbf{\textit{bidirectional benchmarking}} that benchmarking and evaluating 30 T2V video generation models and 48 V2T interpretation models on shared standards. By open-sourcing annotations and maintaining minimal baseline tuning (LoRA adapters with 2-epoch training), we ensure community accessibility for: (1) engaging more T2V video generation models and V2T interpretation models for comparison, (2) developing more effective models based on our framework and training strategies. 

\subsection{Ethical Discussions of Data Collection}
\label{ethical2}
We detail the ethical issues that may emerge during the dataset collection process. All participants in the subjective evaluation are clearly informed of the contents in our experiments.
Specifically, we addressed the ethical problems by getting a written and informed permission from each person featured in the dataset stating that they approved their subjective ratings being used for non-commercial research, thus equipping it with such legal and ethical qualities.
The experiments do not contain any visually improper or NSFW content (both {\it textual} and {\it visual}), because we used extensive manual review during the AIGV generation stage. 
A total of 15 annotators, all postgraduate students in related fields from our laboratory, contributed to this process over 1.5 months, dedicating 3–4 hours daily to annotation tasks. We grouped the 58,500 analyzed videos into 30 sessions due to their vast volume.
Each participant received \$20 per session in accordance with the current ethical standard \cite{silberman2018responsible,otani2023toward}.
The experiment took more than a month to complete, with each participant contributing an average of 60 hours.
All associated AIGVs and their corresponding prompts in the {\bf AIGVE-60K} dataset are released under the {\bf CC BY 4.0} license.

\section{Prompts and Task-specific Challenges}
\label{tasks}

\begin{figure}[!t]
\vspace{-5mm}
	\centering
	\includegraphics[width=1\linewidth]{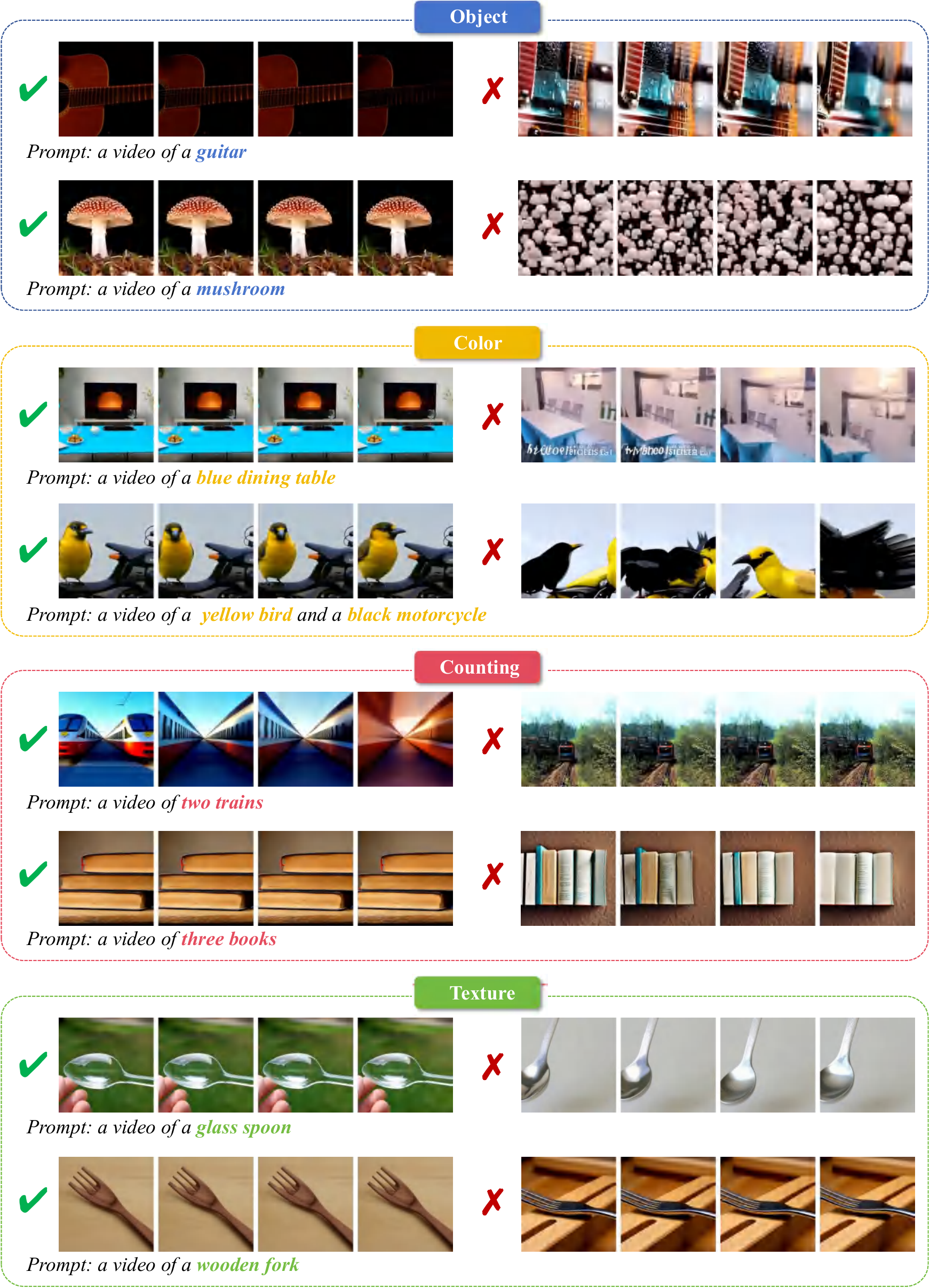}
 \vspace{-2mm}
	\caption{Examples for different task-specific challenges.}
 \vspace{-3mm}
	\label{task1}
\end{figure}

\begin{figure}[!t]
\vspace{-5mm}
	\centering
	\includegraphics[width=1\linewidth]{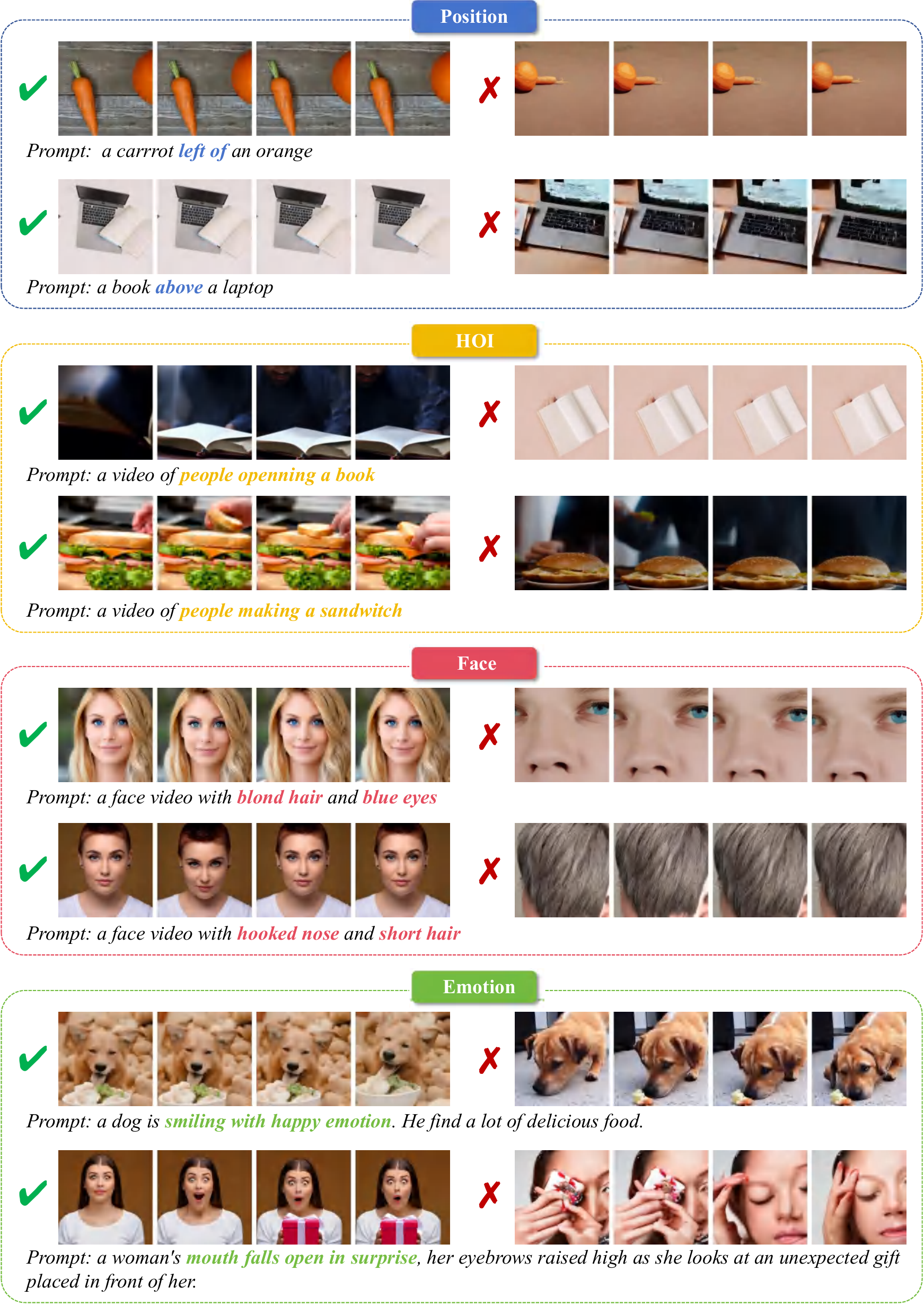}
 \vspace{-2mm}
	\caption{Examples for different task-specific challenges.}
 \vspace{-3mm}
	\label{task2}
\end{figure}

\begin{figure}[!t]
\vspace{-5mm}
	\centering
	\includegraphics[width=1\linewidth]{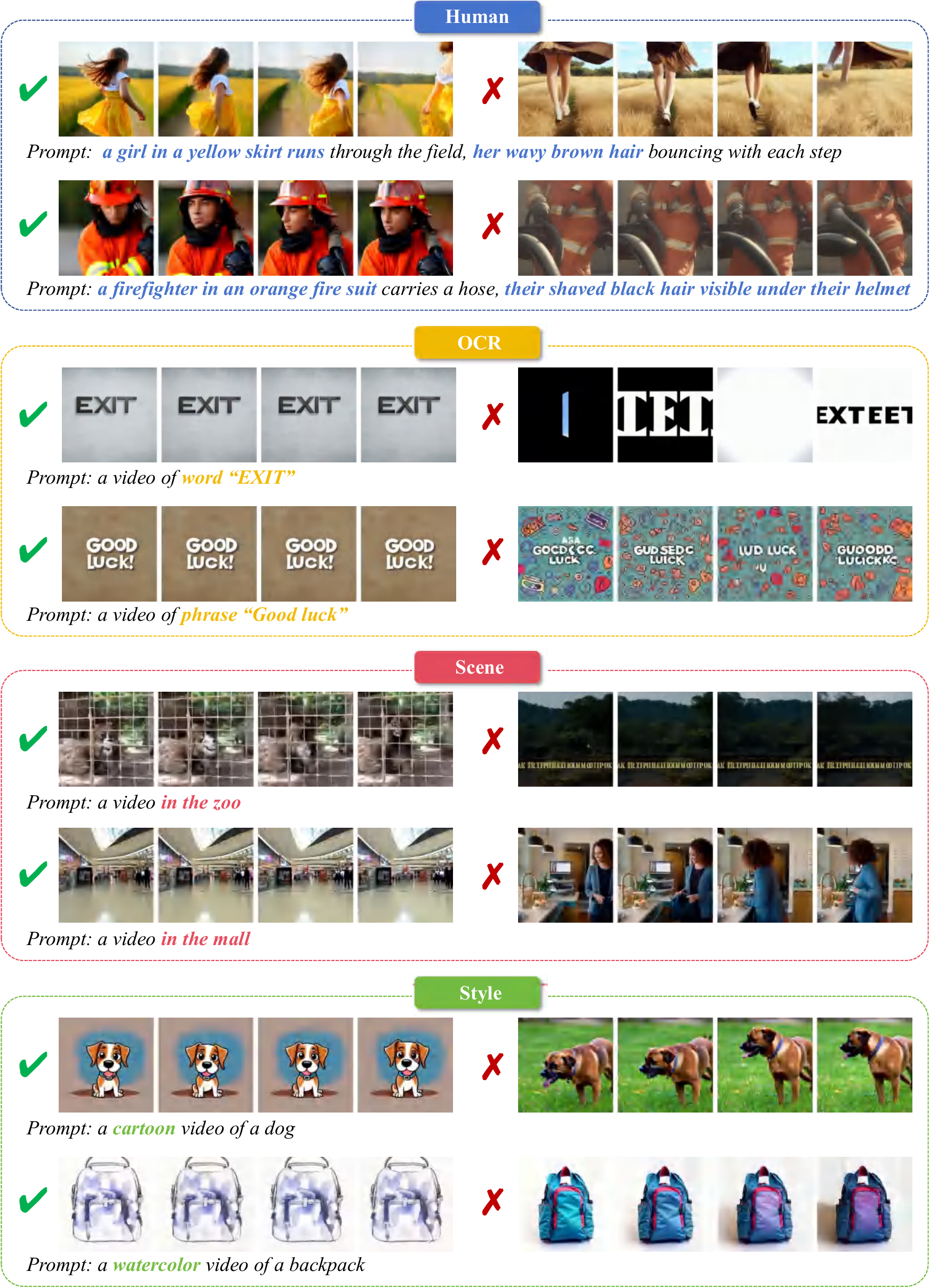}
 \vspace{-2mm}
	\caption{Examples for different task-specific challenges.}
 \vspace{-3mm}
	\label{task3}
\end{figure}

\begin{figure}[!t]
\vspace{-5mm}
	\centering
	\includegraphics[width=1\linewidth]{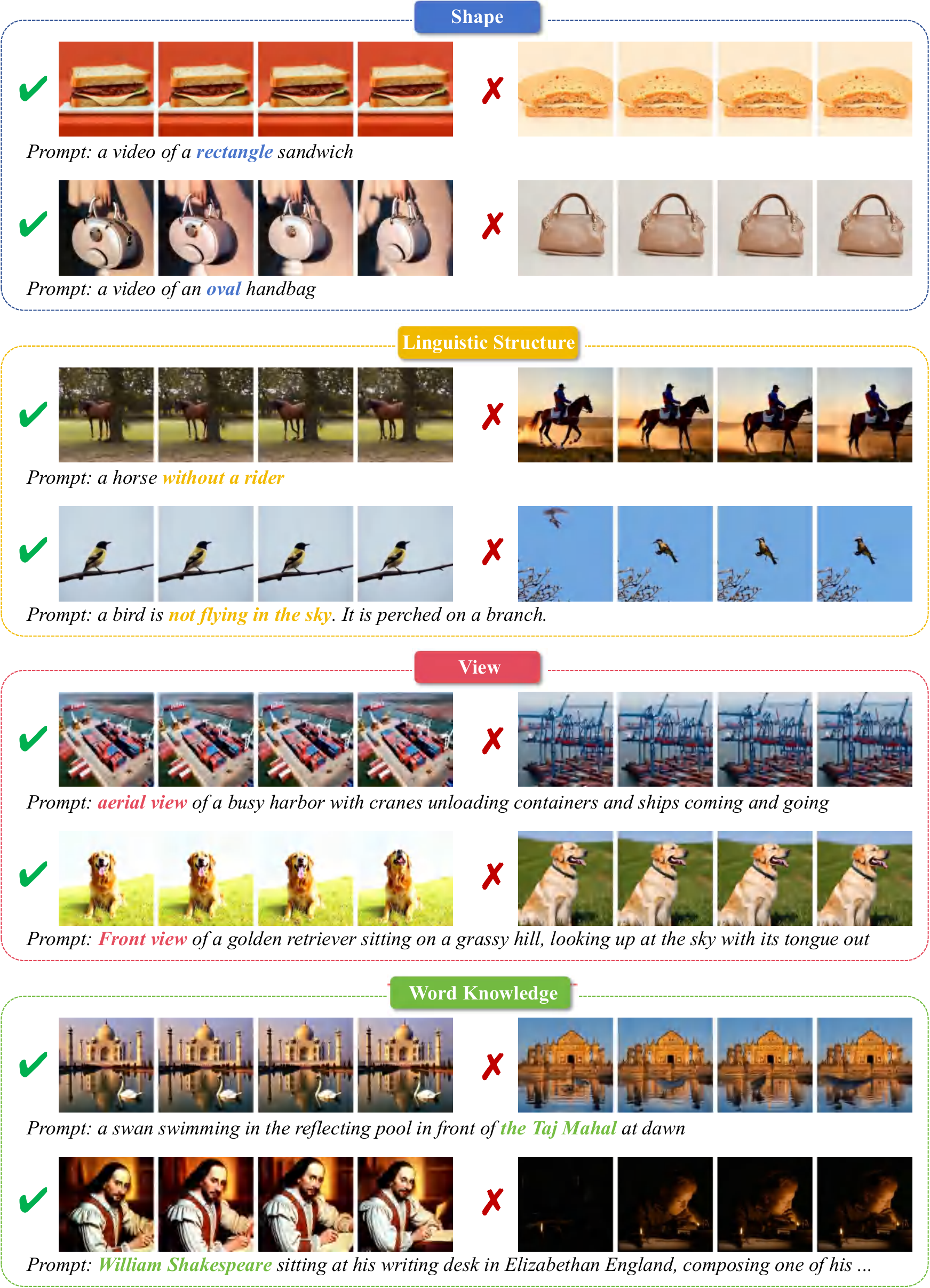}
 \vspace{-2mm}
	\caption{Examples for different task-specific challenges.}
 \vspace{-3mm}
	\label{task4}
\end{figure}

\begin{figure}[!t]
\vspace{-5mm}
	\centering
	\includegraphics[width=1\linewidth]{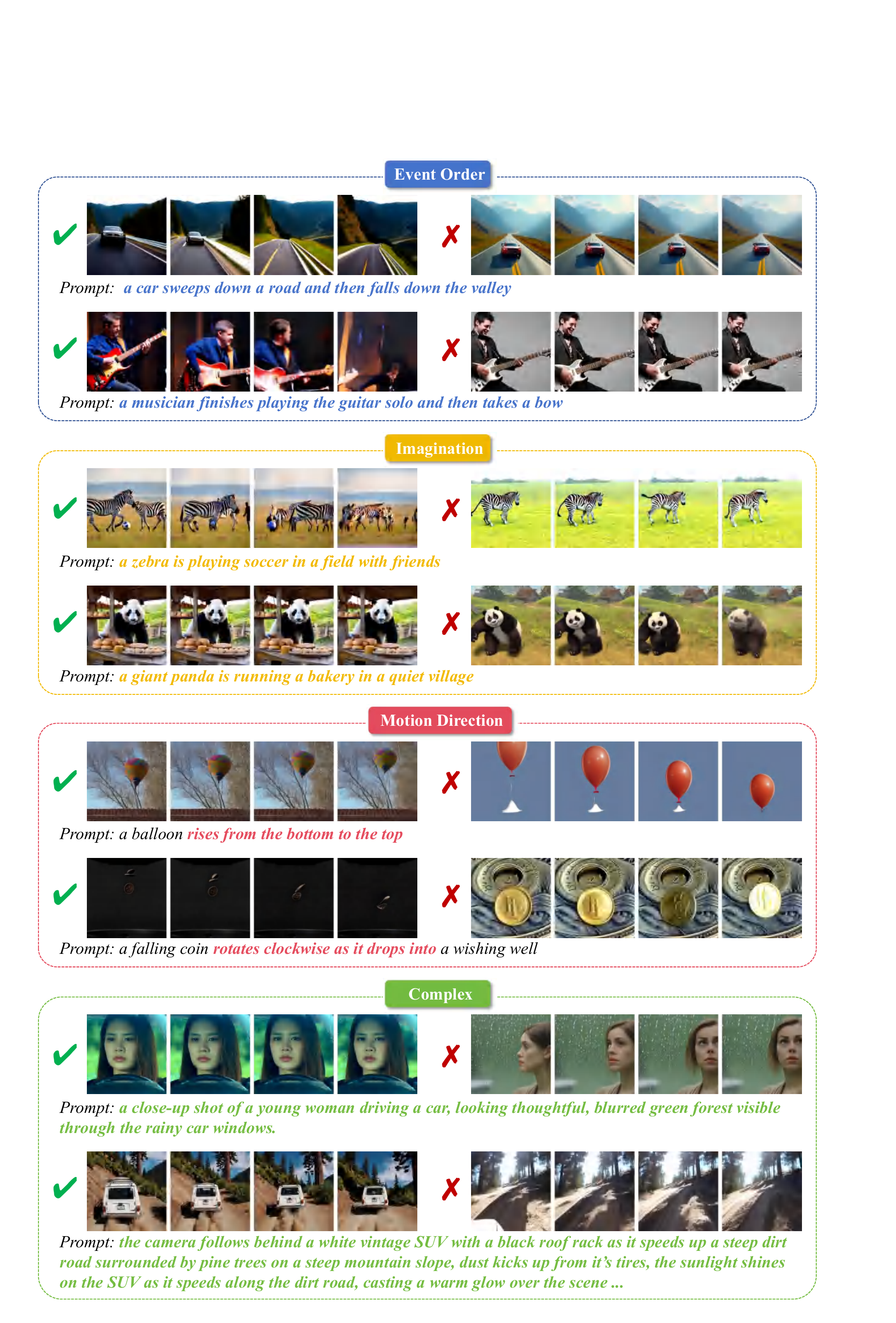}
 \vspace{-2mm}
	\caption{Examples for different task-specific challenges.}
 \vspace{-3mm}
	\label{task5}
\end{figure}

In this study, we systematically investigate the capabilities of text-to-video generation models through a comprehensive evaluation framework. We focus on 20 distinct tasks that vary in complexity and require diverse compositional skills, as detailed in Table \ref{prompt} with their corresponding subcategories, keywords, and example prompts. Prompts of the AIGVE-60K are primarily sourced from 9 existing open-domain real-world text-video pair datasets and AIGC datasets, including InternVid \cite{wang2023internvid}, MSRVTT \cite{xu2016msr-vtt}, WebVid \cite{Bain21}, TGIF \cite{li2016tgif}, FETV \cite{liu2024fetv}, AIGVQA-DB~\cite{wang2024aigv}, GenEval~\cite{ghosh2023geneval}, Sora\cite{videoworldsimulators2024} and  Runaway \cite{runway2024gen3alpha} website. The prompts are classified into 20 tasks according to the subcategories and keywords.
The tasks are adapted from GenEval~\cite{ghosh2023geneval} (6 tasks), T2I-CompBench~\cite{huang2023t2i} (3 tasks), EvalkMi-50K \cite{wang2025lmm4lmmbenchmarkingevaluatinglargemultimodal} (9 tasks), and defined by common AIGC prompt categories (e.g., PartiPrompts~\cite{yu2022scalingautoregressivemodelscontentrich}). To avoid overfitting to template-like input, DeepSeek R1~\cite{guo2025deepseek} to expand and modify the prompts to ensure clarity and diversity, while human verification ensured that task integrity and alignment with evaluation goals were preserved. 
These tasks are carefully designed to assess different aspects of model performance, ranging from basic object rendering to complex spatial and attribute understanding, as shown in Figures \ref{task1}-\ref{task5}. Below, we provide an overview of the main task categories and their associated challenges.
\begin{table*}
\vspace{-7mm}
\caption{Prompt categories with corresponding keywords and examples.
}
\label{prompt}
\renewcommand\arraystretch{1.2}
\resizebox{\linewidth}{!}{
    \begin{tabular}{llp{6cm}}
    \toprule
   Category   & Subcategory / Keywords  &  Prompt examples\\
    
    \midrule   
     Object &  clock, person, bicycle, car, motorcycle, airplane, bus, train, truck, boat,  \dots &   A video of a clock \\
     
     Color &  red, orange, yellow, green, blue, purple, pink, brown, black, white, \dots & A video of a green clock\\
     
     Counting& zero, one, two, three, four, five, six, seven, eight, nine, ten, \dots &A video of four clocks \\
     Texture& glass, cement, stone, rubber, fabric, ceramics, leather, metallic, wooden, plastic, \dots &A video of a wooden forks\\

     Position&left of, right of, above, below, \dots & A video of a cow below an airplane\\

     HOI&hold a stop sign, operate an oven, peel an apple, lie on a bench, carry a book, \dots &a video of people reading a book \\

     Face&hair, mouth, emotion, eyes, necklace, cheeks, nose, skin, \dots &a face video with green eyes and green hair \\

     Emotion&happy, sadness, love, fear, surprise, anger, worry, neutrality &A dog is smiling with happy emotion\\

     Human&human, cloth, cloth-color, hair, hair-color & A man in a blue shirt smiles warmly, his curly black hair framing his face\\
     
     OCR&“HELLO”, “STOP”, “SUCCESSFUL”, “Let it go”, “Thank you”, \dots &A video of phrase “Good luck” \\

     Scene& kitchen, living room, street, swimming pool, playground, waterfall, forest, \dots&A video in the forest \\

     Style&cartoon, realistic, oil painting, vintage, watercolor, line drawing, \dots &A watercolor video of a backpack \\

     Shapes&circle, cylinder, sphere, star, triangle, rectangle, irregular, oval, linear, cone, \dots & A video of a circle chair\\

     View & close-up, ground view, aerial view,  first-person view, wide-angle view, \dots& Aerial view of a harbor \\

     World Knowledge&Great Wall, Great Pyramid, Ha Long Bay, Machu Picchu, Eiffel Tower, Grand Canyon, \dots& Boats in Ha Long Bay\\

    Linguistic Structure& without, no, not, \dots& The garden has no flowers blooming.\\
    Imagination& imaginative objects and scenes, impossible scenarios in the real world, \dots&A panda is flying in the sky \\
     
     Motion Direction&left-to-right, up-to-down, rotate, zoom in, zoom out & A balloon rises from the bottom to the top\\  

     Event Order& and then, \dots & The flowers first wilt and then bloom again\\
     
     Complex& Counting + Color + Shapes + Scene, Style + Color + Position, Human + Emotion, \dots &A video of four blue birds playing on a circle playground \\
      
        \bottomrule
    \end{tabular}
}
\centering
\vspace{-3mm}
\end{table*}
\begin{itemize}
\item \textbf{Object}: evaluates a model's ability to generate a specified object class. The challenge lies in generating a single object or multiple objects.  The model may be required to generate a single object from a predefined category, such as a “cat”, or to combine several objects from distinct categories, such as “dog” and “tree” into a coherent scene. 

\item \textbf{Color}: evaluates a model's proficiency in associating specific color attributes with generated objects. The challenge involves not only generating an object with a single color applied but also handling more complex cases where multiple colors are used across multiple objects. The evaluation focuses on the model's ability to accurately bind color properties to target objects while preserving the integrity and consistency of the objects themselves.

\item \textbf{Counting}: evaluates a model's ability to generate a specific number of objects in a scene. The challenge includes numerical understanding and managing multiple instances without overlap or spatial issues, especially for larger numbers.

\item \textbf{Texture}: evaluates a model's capability to render objects with specific surface textures and material properties (\textit{e.g.}, metallic, wooden, glass). The challenge lies in creating realistic textures that match the object's properties and lighting conditions.

\item \textbf{Position}: evaluates a model's capability to render two objects with specified positional relationships. The challenge encompasses not only object generation but also the accurate representation of specific spatial relationships (\textit{e.g.}, above, below, left of, right of). This requires precise control over object arrangement while maintaining their identities.

\item \textbf{HOI (Human-Object Interaction)}: evaluates a model's ability to generate realistic interactions between humans and objects, ensuring the actions are physically plausible. The challenge is to create recognizable humans and objects while maintaining natural spatial and logical relationships.

\item \textbf{Face}: evaluates a model's ability to generate human faces with specific features (\textit{e.g.}, face shape, nose structure, hairstyle). The challenge is to create realistic and diverse facial representations while maintaining feature consistency, and testing the model's understanding of facial anatomy.

\item \textbf{Emotion}: evaluates a model's ability to convey specific emotions or moods, either through human facial expressions (\textit{e.g.}, happiness, sadness) or through the overall atmosphere of a scene (\textit{e.g.}, serene, love). This evaluates the model's understanding of emotional cues and its ability to translate abstract emotions into visuals.

\item \textbf{Human}: evaluates a model's ability to generate human figures with specific occupational attire, unique accessories, and hairstyles. The challenge lies in creating realistic and coherent human representations while maintaining consistency across these attributes. 

\item \textbf{OCR (Optical Character Recognition)}: evaluates a model's capability to generate readable text within videos, such as words or short sentences. The challenge is to make the text visually coherent with the video and machine-readable by OCR systems, testing the model's understanding of typography and text integration.

\item \textbf{Scene}: evaluates a model's ability to create complex scenes with multiple naturally composed elements in a specific environment (\textit{e.g.}, beach, forest, kitchen). The challenge is to ensure all objects and backgrounds are contextually relevant and spatially consistent, evaluating the model's holistic scene understanding.

\item \textbf{Style}: evaluates a model's proficiency in generating videos in specific artistic styles (\textit{e.g.}, watercolor, oil painting, cartoon). The challenge is to mimic the style's visual characteristics while keeping objects and scenes recognizable, testing the model's ability to apply abstract stylistic concepts consistently.

\item \textbf{Shape}: evaluates a model's ability to generate objects with specific geometric shapes (\textit{e.g.}, spherical, rectangular, triangular, star) while preserving their recognizability. This tests the ability to abstract representations of real-world objects and express them in other shapes.

\item \textbf{View}: evaluates a model's ability to generate videos from specific viewpoints (\textit{e.g.}, first-person, third-person, side view). The challenge is to maintain correct spatial orientation, scale, and proportion across perspectives, testing the model's understanding of spatial geometry.

\item \textbf{World Knowledge}: evaluates a model's knowledge of real-world landmarks, historical sites (\textit{e.g.}, the Great Wall, Eiffel Tower, Great Pyramid), and the physical appearances of famous individuals (\textit{e.g.}, Albert Einstein).  The challenge lies in creating content that accurately aligns with people's perceptions of famous landmarks and the physical appearances of well-known individuals.

\item \textbf{Linguistic Structure}: evaluates a model's ability to interpret and render linguistic structures involving negation (\textit{e.g.}, “without,” “no”). The challenge is to generate videos that accurately reflect the absence of specified objects or features (\textit{e.g.}, a “classroom without people”) while maintaining scene integrity. This tests the model's comprehension of negative constructs.

\item \textbf{Imagination}: evaluates a model's ability to generate imaginative scenes that combine elements from different categories or depict impossible scenarios in the real world (\textit{e.g.}, “a dog is driving a car”). The challenge is to balance creativity with visual plausibility, evaluating the model's capacity for creative thinking and novel concept synthesis.

\item \textbf{Motion Direction}: evaluates a model's ability to generate or simulate motion accurately, ensuring that objects move in the intended direction. This can include simple motion along a single axis, such as left-to-right or up-to-down, as well as more complex trajectories, such as circular paths. It also involves assessing camera motion, where the model must generate realistic and coherent movement of the camera itself, whether it’s panning, zooming in/out, or rotating, to maintain consistent perspective and spatial relationships within the scene.

\item \textbf{Event Order}: evaluates a model's ability to correctly sequence events in a logical order. This typically involves understanding the chronological relationship between two events, where one occurs before the other. For example, the model may be tasked with recognizing that a certain action happens first and then follows another action. The challenge lies in ensuring that the events are correctly ordered.

\item \textbf{Complex}: is designed by combining simpler task components, such as color recognition, object counting, and shape identification, into more intricate and multifaceted challenges.  These tasks require models to integrate and execute multiple simple tasks simultaneously within a single video. Below are some combined forms of complex tasks along with corresponding examples:
\begin{itemize}
\item[(1)] \textbf{Counting + Color + Shapes + Scene}: A video of [number] [color] [class] [action] in a [shape] [scene].
\textbf{Example}: \textit{A video of two white dogs swimming in a triangle-shaped swimming pool}.

       \item[(2)] \textbf{Counting + Color + Shapes + Texture}: A video of [number] [color] [texture] [shape] [class].  
\textbf{Example}:\textit{ A video of two brown wooden rectangular books}.

\item[(3)] \textbf{HOI + Color + Shape + Texture}: A video of [human action] a [color] [texture] [shape] [object].  
\textbf{Example}: \textit{A video of people opening a yellow wooden triangle box}.

\item[(4)] \textbf{Style + Color + Position}: A [style] video of a [color1] [class1] [position] a [color2] [class2].  
\textbf{Example}: \textit{A cartoon video of a yellow dog to the left of a white cat}.

\item[(5)] \textbf{Style + OCR + Color}: A [style] video of [color] text “\texttt{[content]}”.  
\textbf{Example}: \textit{An oil painting of red text “\texttt{CONGRATULATIONS}”}.

\item[(6)] \textbf{OCR + Color + Object}: A video of [color1] text “\texttt{[content]}” on a [color2] [class].  
\textbf{Example}: \textit{A video of green text “\texttt{Happy Birthday}” on a pink cake}.

\item[(7)] \textbf{Counting + Shapes + Object}: A video of [number1] [shape1] [class1] and [number2] [shape2] [class2].  
\textbf{Example}: \textit{A video of six spherical balls and three rectangular cups}.

\item[(8)] \textbf{Counting + Color + Object}: A video of [number1] [color1] [class1] and [number2] [color2] [class2].  
\textbf{Example}: \textit{A video of six red books and four blue pens}.

\item[(9)] \textbf{View + World Knowledge}: A [view] of [famous landmark].  
\textbf{Example}: \textit{An aerial view of the Great Wall}.

\item[(10)] \textbf{Human + Emotion}: A [human description] [action] with [emotion].  
\textbf{Example}: \textit{A girl in a white blouse and navy skirt, wearing a red ribbon tie, smiles with excitement as she receives a trophy during a school award ceremony. Her long brown hair shines as she turns to the audience}.
    \end{itemize}
    Additionally, some complex tasks are identified through the analysis of real-world video-text pairs, where the text prompts reveal the combination of various simpler tasks. The text of these prompts are then categorized as complex based on their inherent multiattribution.
\end{itemize}

\section{Detailed Information of T2V Models }
\label{modelss}
\begin{table*}
\centering
  \vspace{-7mm}
  \caption{An overview and URLs of the adopted 30 T2V generation models. $\spadesuit$Close-source commercial T2V models. $\heartsuit$Open-source lab T2V models. \textsuperscript{$\dag $}Representative variable and optional.}
  \label{models}
    \renewcommand\arraystretch{1.3}
  \resizebox{\linewidth}{!}{\begin{tabular}{lcccccccl}
    \toprule
     Models&Frames &FPS & Resolution& URL\\
    \midrule 
    $\spadesuit$Pixverse~\cite{pixverse_ai} & 161\textsuperscript{$\dag $} & 30\textsuperscript{$\dag $} &640$\times$360\textsuperscript{$\dag $} & \url{https://pixverse.ai/} \\

    $\spadesuit$Wanxiang~\cite{alibaba2024tongyiwanxiang} & 161\textsuperscript{$\dag $} & 30\textsuperscript{$\dag $} & 1280$\times$720\textsuperscript{$\dag $} & \url{https://tongyi.aliyun.com/wanxiang/} \\

    $\spadesuit$Hailuo~\cite{hailuo} & 141\textsuperscript{$\dag $} & 25\textsuperscript{$\dag $} & 1280$\times$720\textsuperscript{$\dag $} & \url{https://hailuoai.video/} \\
    
    $\spadesuit$Jimeng~\cite{jimeng} & 12\textsuperscript{$\dag $}1 & 24\textsuperscript{$\dag $} & 1472$\times$832\textsuperscript{$\dag $}& \url{https://jimeng.jianying.com/} \\
   
    $\spadesuit$Sora~\cite{videoworldsimulators2024} & 150\textsuperscript{$\dag $} & 30\textsuperscript{$\dag $} & 854$\times$480\textsuperscript{$\dag $} & \url{https://openai.com/research/video-generation-models-as-world-simulators} \\

    $\spadesuit$Hunyuan~\cite{li2024hunyuan} & 129\textsuperscript{$\dag $} & 24\textsuperscript{$\dag $} & 1280$\times$720\textsuperscript{$\dag $} & \url{https://aivideo.hunyuan.tencent.com/} \\

     $\spadesuit$Vidu1.5~\cite{vidu_ai} & 60\textsuperscript{$\dag $} & 16\textsuperscript{$\dag $} & 688$\times$384\textsuperscript{$\dag $} & \url{https://www.vidu.studio/zh} \\
     
     $\spadesuit$Gen3~\cite{runway2024gen3alpha} & 128\textsuperscript{$\dag $} & 24\textsuperscript{$\dag $} & 1280$\times$768\textsuperscript{$\dag $} & \url{https://runwayml.com/research/introducing-gen-3-alpha} \\
     
    $\spadesuit$Kling~\cite{kling} & 153\textsuperscript{$\dag $} & 30\textsuperscript{$\dag $} & 1280$\times$720\textsuperscript{$\dag $} & \url{https://klingai.io/} \\

     $\spadesuit$Genmo~\cite{gemo} & 60\textsuperscript{$\dag $} & 15\textsuperscript{$\dag $} & 1728$\times$1728\textsuperscript{$\dag $} & \url{https://www.genmo.ai} \\
     
     $\spadesuit$ChatGLM~\cite{glm2024chatglm} & 151\textsuperscript{$\dag $} & 30\textsuperscript{$\dag $} & 1280$\times$720\textsuperscript{$\dag $} & \url{https://chatglm.cn/video?lang=zh} \\

    $\spadesuit$Xunfei~\cite{xunfei} & 145\textsuperscript{$\dag $} & 24\textsuperscript{$\dag $} &1024$\times$576\textsuperscript{$\dag $} & \url{https://typemovie.art/} \\

   \hdashline
       
    $\heartsuit$Pyramid~\cite{jin2024pyramidal} & 121 & 24 & 1280$\times$768 & \url{https://github.com/jy0205/Pyramid-Flow} \\

     $\heartsuit$Wan2.1~\cite{wan2025} & 81\textsuperscript{$\dag $} & 16\textsuperscript{$\dag $} & 832$\times$480\textsuperscript{$\dag $} & \url{https://github.com/FoundationVision/LlamaGen} \\
        
    $\heartsuit$Allegro~\cite{allegro2024} & 88 & 15 & 1280$\times$720 & \url{https://github.com/rhymes-ai/Allegro} \\

      $\heartsuit$VideoCrafter2~\cite{chen2024videocrafter2} & 16 & 10 &512$\times$320 & \url{https://github.com/AILab-CVC/VideoCrafter} \\
      
    $\heartsuit$CogVideo X1.5~\cite{yang2024cogvideox} & 32 & 8 & 1360$\times$768 & \url{https://github.com/THUDM/CogVideo} \\
    
    $\heartsuit$Animate~\cite{xu2024easyanimatehighperformancelongvideo}& 49 & 8 & 672$\times$384 & \url{https://github.com/aigc-apps/EasyAnimate} \\

    $\heartsuit$Lavie~\cite{wang2023lavie} & 16 & 8 & 512$\times$320 & \url{https://github.com/Vchitect/LaVie} \\

    $\heartsuit$Hotshot-XL~\cite{Mullan_Hotshot-XL_2023} & 8 & 8 & 672$\times$384 & \url{https://github.com/hotshotco/Hotshot-XL} \\

    $\heartsuit$Latte~\cite{ma2025latte} & 16 & 8 & 512$\times$512 & \url{https://github.com/Vchitect/Latte} \\

     $\heartsuit$VideoCrafter1~\cite{chen2023videocrafter1} & 16\textsuperscript{$\dag $} & 10\textsuperscript{$\dag $}& 512$\times$320\textsuperscript{$\dag $} & \url{https://github.com/AILab-CVC/VideoCrafter} \\

      $\heartsuit$Text2Video-Zero~\cite{text2video-zero} &8  & 4 & 512$\times$512 & \url{https://github.com/Picsart-AI-Research/Text2Video-Zero} \\
      
    $\heartsuit$NOVA~\cite{deng2024nova} & 33 & 12 & 768$\times$480 & \url{https://github.com/baaivision/NOVA} \\
    
    $\heartsuit$ModelScope~\cite{wang2023modelscope} & 16 & 8 & 256$\times$256 & \url{https://github.com/modelscope/modelscope} \\
    
    $\heartsuit$Tune-A-Video~~\cite{wu2023tune} & 8 & 8 & 512$\times$512 & \url{https://github.com/showlab/Tune-A-Video} \\

    $\heartsuit$LTX~\cite{HaCohen2024LTXVideo} & 121 & 25 & 704$\times$480 & \url{https://github.com/Lightricks/LTX-Video} \\

    $\heartsuit$LVDM~\cite{he2022lvdm} & 16 & 8 & 256$\times$256 & \url{https://github.com/YingqingHe/LVDM} \\

    $\heartsuit$ZeroScope~\cite{Zeroscope}  & 36 & 8 & 576$\times$320 & \url{https://huggingface.co/cerspense/zeroscope_v2_XL} \\

    $\heartsuit$LWM~\cite{liu2024world}  & 8 & 4 & 256$\times$256 & \url{https://github.com/LargeWorldModel/LWM} \\

    \bottomrule
  \end{tabular}}
    \vspace{-3mm}
\end{table*}

\noindent 
{\bf Pixverse}~\cite{pixverse_ai} is an addictive all-in-one AI video tool that lets you do everything from easy viral effects, to video-to-video restyling, lip-sync, start-frame/end-frame transitions, and extending videos with AI. Whether you want to send quick, hilarious clips to friends or do more advanced workflows, PixVerse is a strong contender for the all-in-one AI video tool because it's easy for beginners and robust enough for power users.  

\noindent 
{\bf Wanxiang}~\cite{alibaba2024tongyiwanxiang} is a multimodal large model developed by Alibaba DAMO Academy, designed to understand and generate content across text, videos, videos, and audio. As part of Alibaba’s Tongyi AI model family, it enables advanced cross-modal interactions, such as text-to-video generation, visual question answering, and more.

\noindent 
{\bf Hailuo}~\cite{hailuo} is an AI platform that lets you create videos from text or videos. It is developed by the Chinese startup MiniMax, which launched in early September 2023. The platform allows users to create high-quality video content from simple text prompts, making it accessible for various applications such as marketing, education, and entertainment.

\noindent 
{\bf Jimeng}~\cite{jimeng} is a text-to-video model that creates high-quality videos from text or videos instantly, developed by Faceu Technology. It allows users to generate short, realistic video clips from text or video prompts, emphasizing accurate interpretation.

\noindent 
{\bf Sora}~\cite{videoworldsimulators2024} has a deep understanding of language, enabling it to accurately interpret prompts and generate compelling characters that express vibrant emotions. Sora can also create multiple shots within a single generated video that accurately preserve characters and visual style and generate complex scenes with multiple characters, specific types of motion, and accurate details of the subject and background. 

\noindent 
{\bf Hunyuan}~\cite{li2024hunyuan} represents parameter-rich and high-performce text-to-video model currently available in the open-source domain. With 13 billion parameters, it is capable of generating videos that exhibit high physical accuracy and scene consistency, thereby actualizing conceptual visions and fostering creative expression.

 \noindent 
{\bf Vidu1.5}~\cite{vidu_ai} Vidu 1.5 introduces the world's first Multiple-Entity Consistency capability, which seamlessly integrates people, objects, and environments to create stunning video effects. With Vidu's Multiple-Entity Consistency feature, images with no relation to each other, be it characters, objects, or even environments, can be integrated into a single video featuring all three characteristics. Moreover, the resulting video from Vidu 1.5 is capable of ensuring visual consistency even with complex inputs that require the processing of multiple subjects or environments. 

\noindent 
{\bf Gen3}~\cite{runway2024gen3alpha} is the first of the next generation of foundation models trained by Runway on a new infrastructure built for large-scale multimodal training. Trained jointly on videos and videos, Gen-3 Alpha powers Runway's text-to-video, video-to-video and text-to-video tools, existing control modes such as motion brush, advanced camera controls, director mode as well as upcoming tools for more fine-grained control over structure, style, and motion.

\noindent 
{\bf Kling}~\cite{kling} is the new generation of AI creative productivity tools, developed by Kuaishou’s Large Model Algorithm Team, providing a wealth of AI videos, AI videos, and related controllable editing capabilities.

\noindent 
{\bf Genmo}~\cite{gemo} is a creative AI co-pilot developed by Genmo AI, designed for video generation and editing. It enables users to produce animations and videos from text prompts or images, and to restyle existing video clips, offering a versatile suite of tools for artistic expression and iterative creation on an accessible platform.

\noindent 
{\bf ChatGLM}~\cite{glm2024chatglm} is a family of large language models developed by Zhipu AI and Tsinghua University's KEG lab, showcasing an evolution from the foundational GLM-130B to the advanced GLM-4. These models are engineered for strong bilingual (Chinese and English) capabilities in text understanding and generation, with recent iterations like GLM-4 integrating an “All Tools” framework to enhance their capacity for complex task execution and interaction with various external functionalities.

\noindent 
{\bf Xunfei}~\cite{xunfei}, developed by iFlytek, is an AI-powered platform designed for the rapid creation of videos from textual input. It streamlines video production by enabling users to effortlessly generate diverse short video content, offering a range of visual styles and templates to facilitate quick and accessible movie-making.

\noindent 
{\bf Pyramid}~\cite{jin2024pyramidal} is an innovative video generative model, distinguished by its novel “Pyramidal flow matching” technique. This method is designed to significantly enhance the efficiency of video synthesis by progressively matching data distributions across multiple scales, enabling high-quality video generation with potentially reduced computational resources.

\noindent 
{\bf Wan2.1}~\cite{wan2025} is an advanced large-scale video generative model, presented as part of the “Wan” open research. This model is engineered for high-fidelity video synthesis, aiming to push the boundaries of generative capabilities while promoting transparency and accessibility in advanced video AI.

\noindent 
{\bf Allegro}~\cite{allegro2024} is a video generation model,  with the ambitious goal of matching commercial-level quality while aiming to “open the black box” of such advanced systems. Allegro seeks to demystify high-fidelity video synthesis by offering a transparent framework or model that endeavors to replicate sophisticated generative capabilities found in commercial offerings.

\noindent 
{\bf VideoCrafter2}~\cite{chen2024videocrafter2} addresses the challenge of training high-quality video models without high-quality video data. By disentangling motion from appearance at the data level, it uses low-quality videos to maintain motion consistency and high-quality images to ensure visual quality.

\noindent 
{\bf CogVideo X1.5}~\cite{yang2024cogvideox}, part of the CogVideoX series, is a text-to-video diffusion model distinguished by its integration of an “Expert Transformer.” This specialized architecture is designed to enable more efficient scaling and task specialization, aiming for enhanced performance in generating complex and high-fidelity videos from textual descriptions.

\noindent 
{\bf Animate}~\cite{xu2024easyanimatehighperformancelongvideo} is a high-performance method specifically designed for long video generation utilizing a transformer-based architecture. This approach focuses on creating extended and temporally consistent video narratives by harnessing the long-range dependency modeling strengths of transformers to maintain coherence over time.

\noindent 
{\bf Lavie}~\cite{wang2023lavie}  is a video generation model engineered to produce high-quality outputs through the use of “cascaded latent diffusion models.” This multi-stage architecture progressively generates and refines video data in the latent space, enabling the creation of detailed and coherent visual sequences by breaking down the complexity of direct high-resolution synthesis.

\noindent 
{\bf Hotshot-XL}~\cite{Mullan_Hotshot-XL_2023} is a text-to-gif model trained to work alongside Stable Diffusion XL. We adopt its official code with default parameters and change the output format from \texttt{GIF} to \texttt{MP4}.

\noindent 
{\bf Latte}~\cite{ma2025latte} is a latent diffusion Transformer model for video generation. It adopts a video Transformer as the backbone and leverages a pretrained variational autoencoder (VAE) to encode input videos into latent space representations. From these latent features, spatial-temporal tokens are extracted and processed by a series of Transformer blocks.
Latte introduces four effective Transformer-based architectural variants by decomposing the input video's spatial and temporal dimensions.

\noindent 
{\bf VideoCrafter1}~\cite{chen2023videocrafter1} introduces two diffusion models for high-quality video generation: a T2V model for text-to-video and an I2V model for image-to-video. The T2V model enhances SD 2.1 with temporal attention layers for consistent video generation, trained on large datasets.

\noindent 
{\bf NOVA}~\cite{deng2024nova} is a model that enables autoregressive image/video generation with high efficiency. It reformulates the video generation problem as non-quantized autoregressive modeling of temporal frame-by-frame prediction and spatial set-by-set prediction. It generalizes well and enables diverse zero-shot generation abilities in one unified model.

\noindent 
{\bf ModelScope}~\cite{wang2023modelscope} is a decomposed diffusion probabilistic model for video generation. Unlike traditional methods that add independent noise to each frame, it separates noise into shared base noise and residual noise, improving spatial-temporal coherence. This approach leverages pretrained video-generation models for efficient frame content prediction while maintaining motion dynamics. 

\noindent 
{\bf Text2Video-Zero}~\cite{text2video-zero} is a zero-shot text-to-video synthesis model without any further fine-tuning or optimization, which introduces motion dynamics between the latent codes and cross-frame attention mechanism to keep the global scene time consistent.

\noindent 
{\bf Tune-A-Video}~\cite{wu2023tune} is a one-shot text-to-video generation model that extends text-to-video models to the spatio-temporal domain. It uses sparse spatio-temporal attention to maintain consistent objects across frames, overcoming computational limitations. It can synthesize novel videos from a single example compatible with personalized and conditional pretrained T2V models.

\noindent 
{\bf LTX}~\cite{HaCohen2024LTXVideo} is a latent diffusion model engineered for real-time video generation. This system's core innovation lies in its ability to synthesize video sequences at interactive speeds, significantly reducing the typical generation latency associated with diffusion-based video models.

\noindent 
{\bf LVDM}~\cite{he2022lvdm} is an efficient video diffusion model operating in a compressed latent space, designed to address the computational challenges of video synthesis. It uses a hierarchical framework to extend video generation beyond training lengths, effectively mitigating performance degradation via conditional latent perturbation and unconditional guidance techniques. 

\noindent 
{\bf ZeroScope}~\cite{Zeroscope}  is specifically designed for upscaling content made with zeroscope\_v2\_576w using vid2vid in the text2video extension by kabachuha. Leveraging this model as an upscaler allows for superior overall compositions at higher resolutions, permitting faster exploration in 576x320 (or 448x256) before transitioning to a high-resolution render.

\noindent 
{\bf LWM}~\cite{liu2024world} is a multimodal autoregressive model trained on extensive video and language data. Using RingAttention, it efficiently handles long-sequence training, expanding context size up to 1M tokens, enabling strong language, video, and video understanding and generation.

\section{More Details of Subjective Experiment}
\label{exp}
\subsection{Annotation Dimension and Criteria}
To comprehensively assess the performance of AI-generated videos (AIGVs), we propose a dual-dimensional evaluation framework that examines both perceptual quality and text-to-video (T2V) correspondence. This approach enables a thorough analysis of different aspects of video generation, providing a holistic understanding of a model's capabilities and limitations.

\begin{itemize}
\item \textbf{Perceptual quality} evaluates the visual characteristics and aesthetic appeal of generated videos. This dimension focuses on multiple aspects of video quality, including \textbf{visual clarity} (the sharpness and resolution of video details), \textbf{naturalness} (the degree to which the video appears realistic and free from artifacts), \textbf{aesthetic appeal} (the composition, color harmony, and overall visual attractiveness), \textbf{temporal smoothness} (the temporal consistency of the video frames and the smoothness of visual elements transition over time), and \textbf{authenticity} (whether the generated video is realistic). High-scoring videos are characterized by exceptional clarity, vivid and well-balanced colors, and meticulous attention to detail, offering an immersive and visually striking experience. In contrast, low scores reflect videos with blurriness, unnatural color tones, faded visuals, and a lack of clarity or detail. This dimension captures the foundational visual attributes that make a video aesthetically pleasing or distracting. For detailed criteria, refer to Figure \ref{sup_bz1}.

\item \textbf{Text-video correspondence} assesses the semantic alignment between the generated video and the input text prompt, including \textbf{content accuracy} (the presence and correct representation of described objects and elements), \textbf{contextual relevance} (the appropriate depiction of scenes and relationships between objects), \textbf{attribute fidelity} (the accurate representation of specific characteristics mentioned in the prompt), and \textbf{semantic consistency} (the logical coherence between visual elements and textual descriptions). Videos with high scores perfectly match the descriptions in the prompt, accurately reflecting all elements with high fidelity. These videos effectively translate textual information into visual content without mismatches. In contrast, videos with lower scores exhibit inconsistencies, missing elements, or mismatched content. For detailed criteria, refer to Figure \ref{sup_bz2}.
\end{itemize}

\subsection{Significance of the Evaluation Perspectives}
The dual-dimensional evaluation framework, which combines perception quality and T2V correspondence, is essential for addressing the inherent trade-offs and complementary aspects of AIGVs. While perception quality emphasizes the visual characteristics that contribute to a video's appeal and realism, T2V correspondence ensures that the generated content remains semantically faithful to the original textual description. 
 A high perception quality score alone does not guarantee semantic accuracy. For example, a video may exhibit exceptional visual quality, characterized by high resolution, vibrant colors, and meticulous detail, yet fail to accurately represent the specific objects, relationships, or attributes described in the text prompt. Conversely, a video may perfectly align with the textual description in terms of content and context but suffer from poor visual quality, such as low resolution, unnatural textures, or inconsistent lighting, which detracts from its overall appeal and usability.
The integration of both dimensions ensures that generated videos achieve a balance between visual excellence and semantic fidelity. This holistic approach not only enhances the evaluation of generative models but also aligns with real-world applications where both video quality and content accuracy are critical. By considering both dimensions, the framework provides a more nuanced understanding of a model's strengths and weaknesses, facilitating targeted improvements in video generation systems.

While high-level scores (perception quality and T2V correspondence) provide a general overview of model performance, they fall short when it comes to accurately evaluating tasks that require a deeper level of understanding and precision. 
For simple tasks, such as identifying the color “white” in an image while the prompt states “blue”, high-level scores might vary slightly depending on the annotators' interpretation of the content, but for such straightforward tasks, the use of task-specific yes/no questions provides a unified and clear evaluation criterion. A simple yes/no answer is sufficient to confirm whether the model has correctly interpreted the task.
In contrast, for more complex tasks such as “3 red pens in the box”, yes/no answer is not adequate. These tasks require a more detailed and precise assessment, as the model needs to recognize and quantify multiple tasks and their accuracy. A simple yes/no response would fail to capture the depth of understanding needed to evaluate whether the model has correctly identified the number of objects and their characteristics.
For such tasks, a more detailed and nuanced assessment is necessary, which is why high-level scores are employed. Together, high-level scores and task-specific QA pairs allow for a comprehensive evaluation framework that ensures both general alignment and task-specific accuracy, enabling the model’s performance to be measured across a wide range of complexity.

\subsection{Annotation Interface}
\label{uiint}
\begin{figure}[!t]
\vspace{-4mm}
	\centering
	\includegraphics[width=\linewidth]{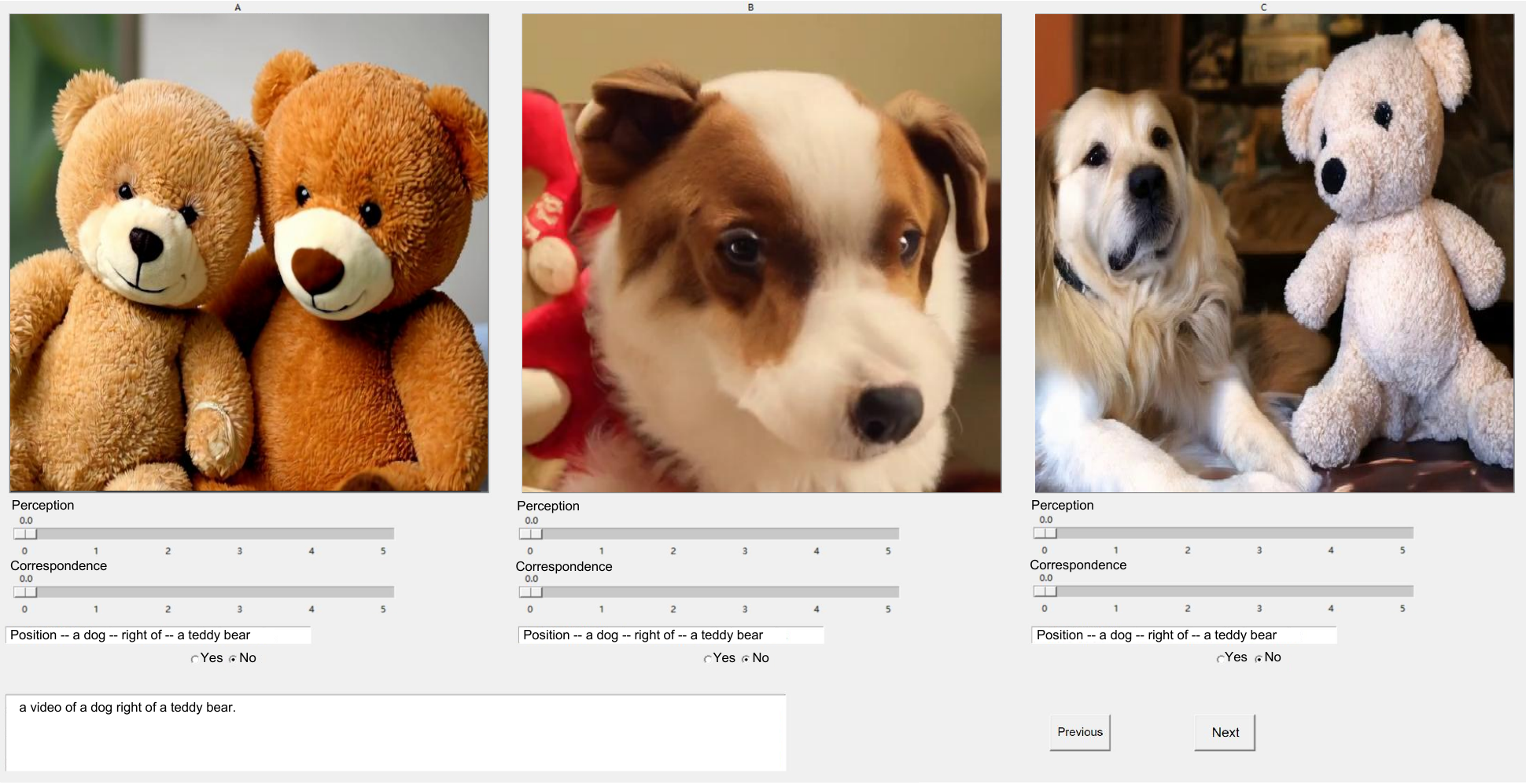}
 \vspace{-5mm}
	\caption{ An example of the simple task annotation interface for human evaluation. The subjects are instructed to rate two dimensions of AI-generated videos, \textit{i.e.}, perception and text-video correspondence, and provide a binary (yes/no) response for a task-specific challenge. Each trial presents three videos generated from 30 models for the same prompt, with absolute scoring applied independently to each video.}
	\label{ui1}
\end{figure}

\begin{figure}[!t]
\vspace{-9mm}
	\centering
	\includegraphics[width=\linewidth]{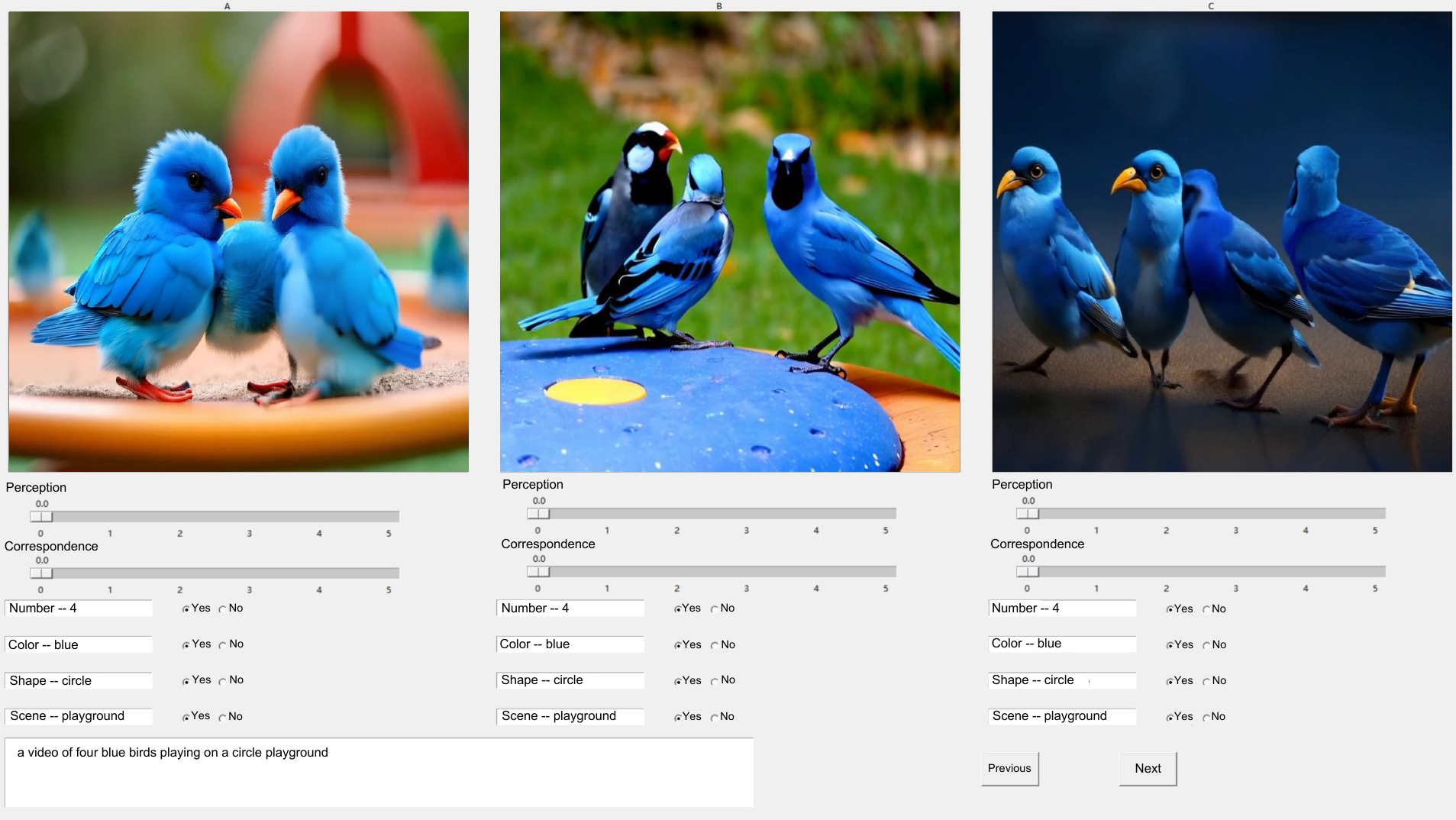}
 \vspace{-5mm}
	\caption{ An example of the complex task annotation interface, which extends the simple task evaluation by incorporating multiple sub-tasks (\textit{e.g.}, Number, Color, Shape, and Scene). The subjects are instructed to rate two dimensions of AI-generated videos, \textit{i.e.}, perception and text-video correspondence, based on the given video and its prompt. Each sub-task is judged independently with a yes/no response. The complex task is considered correct only if all sub-tasks are correctly identified; if any sub-task is incorrect, the entire complex task is marked as incorrect. }
 \vspace{-3mm}
	\label{ui2}
\end{figure}

To ensure a comprehensive and efficient video quality evaluation, we design two custom annotation interfaces tailored for different assessment tasks: simple task annotation and complex task annotation. The simple task annotation interface, shown in Figure \ref{ui1}, is a manual evaluation platform that was developed using the Python tkinter package to facilitate MOS assessments.
The experiment involves evaluating videos based on two independent dimensions and answering a binary question related to a specific task-specific challenge. There are 20 task-specific challenges, including categories such as human, shape, scene, color, etc.
Each trial presents three videos that correspond to the same prompt. These videos are randomly selected from 30 different models. Importantly, participants are instructed to assign absolute scores to each video on the two predefined dimensions, rather than making relative comparisons between the videos.
For each video, participants provide:
(1) Two separate scores representing the two evaluation dimensions.
(2) A binary response (yes/no) to indicate whether the video meets the specified challenge criterion.
 Meanwhile, the complex task annotation interface is illustrated in Figure \ref{ui2}. The complex tasks are composed of multiple subtasks, such as Color, Shape, and Scene. Each subtask is evaluated independently with a yes/no response. The complex task is considered correct only if all its sub-tasks are correct. If any sub-task is incorrect, the entire complex task is marked as incorrect. To ensure uniformity and minimize resolution-related biases in video quality evaluation, all videos displayed in this interface are cropped to a spatial resolution of 1030×1030 pixels.  Navigation options, such as “Previous” and “Next” streamline the workflow, enabling efficient annotation.
\subsection{Annotation Management}
\label{manage}

The annotation process is structured into two primary components: Mean Opinion Score (MOS) annotation and task-specific question-answering (QA) annotation. Each component is designed to evaluate videos across 20 task-specific challenges, including color, position, shapes, view, and \textit{etc.}
The MOS annotation task involves 15 participants to rate each video on a 0-5 Likert scale, assessing both perception quality and T2V correspondence. The question-answering annotation task is similarly conducted with 15 participants, ensuring consistency in the evaluation process. In this task, participants are presented with a series of yes/no questions across the 20 task-specific challenges. To determine the final answer for each question, a majority voting mechanism is employed. This approach ensures that the final decision reflects the collective judgment of the participants, minimizing the impact of individual biases or errors. 

\begin{figure}[!t]
\vspace{-5mm}
	\centering
	\includegraphics[width=1\linewidth]{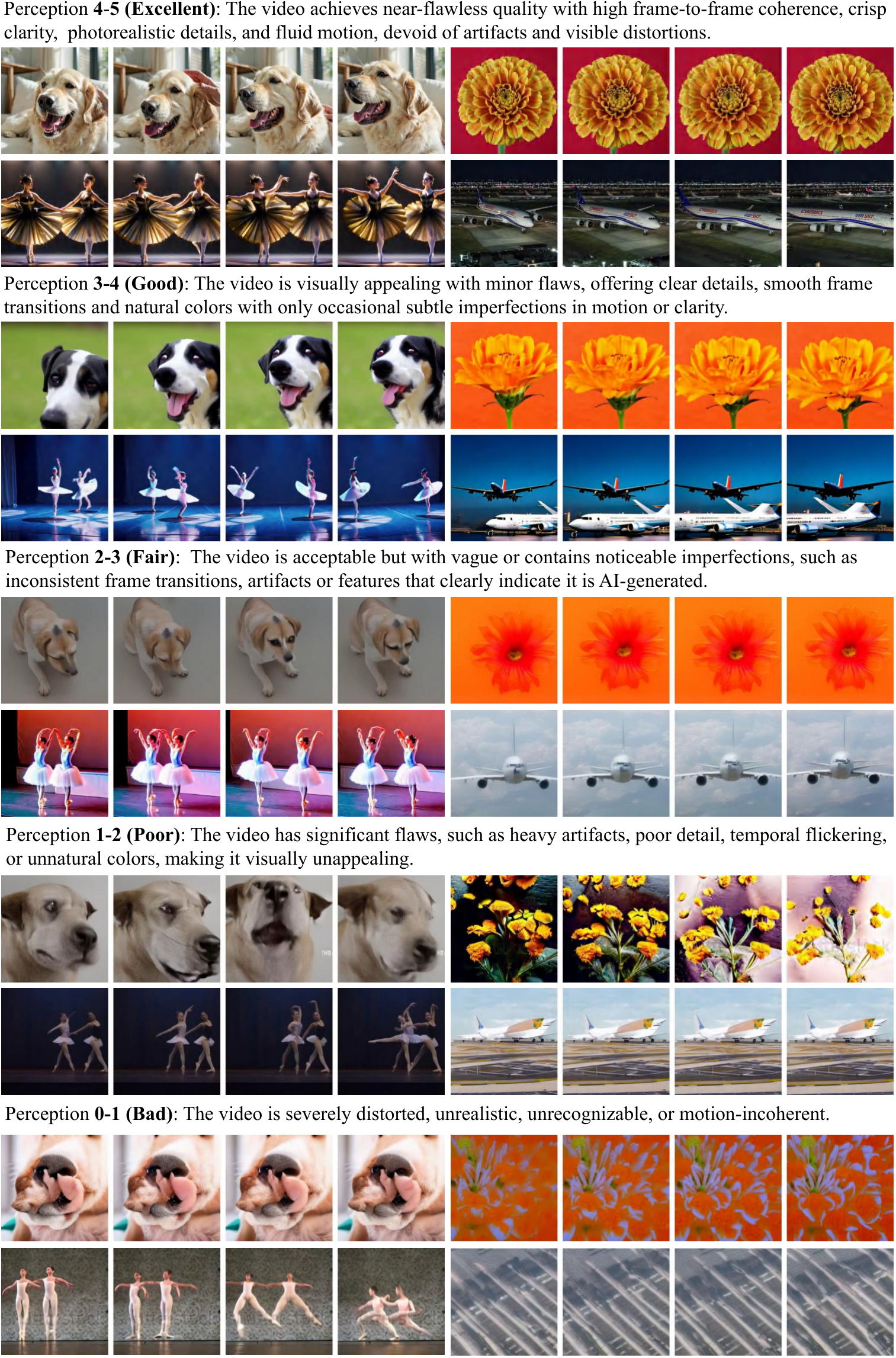}
 \vspace{-2mm}
	\caption{Instructions and examples for manual evaluation of the \textbf{perceptual quality}.}
 \vspace{-3mm}
	\label{sup_bz1}
\end{figure}

\begin{figure}[!t]
\vspace{-5mm}
	\centering
	\includegraphics[width=1\linewidth]{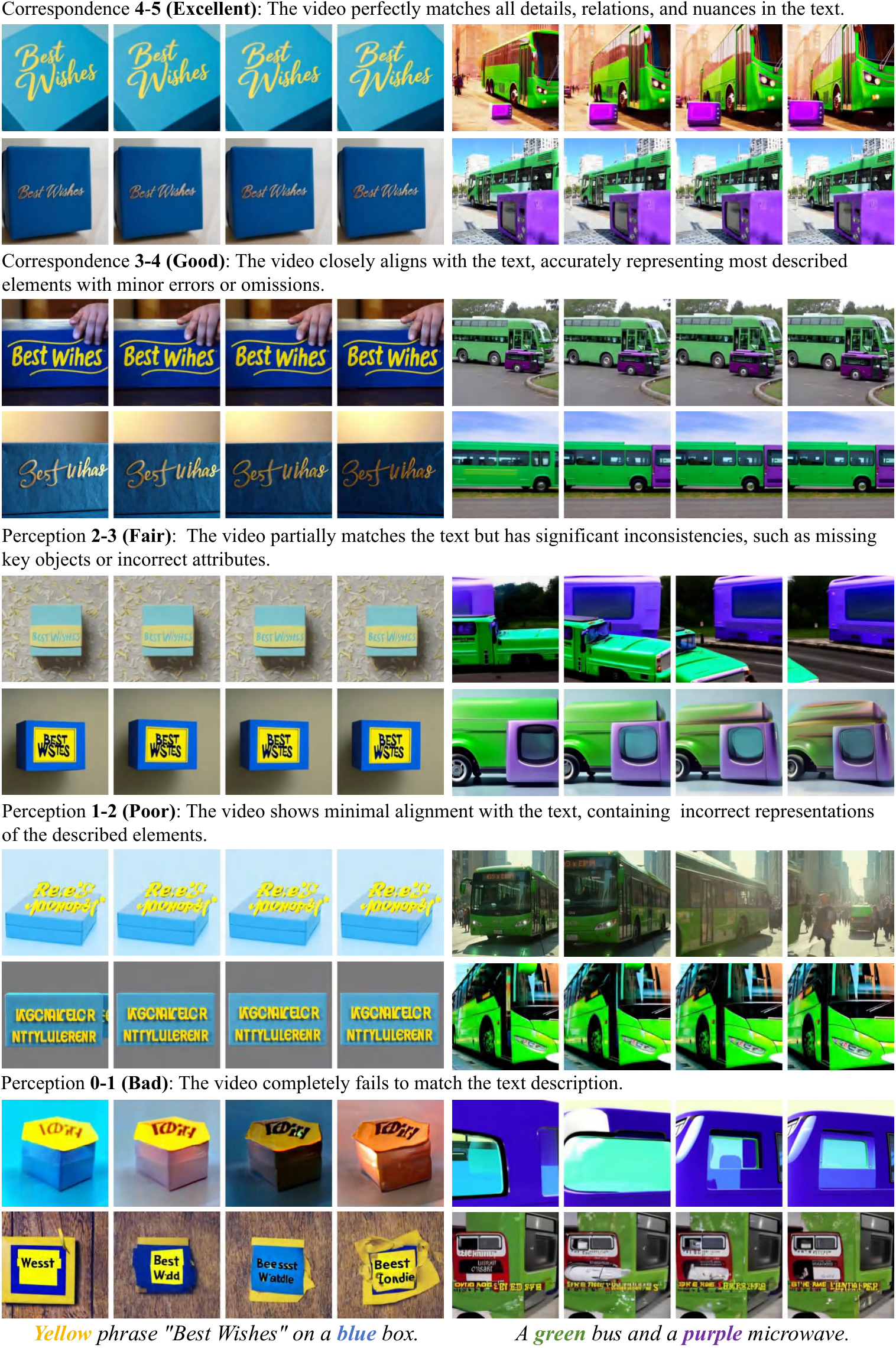}
 \vspace{-1mm}
	\caption{Instructions and examples for manual evaluation of \textbf{T2V correspondence}. Prompt (left): yellow phrase ``Best Wishes” on a blue box. Prompt (right): a green bus and a purple microwave.}
 \vspace{-3mm}
	\label{sup_bz2}
\end{figure}

Prior to engaging in the annotation tasks, all participants undergo a rigorous training process. As illustrated in Figures \ref{sup_bz1}-\ref{sup_bz2}, they are provided with detailed instructions and multiple standardized examples. To ensure a high level of understanding and consistency, a pre-test is conducted to evaluate participants' comprehension of the criteria and their alignment with the standard examples. Participants who do not meet the required accuracy threshold are excluded from further participation, ensuring that only well-prepared individuals contribute to the final dataset.
During the experiment, all evaluations are conducted in a controlled laboratory environment under normal indoor lighting conditions. Participants are seated at a comfortable viewing distance of approximately 60 cm from the screen to minimize visual strain and ensure consistent evaluation conditions. 
While individual preferences may naturally vary, the use of detailed explanations and standardized annotation criteria ensures a high degree of agreement among participants. This consensus is particularly evident in question-answering annotations, where majority voting effectively captures group preferences.  

To ensure the reliability of subjective ratings, we follow the ITU-R BT.500 \cite{series2012methodology} and apply a two-step outlier rejection procedure based on statistical properties of the collected scores. $S_{i,j}$ denote the raw score given by subject $j$ to item $i$, and let $\mu_i$, $\sigma_i$ be the mean and standard deviation across raters for item $i$.
\paragraph{1. Subject-level Outlier Rejection.}
We compute the kurtosis $\beta_i$ of the score distribution for each item $i$, and use it to determine a dynamic threshold coefficient $k_i$:
\[
k_i = 
\begin{cases}
2, & \text{if } 2 \leq \beta_i \leq 4 \\
\sqrt{20}, & \text{otherwise}
\end{cases}
\]

For each subject $j$, we count how many times their score deviates from the mean by more than $k_i\sigma_i$:
\[
P_j = \left| \left\{ i \mid S_{i,j} \geq \mu_i + k_i \sigma_i \right\} \right|, \quad
Q_j = \left| \left\{ i \mid S_{i,j} \leq \mu_i - k_i \sigma_i \right\} \right|
\]

Subject $j$ is considered an outlier and rejected if:
\[
\frac{P_j + Q_j}{N} > 0.05 \quad \text{and} \quad \left| \frac{P_j - Q_j}{P_j + Q_j} \right| < 0.3
\]
where $N$ is the total number of items.

\paragraph{2. Score-level Outlier Rejection.}
For each accepted subject, their score $S_{i,j}$ is retained only if it falls within the interval:
\[
\mu_i - k_i \sigma_i \leq S_{i,j} \leq \mu_i + k_i \sigma_i
\]
Otherwise, the score is excluded from the MOS calculation.

With a subject rejection rate of 3\% and a per-score rejection rate of approximately 2\%, this rigorous and ethically sound annotation management strategy establishes AIGVE-60K as a robust and reliable resource for advancing research in video quality assessment.

\section{More Analysis of AIGVE-60K Database}
\label{analyze}
\subsection{MOS Distribution across 20 Challenges}
\begin{figure}[!t]
\vspace{-7mm}
	\centering
	\includegraphics[width=1\linewidth]{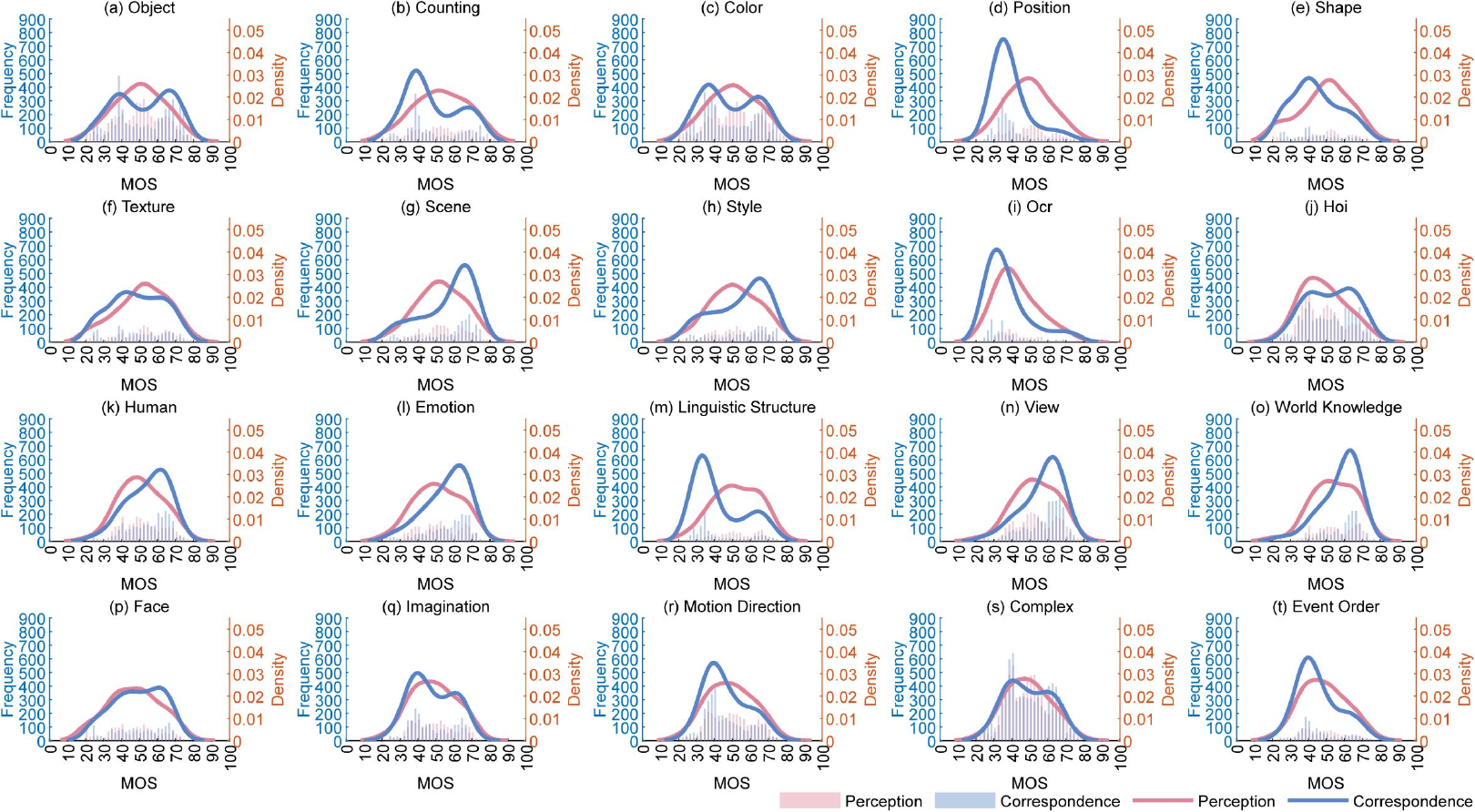}
 \vspace{-4mm}
	\caption{\textbf{Mean Opinion Score (MOS) distribution histograms and kernel density curves} of AIGVE-60K dataset. It includes two dimensions: perception MOS and correspondence MOS. Each dimension contains a total of 58,500 MOS values.}
 \vspace{-3mm}
	\label{sup_20task1}
\end{figure}
As mentioned in the main text, we process and compute the valid subjective evaluation results, obtaining a total of 117,000 Mean Opinion Scores (MOSs) across two dimensions, along with QA accuracy. To better illustrate the generative capabilities of current T2V models in different prompt challenges, we categorize the computed MOSs data into 20 task categories and used the categorized data to plot histograms and kernel density curves (KDC) graphs, as shown in Figure \ref{sup_20task1}.
We can observe that the 30 T2V models we tested exhibit relatively poor text-video alignment in prompt challenges related to position, OCR, linguistic structures, and complexity, with MOSs primarily clustering around 30. In contrast, their performance in other prompt challenges is relatively better. The overall perception MOSs does not show significant differences across different prompt challenges, with scores generally concentrated at a higher level. However, models perform slightly worse in OCR, HOI, and Face-related prompt challenges, where lower MOSs appear more frequently compared to other prompt challenges.
\begin{table*}[tbph]
\vspace{-6mm}
\centering
\renewcommand\arraystretch{1.1}
\caption{Performance comparisons of T2V Models on human-annotated \textbf{perception} MOS.
}
   \resizebox{\linewidth}{!}{\begin{tabular}{l||cccccccccccccccccccc>{\columncolor{mycolor_green}}c:>{\columncolor{mycolor_blue}}c}
  \Xhline{1px}
 Models
&$\text{Object}$&$\text{Color}$&$\text{Count.}$&$\text{Texture}$&$\text{Position}$ &$\text{HOI}$&$\text{Face}$&$\text{Emo.}$&$\text{Human}$&$\text{OCR}$&$\text{Scene}$&$\text{Style}$&$\text{Shape}$&$\text{View}$&$\text{Know.}$&$\text{Ling.}$&$\text{Imag.}$&$\text{Motion}$&$\text{Event}$&$\text{Complex}$& $\text{Overall}$ &Rank\\
    \hline
    Jimeng~\cite{jimeng}&68.03&66.51&65.88&64.83&64.60&66.00&63.93&65.35&67.01&59.16&66.38&65.05&63.56&66.96&63.86&63.21&65.91&62.82&59.13&68.75&65.25&1\\
    Pixverse~\cite{pixverse_ai}&63.97&66.42&64.76&61.30&64.70&65.08&65.25&66.14&68.17&64.26&63.01&64.75&59.34&65.55&61.82&63.17&64.55&60.65&55.64&64.47&63.81&2\\
    Sora~\cite{videoworldsimulators2024}&64.10&61.79&61.22&63.04&62.66&61.97&57.73&62.77&63.82&64.31&62.01&63.13&62.29&62.20&62.71&63.90&62.62&59.00&57.96&62.89&62.09&3\\
    Hailuo~\cite{hailuo}&61.70&60.97&60.97&61.21&60.29&59.36&60.24&65.87&59.80&61.55&59.50&60.75&61.05&60.44&62.30&61.67&61.09&61.38&51.44&61.74&60.58&4\\
    Kling~\cite{kling}&60.70&62.07&62.39&60.01&58.46&61.11&59.45&63.99&64.34&48.21&62.04&61.43&60.69&58.90&60.93&57.17&63.58&58.58&58.40&63.01&60.56&5\\
    Wanxiang~\cite{alibaba2024tongyiwanxiang}&60.60&61.47&60.84&58.70&64.12&59.09&61.11&62.96&62.24&61.37&59.78&59.81&61.54&62.58&60.06&63.68&59.04&58.72&58.02&60.41&60.54&6\\
    Gen3~\cite{runway2024gen3alpha}&58.22&62.37&60.26&55.98&57.42&58.36&61.61&65.00&56.84&61.68&59.16&57.86&53.78&59.41&58.34&58.71&59.21&58.97&56.12&62.53&59.22&7\\
    Hunyuan~\cite{li2024hunyuan}&60.85&60.00&60.83&57.15&60.00&60.18&52.95&60.84&59.23&58.86&57.75&59.00&58.00&60.75&59.36&58.99&59.07&56.93&52.98&59.61&58.81&8\\
    Xunfei~\cite{xunfei}&62.25&60.99&60.26&60.80&60.48&56.70&59.62&58.35&54.80&61.02&58.48&58.59&58.36&60.69&58.43&61.88&58.31&56.96&55.29&52.88&58.60&9\\
    Genmo~\cite{gemo}&62.50&61.47&57.67&49.82&54.50&62.26&52.19&65.16&60.37&49.55&55.87&52.62&52.03&59.30&56.67&61.15&60.82&51.71&51.69&62.07&57.66&10\\
    Pyramid~\cite{jin2024pyramidal}&63.18&55.51&59.91&58.01&58.62&57.28&44.19&60.34&61.49&40.61&58.55&55.68&56.89&59.98&62.14&56.12&62.11&57.19&50.05&61.46&57.45&11\\
    Wan2.1~\cite{wan2025}&59.07&57.94&57.90&58.48&54.69&55.93&62.64&62.06&57.53&51.78&55.80&54.85&56.04&59.05&57.69&55.74&57.25&57.08&53.18&60.55&57.27&12\\
    ChatGLM~\cite{glm2024chatglm}&55.19&55.95&57.38&58.06&59.38&58.84&61.51&61.20&57.84&49.28&55.51&56.52&54.20&56.26&56.54&60.66&53.43&53.16&53.22&58.00&56.39&13\\

    Allegro~\cite{allegro2024}&60.43&54.95&54.82&55.32&52.69&60.28&50.11&62.52&61.63&40.30&58.22&51.73&52.17&61.43&58.75&50.78&61.53&52.54&49.27&55.36&56.08&14\\

    Vidu1.5~\cite{vidu_ai}&56.09&55.93&53.52&57.93&56.07&51.97&59.85&53.51&53.29&55.28&51.32&52.42&56.23&55.83&50.58&55.71&56.74&50.59&51.55&58.78&54.56&15\\

    CogVideo X1.5~\cite{yang2024cogvideox}&54.63&56.76&53.80&48.81&45.18&53.31&43.78&53.96&48.30&38.60&51.09&42.75&47.11&54.83&59.13&48.34&53.75&47.53&44.12&49.81&50.59&16\\
    
    Animate~\cite{xu2024easyanimatehighperformancelongvideo}&53.41&53.18&54.70&44.71&36.88&47.49&52.32&54.40&49.71&38.74&53.62&50.19&49.58&56.78&58.58&47.18&53.68&47.51&45.37&51.73&50.49&17\\

    Lavie~\cite{wang2023lavie}&54.33&52.79&49.66&48.20&42.82&47.63&46.07&52.62&51.21&46.48&48.44&46.87&46.83&53.18&48.05&50.79&50.75&47.54&45.19&50.90&49.30&18\\

    VideoCrafter2~\cite{chen2024videocrafter2}&48.40&50.89&46.50&51.81&44.09&47.06&50.27&46.31&51.24&45.92&49.16&52.89&45.08&48.61&45.25&51.45&47.22&48.99&43.02&47.05&48.11&19\\

    VideoCrafter1~\cite{chen2024videocrafter2}&49.37&51.66&40.35&37.56&35.79&48.27&35.67&47.48&47.49&40.74&44.00&41.85&40.03&47.66&41.99&36.54&48.26&39.92&36.14&47.65&44.12&20\\

    Latte~\cite{ma2025latte}&48.10&47.04&41.26&45.45&40.26&39.25&42.10&45.20&40.69&39.56&43.45&43.47&41.04&43.91&48.58&48.62&44.81&44.98&41.10&46.68&43.81&21\\

    Hotshot-XL~\cite{Mullan_Hotshot-XL_2023}&45.19&46.73&40.55&35.37&33.92&42.75&40.27&42.37&41.59&50.01&42.78&45.72&41.01&48.17&37.79&39.45&43.61&42.65&36.38&44.20&42.66&22\\

    NOVA~\cite{deng2024nova}&43.02&47.73&40.26&43.05&43.30&36.30&38.75&33.44&33.45&36.50&42.63&39.84&44.19&48.40&42.23&43.37&38.38&40.16&42.23&42.92&41.18&23\\

    Text2Video-Zero~\cite{text2video-zero}&42.28&41.29&41.74&39.40&32.11&39.89&44.01&43.35&44.34&32.78&41.22&40.60&37.31&42.24&41.92&37.43&44.95&39.45&36.96&40.33&40.53&24\\

    LTX~\cite{HaCohen2024LTXVideo}&41.52&46.48&36.98&35.44&35.72&39.00&33.22&42.84&43.33&32.43&40.36&33.39&32.35&50.32&49.53&32.17&43.28&37.29&36.66&43.88&40.11&25\\

    ModelScope~\cite{wang2023modelscope}&43.82&42.94&37.64&36.89&37.25&38.97&33.68&39.16&34.97&36.79&36.41&39.56&34.09&39.73&34.50&34.53&37.54&35.96&36.48&40.78&38.00&26\\

    Tune-A-Video~\cite{wu2023tune} &36.29&36.48&35.55&33.75&30.55&35.67&36.12&38.28&39.76&33.97&35.64&36.78&32.48&34.33&40.21&33.00&37.90&32.83&32.44&34.36&35.41&27\\
    
    LVDM~\cite{he2022lvdm}&37.95&37.35&32.14&32.57&29.72&32.46&28.48&29.29&32.44&34.78&33.78&33.03&34.25&38.69&31.92&34.27&33.43&33.89&32.76&35.58&33.84&28\\

    ZeroScope~\cite{Zeroscope}&34.34&32.47&26.77&23.01&27.34&34.06&21.34&32.53&30.51&33.83&28.50&31.49&19.89&29.44&32.06&28.55&30.40&30.46&30.91&32.76&30.08&29\\
    LWM~\cite{liu2024world}&29.18&26.76&24.70&26.24&23.70&27.78&29.80&27.04&25.51&28.79&27.82&24.45&24.52&31.29&27.46&29.76&25.80&28.10&28.45&26.85&27.39&30\\

    \Xhline{1px}
  \end{tabular}}\label{mos1}

 \vspace{-2mm}
\end{table*}
\begin{table*}[tbph]
\vspace{-2mm}
\centering
\renewcommand\arraystretch{1.1}
\caption{Performance comparisons of T2V Models on human-annotated \textbf{correspondence} MOS.
}
   \resizebox{\linewidth}{!}{\begin{tabular}{l||cccccccccccccccccccc>{\columncolor{mycolor_green}}c:>{\columncolor{mycolor_blue}}c}
  \Xhline{1px}

 Models
&$\text{Object}$&$\text{Color}$&$\text{Count.}$&$\text{Texture}$&$\text{Position}$ &$\text{HOI}$&$\text{Face}$&$\text{Emo.}$&$\text{Human}$&$\text{OCR}$&$\text{Scene}$&$\text{Style}$&$\text{Shape}$&$\text{View}$&$\text{Know.}$&$\text{Ling.}$&$\text{Imag.}$&$\text{Motion}$&$\text{Event}$&$\text{Complex}$& $\text{Overall}$ &Rank\\
    \hline
    Wanxiang~\cite{alibaba2024tongyiwanxiang}&64.11&61.69&60.87&59.28&59.94&61.61&57.60&61.67&65.02&54.63&61.83&59.40&63.87&62.94&62.10&62.94&59.79&54.39&50.01&60.16&60.37&1\\
    Pixverse~\cite{pixverse_ai}&62.86&61.25&62.56&63.05&63.57&60.72&60.57&55.99&59.56&66.51&59.72&60.01&57.82&58.53&62.39&59.98&60.14&55.69&48.11&59.29&59.97&2\\

    Hailuo~\cite{hailuo}&62.76&61.45&53.52&61.32&54.19&59.74&58.71&65.39&61.67&58.57&61.16&60.19&60.54&60.07&63.55&61.39&62.83&55.49&45.12&60.29&59.74&3\\
    Sora~\cite{videoworldsimulators2024}&63.12&59.98&56.67&62.97&52.07&60.29&55.57&62.15&64.17&63.83&61.50&60.26&56.38&62.72&60.60&65.81&62.48&52.76&46.76&57.84&59.68&4\\
    
    Vidu1.5~\cite{vidu_ai}&62.67&61.91&53.22&59.30&54.34&59.00&59.88&57.97&59.15&47.33&58.42&60.87&56.86&59.48&60.04&56.40&60.30&55.39&51.53&60.42&58.25&5\\
    
    Jimeng~\cite{jimeng}&61.53&55.74&55.40&62.48&53.58&60.45&55.72&61.37&61.67&41.04&61.06&50.89&61.42&61.69&58.37&61.69&59.82&50.18&47.01&63.53&57.86&6\\
    
    Hunyuan~\cite{li2024hunyuan}&59.78&58.94&58.94&58.14&57.95&60.32&52.23&61.20&58.17&51.02&60.03&55.76&56.94&59.33&61.76&43.51&58.69&52.35&46.01&59.98&57.25&7\\
    
    Gen3~\cite{runway2024gen3alpha}&54.94&57.85&45.91&57.07&49.53&56.11&58.16&61.09&53.52&58.67&60.25&60.46&48.55&56.44&56.39&58.33&59.43&51.11&38.83&58.47&55.72&8\\

    Kling~\cite{kling}&60.66&58.57&56.60&54.61&51.39&57.54&52.57&59.03&59.83&32.76&58.24&57.66&56.25&58.41&55.87&51.99&57.55&53.92&44.89&58.07&55.57&9\\
    
    ChatGLM~\cite{glm2024chatglm}&51.82&51.31&48.93&59.83&56.13&55.44&55.17&62.07&46.96&37.21&58.13&58.83&54.10&56.05&60.79&60.22&55.93&47.94&45.96&54.32&53.98&10\\

    Genmo~\cite{gemo}&58.36&56.15&42.75&51.83&57.73&53.06&52.21&60.03&52.38&43.20&55.38&57.11&51.15&54.68&57.24&59.43&55.34&51.93&43.60&55.60&53.78&11\\

    Xunfei~\cite{xunfei}&60.42&55.37&57.22&57.88&52.09&54.35&54.43&55.90&52.42&53.73&52.94&53.20&51.41&56.72&59.02&59.47&53.12&45.37&38.33&47.23&53.46&12\\

    Wan2.1~\cite{wan2025}&57.86&54.74&48.39&54.50&42.45&55.43&59.01&64.19&60.22&42.40&51.73&47.77&42.02&58.51&57.45&40.63&54.17&46.33&39.43&55.03&52.33&13\\

    VideoCrafter2~\cite{chen2024videocrafter2}&53.29&51.28&45.34&54.77&41.20&52.88&52.08&54.18&57.15&40.26&54.19&58.10&51.08&56.39&53.94&50.25&50.05&45.20&39.26&52.45&51.07&14\\

    Allegro~\cite{allegro2024}&52.79&50.36&46.87&49.98&39.82&53.12&53.04&54.19&51.08&32.03&53.68&46.06&45.62&58.72&58.41&40.39&57.05&50.19&41.66&54.12&50.68&15\\

    Pyramid~\cite{jin2024pyramidal}&56.07&46.65&53.95&53.22&45.62&53.74&48.07&56.80&54.23&31.58&52.87&51.28&51.35&53.93&54.11&40.52&50.72&44.05&38.25&54.84&50.17&16\\

    CogVideo~\cite{yang2024cogvideox}&53.85&50.62&52.66&48.11&41.36&55.25&48.95&55.62&49.81&35.12&50.49&47.68&44.99&52.92&60.80&41.26&54.39&43.37&40.31&49.84&49.73&17\\
    Animate~\cite{xu2024easyanimatehighperformancelongvideo}&51.73&53.69&54.71&50.15&31.35&49.75&52.52&53.45&48.80&34.16&52.99&55.52&47.54&54.43&58.86&37.28&51.41&44.40&40.26&49.03&49.30&18\\

    Lavie~\cite{wang2023lavie}&51.67&45.55&56.42&51.23&33.44&49.08&50.01&51.64&49.86&39.97&47.92&48.77&51.93&52.13&52.23&47.33&50.57&42.31&38.62&50.07&48.22&19\\

    Hotshot-XL~\cite{Mullan_Hotshot-XL_2023}&52.72&49.17&46.90&41.35&32.02&49.99&44.89&48.26&49.14&52.34&49.08&54.94&47.23&52.36&49.52&38.66&50.16&41.62&37.68&50.67&47.75&20\\
    NOVA~\cite{deng2024nova}&52.23&51.77&48.03&50.54&40.35&45.14&45.10&45.48&43.94&33.84&50.91&50.35&46.09&55.43&53.20&38.06&46.33&41.32&37.32&53.45&47.18&21\\

    Latte~\cite{ma2025latte}&52.71&48.64&39.74&47.05&34.00&46.20&50.28&52.48&48.95&32.75&48.53&52.10&44.60&46.35&52.14&47.24&49.29&43.85&36.63&50.76&46.73&22\\

    Text2Video-Zero~\cite{text2video-zero}&48.01&46.01&51.74&50.66&29.73&44.75&48.70&41.89&42.77&32.41&45.48&51.36&50.97&48.41&47.39&46.89&44.64&41.21&34.79&45.84&44.89&23\\

    VideoCrafter1~\cite{chen2024videocrafter2}&50.34&48.01&41.58&44.32&36.07&45.78&38.29&43.64&47.07&36.98&46.76&54.15&46.31&46.91&46.15&40.83&46.52&39.80&36.69&45.32&44.67&24\\

    ModelScope~\cite{wang2023modelscope}&46.00&46.29&40.27&45.07&40.26&46.29&37.63&46.23&45.97&41.93&41.65&53.65&43.38&41.21&46.84&41.24&44.67&40.09&37.06&46.05&43.73&25\\

    Tune-A-Video~\cite{wu2023tune} &49.46&44.66&41.35&45.05&30.22&46.58&48.32&42.47&47.93&35.55&40.33&50.30&42.65&42.82&47.71&41.67&40.74&36.98&33.35&42.05&42.69&26\\
    LVDM~\cite{he2022lvdm}&49.90&43.44&40.96&43.66&32.91&44.49&33.45&42.63&39.83&34.30&41.65&44.66&42.43&47.83&43.24&38.82&44.22&38.41&35.08&46.48&42.20&27\\
    
    LTX~\cite{HaCohen2024LTXVideo}&45.02&42.95&38.98&41.45&31.59&44.57&36.90&44.92&47.19&28.34&43.40&39.93&31.76&50.90&43.98&32.21&42.48&40.73&33.93&43.67&41.28&28\\

    ZeroScope~\cite{Zeroscope}&40.75&36.60&31.47&24.04&26.60&40.65&24.88&44.56&37.98&30.92&34.57&39.91&22.47&34.46&39.52&32.29&34.99&32.23&31.76&38.60&34.69&29\\
    LWM~\cite{liu2024world}&35.40&31.53&28.42&27.16&25.42&33.06&34.24&32.83&31.42&27.82&34.15&25.71&25.11&35.92&31.44&33.71&30.84&31.31&30.21&31.84&31.49&30\\

    \Xhline{1px}
  \end{tabular}}\label{mos2}

 \vspace{-2mm}
\end{table*}
\begin{table*}[tbph]
\vspace{-2mm}
\centering
\renewcommand\arraystretch{1.1}
\caption{Performance comparisons of T2V Models on human-annotated \textbf{task-specific accuracy}.
}
   \resizebox{\linewidth}{!}{\begin{tabular}{l||cccccccccccccccccccc>{\columncolor{mycolor_green}}c:>{\columncolor{mycolor_blue}}c}
  \Xhline{1px}
   Models
&$\text{Object}$&$\text{Color}$&$\text{Count.}$&$\text{Texture}$&$\text{Position}$ &$\text{HOI}$&$\text{Face}$&$\text{Emo.}$&$\text{Human}$&$\text{OCR}$&$\text{Scene}$&$\text{Style}$&$\text{Shape}$&$\text{View}$&$\text{Know.}$&$\text{Ling.}$&$\text{Imag.}$&$\text{Motion}$&$\text{Event}$&$\text{Complex}$& $\text{Overall}$ &Rank\\
    \hline
    Pixverse~\cite{pixverse_ai}&100.0&100.0&100.0&100.0&90.00&96.67&91.67&55.56&100.0&100.0&92.59&80.00&72.73&94.44&100.0&80.00&100.0&82.35&54.55&94.74&91.33&1\\
    Wanxiang~\cite{alibaba2024tongyiwanxiang}&100.0&95.00&80.00&100.0&80.00&100.0&91.67&100.0&90.00&75.00&96.30&90.00&100.0&100.0&100.0&90.00&87.50&64.71&54.55&89.47&90.33&2\\
    Hailuo~\cite{hailuo}&88.89&90.00&70.00&100.0&60.00&86.67&100.0&100.0&90.00&83.33&88.89&90.00&81.82&88.89&100.0&90.00&100.0&82.35&45.45&94.74&87.67&3\\
    Vidu1.5~\cite{vidu_ai}&94.44&95.00&70.00&90.91&60.00&93.33&91.67&88.89&100.0&50.00&92.59&100.0&81.82&88.89&100.0&80.00&87.50&88.24&54.55&94.74&87.00&4\\
    Sora~\cite{videoworldsimulators2024}&88.89&95.00&70.00&100.0&50.00&90.00&83.33&100.0&90.00&91.67&96.30&80.00&63.64&100.0&90.91&100.0&95.83&52.94&45.45&89.47&85.67&5\\
    Jimeng~\cite{jimeng}&94.44&85.00&70.00&90.91&60.00&90.00&75.00&100.0&100.0&33.33&92.59&50.00&81.82&88.89&81.82&90.00&95.83&47.06&45.45&100.0&81.33&6\\
    
    Hunyuan~\cite{li2024hunyuan}&83.33&85.00&90.00&81.82&90.00&86.67&66.67&88.89&80.00&50.00&96.30&60.00&81.82&83.33&100.0&30.00&87.50&70.59&36.36&89.47&79.67&7\\
    
    Genmo~\cite{gemo}&83.33&90.00&30.00&63.64&80.00&73.33&91.67&100.0&60.00&25.00&92.59&80.00&54.55&88.89&90.91&80.00&87.50&76.47&18.18&84.21&75.67&8\\
    
    Gen3~\cite{runway2024gen3alpha}&77.78&80.00&40.00&90.91&40.00&76.67&75.00&100.0&70.00&75.00&92.59&90.00&63.64&77.78&81.82&90.00&83.33&52.94&18.18&89.47&75.33&9\\
    
    ChatGLM~\cite{glm2024chatglm}&66.67&75.00&60.00&90.91&60.00&73.33&83.33&100.0&60.00&16.67&96.30&80.00&63.64&94.44&100.0&90.00&87.50&47.06&18.18&78.95&74.00&10\\
    Kling~\cite{kling}&94.44&85.00&60.00&72.73&50.00&83.33&66.67&77.78&80.00&8.33&85.19&90.00&81.82&94.44&63.64&60.00&79.17&64.71&27.27&78.95&73.67&11\\
        
    Xunfei~\cite{xunfei}&88.89&65.00&90.00&81.82&50.00&73.33&75.00&77.78&70.00&66.67&62.96&60.00&63.64&66.67&90.91&80.00&70.83&35.29&0.00&57.89&66.33&12\\
    
    VideoCrafter2~\cite{chen2024videocrafter2}&66.67&55.00&50.00&72.73&30.00&76.67&66.67&77.78&90.00&16.67&88.89&90.00&63.64&88.89&81.82&60.00&62.50&41.18&9.09&78.95&65.67&13\\
    
    CogVideo X1.5~\cite{yang2024cogvideox}&77.78&65.00&60.00&54.55&20.00&86.67&75.00&88.89&70.00&16.67&74.07&60.00&45.45&77.78&100.0&30.00&83.33&35.29&27.27&68.42&64.67&14\\
    
    Pyramid~\cite{jin2024pyramidal}&77.78&45.00&70.00&72.73&30.00&83.33&58.33&100.0&80.00&8.33&77.78&60.00&63.64&72.22&81.82&30.00&75.00&29.41&18.18&84.21&63.67&15\\
    
    Allegro~\cite{allegro2024}&72.22&75.00&50.00&63.64&20.00&83.33&58.33&66.67&70.00&8.33&70.37&40.00&36.36&88.89&72.73&20.00&87.50&52.94&18.18&84.21&63.00&16\\
    
    Wan2.1~\cite{wan2025}&83.33&65.00&40.00&72.73&40.00&73.33&91.67&88.89&80.00&25.00&77.78&50.00&27.27&83.33&72.73&30.00&70.83&41.18&0.00&68.42&62.67&17\\
    
    Animate~\cite{xu2024easyanimatehighperformancelongvideo}&66.67&70.00&80.00&72.73&10.00&56.67&75.00&77.78&80.00&8.33&70.37&90.00&54.55&83.33&90.91&10.00&70.83&29.41&27.27&63.16&60.67&18\\
    
    Hotshot-XL~\cite{Mullan_Hotshot-XL_2023}&66.67&65.00&40.00&36.36&0.00&70.00&50.00&77.78&70.00&75.00&62.96&80.00&63.64&72.22&72.73&20.00&62.50&35.29&9.09&68.42&57.67&19\\

    NOVA~\cite{deng2024nova}&72.22&70.00&50.00&63.64&20.00&56.67&58.33&55.56&50.00&8.33&81.48&80.00&54.55&77.78&72.73&10.00&54.17&23.53&0.00&84.21&56.00&20\\
    Lavie~\cite{wang2023lavie}&55.56&40.00&80.00&54.55&10.00&70.00&66.67&66.67&90.00&16.67&55.56&60.00&63.64&72.22&81.82&50.00&66.67&29.41&9.09&47.37&55.00&21\\
    Latte~\cite{ma2025latte}&61.11&65.00&20.00&54.55&10.00&60.00&66.67&77.78&70.00&0.00&70.37&60.00&54.55&44.44&81.82&50.00&58.33&47.06&0.00&78.95&54.33&22\\

    Text2Video-Zero~\cite{text2video-zero}&66.67&75.00&80.00&54.55&0.00&56.67&58.33&33.33&50.00&0.00&59.26&60.00&54.55&66.67&45.45&50.00&45.83&11.76&0.00&52.63&48.67&23\\

    ModelScope~\cite{wang2023modelscope}&44.44&55.00&40.00&63.64&20.00&60.00&25.00&66.67&60.00&25.00&40.74&80.00&54.55&44.44&72.73&40.00&54.17&29.41&0.00&57.89&47.33&24\\
    VideoCrafter1~\cite{chen2024videocrafter2}&55.56&60.00&40.00&36.36&10.00&53.33&33.33&33.33&70.00&8.33&62.96&90.00&54.55&50.00&45.45&30.00&54.17&11.76&9.09&57.89&46.00&25\\

    Tune-A-Video~\cite{wu2023tune}&72.22&45.00&30.00&54.55&0.00&56.67&58.33&55.56&80.00&16.67&25.93&60.00&45.45&55.56&54.55&30.00&41.67&23.53&9.09&36.84&43.00&26\\
    
    LVDM~\cite{he2022lvdm}&83.33&45.00&10.00&36.36&10.00&53.33&16.67&44.44&20.00&8.33&40.74&60.00&54.55&55.56&18.18&30.00&45.83&23.53&0.00&68.42&40.33&27\\
    LTX~\cite{HaCohen2024LTXVideo}&55.56&50.00&40.00&45.45&0.00&53.33&8.33&44.44&50.00&0.00&40.74&40.00&0.00&72.22&36.36&10.00&41.67&29.41&9.09&36.84&37.00&28\\
    ZeroScope~\cite{Zeroscope}&50.00&35.00&20.00&0.00&0.00&30.00&8.33&55.56&20.00&0.00&22.22&50.00&9.09&16.67&9.09&10.00&16.67&5.88&0.00&47.37&22.00&29\\
    LWM~\cite{liu2024world}&16.67&25.00&10.00&9.09&0.00&13.33&8.33&11.11&10.00&0.00&11.11&0.00&0.00&16.67&0.00&10.00&8.33&0.00&0.00&5.26&9.00 &30\\

    \Xhline{1px}
  \end{tabular}}\label{mos3}

 \vspace{-2mm}
\end{table*}
\subsection{T2V Model Performance across 20 Challenges}

Tables \ref{mos1}-\ref{mos3} provide detailed performance comparisons of the 30 T2V models across 20 task-specific challenges on three types of human annotations: perception MOS, T2V correspondence MOS, and task-specific accuracy. 
For perception quality, as shown in Table \ref{mos1}, Jimeng~\cite{jimeng} stands out with the highest MOS and performs particularly well in the category “Complex”. This is likely because complex combinations tend to be visually appealing and often involve the harmonious integration of various components, which makes them more engaging and immersive. 
For T2V correspondence, as demonstrated in Table \ref{mos2}, Wanxiang~\cite{alibaba2024tongyiwanxiang} leads the way, demonstrating strong alignment between the generated videos and the textual descriptions, but has a relatively lower performance in perception quality. In contrast, models such as Kling~\cite{kling} excel in perception quality, delivering high MOS scores, but cannot perform as well in terms of T2V correspondence.
The contrasting trends in performance between perception quality and T2V correspondence emphasize the importance of evaluating both dimensions independently. While perception quality focuses on the visual aspects of the generated videos, T2V correspondence measures how well the video aligns with the content described in the text prompt. This dual evaluation ensures a more comprehensive understanding of a model’s abilities, where one dimension evaluates aesthetic quality, and the other checks the accuracy of the video-text alignment.
Most models perform well in the “Object” category but fall short in the “OCR” category. This suggests that while the T2V models excel at generating visually coherent objects, they struggle with accurately interpreting and rendering textual elements.
In terms of task-specific accuracy, the differences between models become particularly evident. Some models achieve near-perfect accuracy, with scores reaching as high as 100\%, demonstrating their exceptional ability to handle specific tasks with precision. On the other hand, certain models perform poorly, with accuracy scores as low as 0\%, indicating significant limitations in their ability to correctly complete the designated tasks. This stark contrast highlights the varying capabilities of the models and underscores the importance of task-specific evaluation in assessing the true performance of text-to-video models across different challenges.

\section{Details of Loss Function}
\label{loss}
The training process for LOVE is divided into two progressive stages, each utilizing a specific loss function to target distinct objectives: language loss for training, aligning visual and language features to give visual question answers across the 20 task-specific challenges, L1 loss for quality regression to generate accurate perception and correspondence scores.

\paragraph{(1) Training with language loss.}
In the first stage, we train the projector to align visual and language features using the standard language loss. This involves ensuring that the visual tokens extracted from the vision encoder and temporal encoder correspond effectively to the language representations from the LLM. The language loss, calculated using a cross-entropy function, measures the model’s ability to predict the correct token given the prior context:
\begin{eqnarray}
\begin{aligned}
&\mathcal{L}_{\text{language}} = -\frac{1}{N} \sum_{i=1}^N \log P(y_{\text{label}} | y_{\text{pred}})
\end{aligned}
\end{eqnarray}
where $P(y_{\text{label}} | y_{\text{pred})}$ represents the probability assigned to the correct token $y_{\text{label}}$ by the model, $y_{\text{pred}}$ is the predicted token, and $N$ is the total number of tokens. By minimizing this loss, the model learns to generate coherent textual descriptions of video content, laying the foundation for subsequent stages.

\paragraph{(2) Quality Regression with L1 loss.}
Once the model can produce coherent descriptions of video content, the focus shifts to fine-tuning the quality regression module to output stable and precise numerical quality scores. The quality regression module takes the aligned visual tokens as input and predicts a quality score that reflects the overall video quality.
Using the AIGVE-60K, which contains human-annotated MOS for each video, the model is trained to align its predictions with human ratings. The training objective minimizes the difference between the predicted quality score $Q_{predict}$ and the ground-truth MOS  $Q_{label}$ using the L1 loss function:
\begin{eqnarray}
\begin{aligned}
\label{loss_function}
\mathcal{L}_{\text{MOS}} = \frac{1}{N} \sum_{i=1}^N \left| Q_{\text{predict}}(i) - Q_{\text{label}}(i) \right|
\end{aligned}
\end{eqnarray}
where $Q_{\text{predict}}(i)$ is the score predicted by the regressor $i$ and $Q_{\text{label}}(i)$ is the corresponding ground-truth MOS derived from subjective experiments, and $N$  is the number of videos in the batch. This loss function ensures that the predicted scores remain consistent with human evaluations, enabling the model to accurately assess the quality of AI-generated videos in numerical form.
\section{Implemention Details}
\label{implementation}

\subsection{Detailed Information of Evaluation Criteria}
\noindent
We adopt the widely used metrics in VQA literature \cite{sun2022deep, wu2022fast,wu2023dover}: Spearman rank-order correlation coefficient (SRCC), Pearson linear correlation coefficient (PLCC), and Kendall’s Rank Correlation Coefficient (KRCC) as our evaluation criteria. SRCC quantifies the extent to which the ranks of two variables are related, which ranges from -1 to 1. Given $N$ action videos, SRCC is computed as:
\begin{equation}
SRCC = 1 - \frac{{6\sum\nolimits_{n = 1}^N {{{({v_n} - {p_n})}^2}} }}{{N({N^2} - 1)}},
\end{equation}
where $v_n$ and $p_n$ denote the rank of the ground truth $y_n$ and the rank of predicted score ${\hat y_n}$ respectively. The higher the SRCC, the higher the monotonic correlation between ground truth and predicted score.
Similarly, PLCC measures the linear correlation between predicted scores and ground truth scores, which can be formulated as:
\begin{equation}
PLCC = \frac{{\sum\nolimits_{n = 1}^N {({y_n} - \bar y)({{\hat y}_n} - \bar {\hat y})} }}{{\sqrt {\sum\nolimits_{n = 1}^N {{{({y_n} - \bar y)}^2}} } \sqrt {\sum\nolimits_{n = 1}^N {{{({{\hat y}_n} - \bar {\hat y})}^2}} } }},
\end{equation}
where $\bar y$ and $\bar {\hat y}$ are the mean of ground truth and predicted score respectively.

\noindent
We also adopt the Kendall Rank Correlation Coefficient (KRCC) as an evaluation metric, which measures the ordinal association between two variables. For a pair of ranks $(v_i, p_i)$ and $(v_j, p_j)$, the pair is concordant if:
\begin{equation}
(v_i - v_j)(p_i - p_j) > 0,
\end{equation}
and discordant if $<  $ 0.
 Given $N$ AIGVs, KRCC is computed as:
\begin{equation}
KRCC =  \frac{{C - D}}{{\frac{1}{2}N(N-1)}},
\end{equation}
where $C$ and $D$ denote the number of concordant and discordant pairs, respectively.

\subsection{Detailed Information of Evaluation Methods}

\noindent
\textbf{BMPRI} \cite{quality:BMPRI}, \textbf{BPRI}  \cite{min2017blind}, \textbf{BRISQUE} \cite{mittal2012no}, \textbf{HOSA} \cite{xu2016blind}, \textbf{NIQE} \cite{mittal2012making}, \textbf{QAC} \cite{xue2013learning} are conventional handcrafted IQA methods. These methods primarily rely on the extraction of natural image features, which are designed to capture perceptual characteristics such as texture, structure, and color information. For our video input, we employ a strategy of uniformly sampling 8 frames from each video. This ensures that we capture key frames of the videos, while maintaining computational efficiency. The scores obtained for each frame are then averaged to produce a final score for the video. 

\noindent
\textbf{V-Aesthetic Quality}~\cite{huang2024vbench}, \textbf{V-Imaging Quality}~\cite{huang2024vbench}, \textbf{V-Overall Consistency}~\cite{huang2024vbench}, \textbf{V-Subject Consistency}~\cite{huang2024vbench},\textbf{V-Temporal Flickering}~\cite{huang2024vbench} are metrics derived from VBench~\cite{huang2024vbench}, which incorporate a wide range of smaller, specialized sub-metrics designed to evaluate various aspects of video quality. 
\textbf{V-Aesthetic} Quality evaluates the visual appeal of the generated video, considering factors such as layout, color richness, and the artistic quality of the subjects. Using the LAION~\cite{schuhmann2022laion} aesthetic predictor, each video frame is rated on a scale from 0 to 10, normalized to a 0-1 range. The final aesthetic score for the video is obtained by averaging the scores of all frames, with higher scores indicating better aesthetic quality. 
\textbf{V-Imaging Quality} measures low-level distortions like noise and blur, using the MUSIQ~\cite{MUSIQ} predictor trained on the SPAQ~\cite{fang2020perceptual} dataset to compute frame-wise quality scores.
\textbf{V-Overall Consistency} measures the overall consistency between the video and the corresponding text prompt, focusing on both semantic and style alignment. This is assessed using ViCLIP~\cite{wang2023internvid}, a model that evaluates how well the text and generated video align in terms of meaning and stylistic elements.
\textbf{V-Subject Consistency} is measured using DINO~\cite{liu2023grounding} features, with the subject consistency score calculated based on cosine similarity between frames. 
\textbf{V-Temporal Flickering} evaluates temporal consistency by focusing on flickering caused by lighting or shaky camera motions, using static video scenes to isolate this issue and calculating flickering on a frame-by-frame basis.

\noindent
{\bf VSFA}~\cite{VSFA} is an objective no-reference video quality assessment method by integrating two eminent effects of the human visual system, namely, content-dependency and temporal-memory effects into a deep neural network. 
\begin{table*}
\centering
\vspace{-9mm}
  \caption{An overview and URLs of the adopted 30 V2T interpretation models.}
  \label{models22}
  
    \renewcommand\arraystretch{1}
  \resizebox{0.8\linewidth}{!}{\begin{tabular}{ll}
            \toprule 
        \textbf{Methods} & \textbf{URL} \\
        \hline
$\clubsuit$VSFA \cite{VSFA} & \url{https://github.com/lidq92/VSFA} \\
$\clubsuit$BVQA \cite{li2022blindly} & \url{https://github.com/vztu/BVQA_Benchmark} \\
$\clubsuit$SimpleVQA \cite{sun2022deep} & \url{https://github.com/Raykshj/SimpleVQA} \\
$\clubsuit$FAST-VQA \cite{wu2022fast} & \url{https://github.com/VQAssessment/FAST-VQA-and-FasterVQA} \\
$\clubsuit$DOVER \cite{wu2023dover} & \url{https://github.com/VQAssessment/DOVER} \\
\hline

$\heartsuit$CLIPScore~\cite{hessel2021clipscore}& \url{https://github.com/jmhessel/clipscore} \\
$\heartsuit$BLIPScore~\cite{li2022blip} & \url{https://github.com/salesforce/BLIP} \\
$\heartsuit$AestheticScore \cite{schuhmann2022laion} & \url{https://github.com/sorekdj60/AestheticScore} \\
$\heartsuit$ImageReward \cite{xu2023imagereward} & \url{https://github.com/THUDM/ImageReward} \\
$\heartsuit$PickScore \cite{kirstain2023pick} & \url{https://github.com/yuvalkirstain/PickScore} \\
$\heartsuit$HPSv2~\cite{wu2023human} & \url{https://github.com/tgxs002/HPSv2}\\
$\heartsuit$VQAScore \cite{li2024evaluating} & \url{https://github.com/linzhiqiu/t2v_metrics} \\
$\heartsuit$FGA-BLIP2~\cite{han2024evalmuse40kreliablefinegrainedbenchmark}& \url{https://github.com/DYEvaLab/EvalMuse}\\
\hline
$\bigstar$DeepseekVL2~\cite{wu2024deepseekvl2mixtureofexpertsvisionlanguagemodels}& \url{https://github.com/deepseek-ai/DeepSeek-V2}\\
$\bigstar$VideoLlava~\cite{lin2023video}& \url{https://github.com/PKU-YuanGroup/Video-LLaVA}\\
$\bigstar$VideoLlama3~\cite{damonlpsg2025videollama3}& \url{https://github.com/DAMO-NLP-SG/VideoLLaMA3}\\
$\bigstar$mPLUG-OWL3 \cite{ye2024mplug} & \url{https://github.com/X-PLUG/mPLUG-Owl} \\ 
$\bigstar$Qwen-VL~\cite{Qwen2.5-VL}& \url{https://github.com/QwenLM/Qwen2.5-VL}\\
$\bigstar$Llama3.2-Vision~\cite{meta2024llama}& \url{https://huggingface.co/meta-llama/Llama-3.2-11B-Vision}\\
$\bigstar$CogAgent~\cite{hong2024cogagentvisuallanguagemodel}& \url{https://github.com/THUDM/CogAgent}\\
$\bigstar$LLaVA-NEXT \cite{liu2024llavanext} & \url{https://github.com/LLaVA-VL/LLaVA-NeXT} \\
$\bigstar$InternVideo2.5 \cite{wang2025internvideo} & \url{https://github.com/OpenGVLab/InternVideo} \\
$\bigstar$InternVL~\cite{chen2024expanding} & \url{https://github.com/OpenGVLab/InternVL} \\

\hline
$\triangle$Gemini1.5-pro \cite{Gemini}&\url{https://gemini.google.com} \\
$\triangle$Claude3.5 \cite{Claude3.5}&\url{https://claude.ai} \\
$\triangle$Grok2 Vision \cite{Grok2}&\url{https://grok.com} \\
$\triangle$ChatGPT-4o \cite{GPT4}&\url{https://chatgpt.com} \\
\bottomrule
    \end{tabular}}
    \vspace{-8mm}
\end{table*}

\noindent
{\bf BVQA}~\cite{li2022blindly} leverages the transferred knowledge from IQA databases with authentic distortions and large-scale action recognition with rich motion patterns for better video representation. 

\noindent
{\bf SimpleVQA}~\cite{sun2022a} adopts an end-to-end spatial feature extraction network to directly learn the quality-aware spatial feature representation from raw pixels of the video frames and extract the motion features to measure the temporal-related distortions. A pre-trained SlowFast model is used to extract motion features. 

\noindent
{\bf FAST-VQA}~\cite{wu2022fast} proposes a grid mini-patch sampling strategy, which allows consideration of local quality by sampling patches at their raw resolution and covers global quality with contextual relations via mini-patches sampled in uniform grids. It overcomes the high computational costs when evaluating high-resolution videos. 

\noindent
{\bf DOVER}~\cite{wu2023dover} is a disentangled objective video quality evaluator that learns the quality of videos based on technical and aesthetic perspectives.

\noindent
{\bf CLIPScore}~\cite{hessel2021clipscore} is an image captioning metric  and passes both the image and the candidate caption through their respective feature extractors, then computing the cosine similarity between the text and video embeddings.

\noindent
{\bf BLIPScore}~\cite{li2022blip} provides more advanced multi-modal feature extraction capabilities. Using the same methodology as CLIPScore ~\cite{hessel2021clipscore}, it computes the cosine similarity between the text and visual embeddings, but benefits from enhanced pre-training strategy, which is designed to better capture fine-grained relationships between text and visual content.

\noindent
{\bf ImageReward}~\cite{xu2023imagereward} builds upon the BLIP model~\cite{li2022blip} by introducing an additional MLP layer on top of BLIP’s output. Instead of directly computing a similarity score, the MLP generates a scalar value representing the preference for one video over another in comparative settings. 

\noindent
{\bf AestheticScore}~\cite{schuhmann2022laion} is given by an aesthetic predictor introduced by LAION \cite{schuhmann2022laion}

\noindent
{\bf PickScore}~\cite{kirstain2023pick}
is trained by fine-tuning CLIP-H on human preference data, aiming to maximize the probability of a preferred video being selected. PickScore exhibits strong correlation with human rankings, outperforming traditional metrics like FID and aesthetics predictors.

\noindent
{\bf HPSv2}~\cite{wu2023human}
 is designed for better aligning their outputs with human preferences. HPS is based on a fine-tuned CLIP model that accurately predicts human preferences over generated contents. 
 
\noindent
{\bf VQAScore}~\cite{li2024evaluating}
 is designed to assess the alignment between generated videos and text prompts, particularly for compositional text-to-visual generation tasks. It can be used in a black-box manner, requiring no fine-tuning or additional prompt decomposition. 
 
\noindent
\textbf{FGA-BLIP2}~\cite{han2024evalmuse40kreliablefinegrainedbenchmark}
 utilizes vision-language models to jointly fine-tune image-text alignment scores and element-level annotations. This approach enables the model to generate overall scores while determining whether the generated images match the elements specified in the prompt. 
 
\noindent
{\bf DeepseekVL2}~\cite{wu2024deepseekvl2mixtureofexpertsvisionlanguagemodels}
is an advanced series of mix-of-experts (MoE) vision language models. It introduces a dynamic tiling vision encoding strategy, allowing efficient processing of high-resolution videos with varying aspect ratios, enhancing tasks like visual grounding and document analysis. It also leverages the Multi-head Latent Attention (MLA) mechanism for the language component, which reduces computational costs and improves inference efficiency. 

\noindent
\textbf{VideoLlava}~\cite{liu2024improved} Video-LLaVA binds visual signals to the language feature space, unifying visual representations, and proposes a solution to align before
projection. It enables LLM to perform visual reasoning capabilities on both images and
videos simultaneously.

\noindent
\textbf{VideoLlama3}~\cite{yao2024minicpm}
is a vision-language model trained through a four-stage paradigm. Key innovations include using Rotary Position Embedding (RoPE) for dynamic image resolution and compressing video tokens for more efficient representation, enabling the model to achieve good performance in both image and video understanding benchmarks.

\noindent
{\bf LLaVA-NeXT}~\cite{liu2024llavanext} improves on LLaVA-1.5~\cite{liu2024improved} by increasing input video resolution and enhances visual detail, reasoning, and OCR capabilities. It also improves world knowledge and logical reasoning while maintaining LLaVA's minimalist design and data efficiency, using under 1M visual instruction tuning samples. 

\noindent
{\bf mPLUG-Owl3}~\cite{ye2024mplug} is a versatile multi-modal large language model designed to handle long video sequences, interleaved video-text, and lengthy video inputs. It introduces Hyper Attention blocks that efficiently integrate vision and language into a shared semantic space, allowing for the processing of extended multi-video scenarios. 

\noindent
{\bf Qwen2-VL}~\cite{Qwen2-VL}
is an advanced large vision-language model designed to process videos, videos, and text with dynamic resolution handling and multimodal rotary position embedding (M-RoPE). The model features strong capabilities in OCR, video comprehension, multilingual support, and robust agent functionalities for device operations. 

\noindent
{\bf Qwen2.5-VL}~\cite{Qwen2.5-VL}
 is the latest flagship model in the Qwen vision-language series, featuring significant improvements in visual recognition, object localization, document parsing, and long-video comprehension. Building on the Qwen2-VL architecture, it introduces key enhancements such as dynamic resolution processing for videos and videos, absolute time encoding for temporal dynamics, and window attention to optimize inference efficiency. 
 
\noindent
{\bf Llama3.2-Vision}~\cite{meta2024llama}
 excels in video reasoning tasks, such as document-level understanding, chart and graph captioning, and visual grounding. These models can reason with videos, such as answering questions based on graphs or maps, and generate captions that describe visual scenes. 
 
\noindent
\textbf{Llava-one-vision}~\cite{xiong2024llavaovchat} 
is an open-source large multimodal model (LMM) designed to enhance vision-and-language tasks in single-image, multi-image, and video scenarios. It utilizes a cost-efficient architecture connecting vision encoders with LLMs, demonstrating strong video understanding through task transfer from images.

\noindent
{\bf CogAgent}~\cite{hong2024cogagentvisuallanguagemodel}
is designed to facilitate understanding and navigation of graphical user interfaces (GUIs). It utilizes both low and high-resolution video encoders to recognize small text and page elements. CogAgent excels in GUI tasks like navigation and decision-making. CogAgent’s innovative design includes a cross-attention branch to balance high-resolution inputs and computational efficiency. 

\noindent
{\bf InternVL2.5}~\cite{chen2024expanding}
 demonstrates strong performance in various benchmarks, including multi-discipline reasoning, document and video understanding, and multimodal hallucination detection. The model features enhanced vision encoders, larger dataset sizes, and improved test-time scaling.
 
\noindent
{\bf InternVL3}~\cite{chen2024expanding}
is an advanced multimodal large language model series that surpasses its predecessor, InternVL 2.5, in multimodal perception and reasoning. It extends its capabilities to tool usage, GUI agents, industrial image analysis, 3D vision perception, and more. Built upon Native Multimodal Pre-Training, InternVL3 integrates language and multimodal training in a single stage, enhancing its performance without requiring additional bridging modules. 

\noindent
{\bf InternVideo2.5}~\cite{wang2025internvideo} 
is an advanced multimodal large language model designed to enhance video understanding by focusing on long and rich context modeling. This approach improves the model's ability to perceive fine-grained details and capture long-term temporal structures in videos. By incorporating dense vision task annotations and utilizing direct preference optimization (DPO), it creates compact spatiotemporal representations through adaptive hierarchical token compression.

\subsection{Question design for LLM-based models}
For LMM-based methods, we not only need to input the video to be evaluated, but also the corresponding prompt and instructions to guide the model to output the result we want. Three different questions need to be input for each video to be evaluated. When designing questions from the two dimensions of Perception and T2V correspondence, all videos have a unified template, but to obtain the question-answer pair for each video, different questions need to be designed according to the challenge corresponding to the prompt used to generate the video. We have a total of 20 tasks, so there are 20 question templates for the task-specific challenges. The specific question templates are listed as follows:
\begin{itemize}
    \item \textbf{Perception Quality}:
    Suppose you are now a volunteer for subjective quality evaluation of videos, and you are now required to rate the perception quality of the given videos on a scale of 0-100. Results are accurate to the nearest digit. Answer only one score.
    \item \textbf{T2V Correspondence}:
Please rate the consistency between the video and the text description “\textless prompt \textgreater”. The rating scale is from 0 to 100, with higher scores for descriptions that include important content from the video and lower scores for descriptions that lack important content. Results are accurate to the nearest digit. Answer only a score.
    \item \textbf{Task-specific Questions}:

    \begin{itemize}
    \item[(1)] \textbf{Object}: Does the video contain \textless class\_name \textgreater? Answer yes or no.
    \item[(2)] \textbf{Color}: Does the video contain \textless class\_name \textgreater in the color of \textless class\_color \textgreater? Answer yes or no.
    \item[(3)] \textbf{Counting}: Does the video contain \textless class\_count \textgreater \textless class\_name \textgreater? Answer yes or no.
    \item[(4)] \textbf{Texture}: Does the video contain a \textless class\_texture \textgreater \textless class\_name \textgreater? Answer yes or no.
    \item[(5)] \textbf{Position}: Does the video contain both \textless class1\_name \textgreater and \textless class2\_name \textgreater, and are they positioned as described in “\textless prompt \textgreater”? Answer yes or no.
    \item[(6)] \textbf{HOI}: Does the video contain both a person and \textless object\_name \textgreater, and is the person's action \textless verb\_ing \textgreater? Answer yes or no.
    \item[(7)] \textbf{Face}: Does the face in the video have \textless first\_feature \textgreater  \textless second\_feature \textgreater and \textless third\_feature \textgreater? Answer yes or no.
     \item[(8)] \textbf{Emotion}: If there is a person in the video, is their emotion \textless emotion\_class \textgreater? If there is no person, does the overall mood of the video convey \textless emotion\_class \textgreater? Answer yes or no.
     \item[(9)] \textbf{Human}: Do the appearance, hairstyle, accessories, and profession of the person in the video match the description in “\textless prompt \textgreater”? Answer yes or no.
     \item[(10)] \textbf{OCR}: Does the video contain the text “\textless OCR \textgreater” with all letters correct? Answer yes or no.
     \item[(11)] \textbf{Scene}: Does the video depict a \textless scene\_name \textgreater scene? Answer yes or no.
     \item[(12)] \textbf{Style}: Is the style of the video \textless style\_name \textgreater? Answer yes or no.
    \item[(13)] \textbf{Shapes}: Does the video contain a \textless class\_shape \textgreater \textless class\_name \textgreater? Answer yes or no.
    \item[(14)] \textbf{View}: Is the perspective shown in the video \textless view\_class \textgreater? Answer yes or no.
    \item[(15)] \textbf{World Knowledge}: Does the video contain a famous landmark or celebrity \textless knowledge\_class \textgreater? Answer yes or no.
    \item[(16)] \textbf{Linguistic Structure}: Does the scene depicted in the video exclude \textless class\_name \textgreater? Answer yes or no.
    \item[(17)] \textbf{Imagination}: Does the video content show imaginative elements, and does it match the description in “\textless prompt \textgreater”? Answer yes or no.
    \item[(18)] \textbf{Motion Direction}: Does the video show the trajectory and path of \textless class\_name \textgreater as described in as described in “\textless prompt \textgreater”? Answer yes or no.
    \item[(19)] \textbf{Event Order}: Does the video show the sequence of actions of \textless class\_name \textgreater as described in “\textless prompt \textgreater”? Answer yes or no.
    \item[(20)] \textbf{Complex}: The questions for a complex challenge are a combination of the questions for the individual challenges described above. For example, for a complex challenge consisting of a combination of task 1, task 2, etc., the question template is: Are the text descriptions of the pictures: \textless task1\_question \textgreater, \textless task1\_question \textgreater \dots all correct? Answer yes or no.
    \end{itemize}
\end{itemize}
The content in “\textless  \textgreater” in the above question template needs to be determined based on the specific prompt content.
\section{More Results Comparisons}
\label{comparison}
\begin{table*}[t]
\vspace{-7mm}
\centering
\renewcommand\arraystretch{1.2}
\caption{Performance comparisons of V2T Interpretation Models on \textbf{perception} SRCC.
}
   \resizebox{1\linewidth}{!}{\begin{tabular}{l||cccccccccccccccccccc>{\columncolor{mycolor_green}}c:>{\columncolor{mycolor_blue}}c}
  \Xhline{1px}
 Models
&$\text{Object}$&$\text{Color}$&$\text{Count.}$&$\text{Texture}$&$\text{Position}$ &$\text{HOI}$&$\text{Face}$&$\text{Emo.}$&$\text{Human}$&$\text{OCR}$&$\text{Scene}$&$\text{Style}$&$\text{Shape}$&$\text{View}$&$\text{Know.}$&$\text{Ling.}$&$\text{Imag.}$&$\text{Motion}$&$\text{Event}$&$\text{Complex}$& $\text{Overall}$ &Rank\\
    \hline
    $\spadesuit$BMPRI \cite{quality:BMPRI}&0.500&0.532&0.571&0.527&0.669&0.592&0.581&0.653&0.631&0.470&0.610&0.565&0.586&0.573&0.669&0.566&0.583&0.581&0.573&0.598 &0.574 &9\\
    $\spadesuit$BPRI  \cite{min2017blind}&0.326&0.279&0.318&0.373&0.362&0.387&0.462&0.498&0.386&0.223&0.426&0.196&0.248&0.368&0.427&0.301&0.397&0.364&0.358&0.429&0.356&33\\
    $\spadesuit$BRISQUE \cite{mittal2012no}&0.515&0.573&0.564&0.563&0.697&0.603&0.609&0.605&0.659&0.623&0.580&0.603&0.642&0.507&0.603&0.654&0.566&0.606&0.626&0.549 &0.584 &8\\
    $\spadesuit$HOSA \cite{xu2016blind}&0.594&0.610&0.689&0.612&0.753&0.668&0.690&0.681&0.698&0.622&0.636&0.674&0.676&0.620&0.683&0.684&0.634&0.650&0.674&0.642 &0.647 &3\\
    $\spadesuit$NIQE \cite{mittal2012making}&0.603&0.625&0.681&0.627&0.773&0.671&0.692&0.675&0.687&0.621&0.646&0.661&0.701&0.634&0.695&0.692&0.635&0.654&0.701&0.642 &0.654 &2\\
    $\spadesuit$QAC \cite{xue2013learning}&0.611&0.573&0.637&0.599&0.711&0.638&0.56&0.644&0.585&0.499&0.612&0.567&0.579&0.591&0.671&0.653&0.581&0.583&0.635&0.601 &0.596 &7\\
    \hdashline
    
    $\diamondsuit$V-Aesthetic Quality~\cite{huang2024vbench}&0.479&0.379&0.614&0.651&0.525&0.525&0.504&0.420&0.545&0.563&0.503&0.573&0.573&0.462&0.534&0.510&0.522&0.509&0.547&0.450&0.503&20\\
    $\diamondsuit$V-Imaging Quality~\cite{huang2024vbench}&0.180&0.251&0.358&0.217&0.286&0.359&0.333&0.271&0.370&0.241&0.291&0.349&0.313&0.176&0.392&0.273&0.273&0.253&0.333&0.309&0.281&37\\
    $\diamondsuit$V-Overall Consistency~\cite{huang2024vbench}&0.092&0.171&0.135&0.246&0.225&0.122&0.016&0.054&0.113&0.328&0.122&0.036&0.117&0.170&0.064&0.130&0.159&0.079&0.049&0.203&0.156&44\\
    $\diamondsuit$V-Subject Consistency~\cite{huang2024vbench}&0.305&0.365&0.444&0.366&0.432&0.457&0.450&0.316&0.404&0.242&0.362&0.431&0.407&0.332&0.310&0.365&0.399&0.245&0.338&0.414&0.344 &35\\
    $\diamondsuit$V-Temporal Flickering~\cite{huang2024vbench}&0.397&0.368&0.431&0.448&0.512&0.488&0.488&0.533&0.364&0.240&0.448&0.428&0.494&0.462&0.430&0.398&0.438&0.371&0.470&0.496&0.408&28\\
    \hdashline
    
    $\clubsuit$VSFA \cite{VSFA}&0.329&0.383&0.502&0.325&0.512&0.394&0.224&0.238&0.327&0.451&0.447&0.393&0.356&0.309&0.504&0.375&0.365&0.420&0.424&0.456&0.375&31\\
    $\clubsuit$BVQA \cite{li2022blindly}&0.375&0.394&0.263&0.180&0.191&0.388&0.092&0.363&0.388&0.182&0.313&0.168&0.179&0.384&0.289&0.124&0.326&0.294&0.319&0.275&0.309&36\\
    $\clubsuit$SimpleVQA \cite{sun2022deep}&0.536&0.512&0.650&0.616&0.686&0.596&0.622&0.572&0.536&0.543&0.568&0.653&0.629&0.524&0.592&0.538&0.569&0.479&0.624&0.602 &0.563 &10\\
    $\clubsuit$FAST-VQA \cite{wu2022fast}&0.624&0.610&0.677&0.658&0.756&0.676&0.726&0.639&0.592&0.588&0.632&0.661&0.699&0.625&0.638&0.63&0.646&0.650&0.665&0.640&0.639 &5\\
    $\clubsuit$DOVER \cite{wu2023dover}&0.616&0.663&0.699&0.727&0.760&0.652&0.779&0.686&0.592&0.649&0.659&0.661&0.776&0.636&0.597&0.565&0.629&0.610&0.662&0.683&0.641 &4\\
    \hdashline
    $\heartsuit$CLIPScore~\cite{hessel2021clipscore}&0.008&0.106&0.032&0.118&0.155&0.034&0.089&0.024&0.129&0.350&0.051&0.054&0.103&0.008&0.112&0.107&0.239&0.047&0.038&0.093&0.095&46\\
    $\heartsuit$BLIPScore~\cite{li2022blip}&0.172&0.199&0.133&0.254&0.342&0.229&0.076&0.057&0.192&0.384&0.115&0.086&0.225&0.221&0.181&0.193&0.183&0.011&0.003&0.208&0.188&40\\
    $\heartsuit$AestheticScore~\cite{schuhmann2022laion}&0.509&0.490&0.650&0.666&0.657&0.572&0.602&0.594&0.627&0.558&0.527&0.639&0.581&0.489&0.556&0.510&0.589&0.584&0.592&0.548&0.552&12\\

    $\heartsuit$ImageReward~\cite{xu2023imagereward}&0.341&0.414&0.419&0.526&0.606&0.417&0.340&0.382&0.418&0.600&0.442&0.449&0.452&0.448&0.300&0.279&0.400&0.364&0.167&0.407&0.418&27\\
    $\heartsuit$PickScore~\cite{kirstain2023pick}&0.383&0.312&0.547&0.540&0.661&0.518&0.577&0.441&0.296&0.572&0.370&0.543&0.560&0.476&0.291&0.339&0.448&0.306&0.456&0.346&0.403&29\\
    $\heartsuit$HPSv2~\cite{wu2023human}&0.463&0.482&0.642&0.618&0.718&0.573&0.663&0.515&0.591&0.620&0.551&0.615&0.695&0.549&0.454&0.441&0.513&0.489&0.471&0.562 &0.542 &14\\
    $\heartsuit$VQAScore~\cite{li2024evaluating}&0.109&0.171&0.153&0.073&0.094&0.146&0.032&0.091&0.162&0.357&0.273&0.068&0.039&0.187&0.113&0.140&0.218&0.196&0.067&0.266&0.168&42\\
    $\heartsuit$FGA-BLIP2~\cite{han2024evalmuse40kreliablefinegrainedbenchmark}&0.482&0.523&0.533&0.587&0.716&0.566&0.598&0.483&0.474&0.610&0.501&0.586&0.568&0.556&0.388&0.436&0.493&0.521&0.345&0.508&0.518 &19\\

    \hdashline

$\bigstar$DeepseekVL2 (1B)~\cite{wu2024deepseekvl2mixtureofexpertsvisionlanguagemodels}&0.048&0.003&0.018&0.032&0.042&0.004&0.077&0.006&0.034&0.063&0.001&0.052&0.025&0.108&0.005&0.015&0.017&0.096&0.039&0.028&0.012&48\\
$\bigstar$VideoLlava (7B)~\cite{lin2023video}&0.193&0.108&0.185&0.411&0.040&0.175&0.146&0.118&0.129&0.14&0.152&0.418&0.293&0.233&0.180&0.316&0.199&0.081&0.151&0.192&0.181&41\\
$\bigstar$mPLUG-Owl3 (7B)~\cite{ye2024mplug}&0.316&0.392&0.333&0.318&0.292&0.453&0.313&0.349&0.365&0.334&0.353&0.353&0.233&0.357&0.329&0.283&0.280&0.345&0.366&0.453&0.353&34\\
$\bigstar$VideoLlama3 (8B)~\cite{damonlpsg2025videollama3} &0.365&0.390&0.498&0.346&0.406&0.424&0.335&0.179&0.507&0.287&0.466&0.480&0.381&0.438&0.280&0.418&0.441&0.406&0.310&0.435&0.392 &30\\

$\bigstar$Qwen2-VL (7B) \cite{wang2024qwen2}&0.315&0.344&0.440&0.370&0.376&0.383&0.386&0.362&0.525&0.370&0.395&0.405&0.442&0.279&0.135&0.296&0.341&0.364&0.382&0.341&0.357&32\\

$\bigstar$Qwen2.5-VL (7B)~\cite{Qwen2.5-VL}&0.506&0.500&0.616&0.653&0.592&0.560&0.630&0.537&0.543&0.433&0.489&0.546&0.608&0.487&0.494&0.514&0.571&0.540&0.558&0.558&0.541 &15\\

$\bigstar$Llama3.2-Vision (11B)~\cite{meta2024llama}&0.027&0.094&0.107&0.113&0.267&0.067&0.056&0.116&0.072&0.120&0.144&0.205&0.215&0.086&0.056&0.120&0.077&0.064&0.107&0.076&0.094&47\\
$\bigstar$CogAgent (18B)~\cite{hong2024cogagentvisuallanguagemodel}&0.150&0.167&0.117&0.059&0.106&0.107&0.231&0.096&0.089&0.123&0.094&0.259&0.158&0.101&0.127&0.200&0.096&0.093&0.167&0.114&0.124 &45\\
$\bigstar$InternVL2.5 (8B)~\cite{chen2024expanding}&0.309&0.258&0.366&0.330&0.307&0.299&0.187&0.175&0.339&0.179&0.310&0.423&0.215&0.255&0.228&0.320&0.288&0.304&0.319&0.261&0.280&38\\
$\bigstar$InternVL3 (9B)~\cite{wang2024mpo}&0.229&0.257&0.298&0.371&0.310&0.293&0.348&0.289&0.266&0.205&0.274&0.300&0.313&0.251&0.259&0.244&0.322&0.218&0.246&0.281&0.273&39\\
$\bigstar$InternVideo2.5 (8B)~\cite{wang2025internvideo}&0.799&0.741&0.832&0.828&0.871&0.833&0.853&0.789&0.791&0.818&0.810&0.855&0.850&0.748&0.796&0.837&0.787&0.829&0.823&0.761&0.156&43\\
$\bigstar$LLaVA-NeXT (8B)~\cite{liu2024llavanext}&0.463&0.421&0.613&0.551&0.567&0.532&0.442&0.474&0.501&0.525&0.495&0.663&0.549&0.432&0.481&0.487&0.471&0.501&0.523&0.461&0.489&23\\

$\bigstar$InternVL2.5 (38B)~\cite{chen2024expanding}&0.567&0.576&0.673&0.717&0.750&0.666&0.659&0.609&0.661&0.582&0.620&0.741&0.661&0.572&0.579&0.660&0.690&0.632&0.584&0.678&0.623 &6\\

$\bigstar$InternVL3 (38B)~\cite{wang2024mpo}&0.460&0.463&0.535&0.582&0.551&0.554&0.504&0.438&0.530&0.421&0.526&0.574&0.453&0.465&0.374&0.441&0.579&0.511&0.497&0.534 &0.495 &22\\

$\bigstar$Qwen2-VL (72B) \cite{wang2024qwen2}&0.449&0.398&0.53&0.498&0.412&0.522&0.521&0.516&0.549&0.341&0.484&0.501&0.452&0.464&0.464&0.474&0.439&0.466&0.497&0.558 &0.463 &26\\
$\bigstar$Qwen2.5-VL (72B)~\cite{Qwen2.5-VL}&0.399&0.404&0.506&0.464&0.478&0.438&0.560&0.432&0.424&0.375&0.400&0.502&0.409&0.378&0.407&0.409&0.482&0.377&0.472&0.465 &0.425 &25\\
$\bigstar$Llava-one-vision (72B)~\cite{xiong2024llavaovchat}&0.517&0.485&0.623&0.636&0.634&0.550&0.573&0.473&0.547&0.376&0.552&0.636&0.520&0.479&0.487&0.558&0.579&0.476&0.564&0.517 &0.529 &17\\
$\bigstar$InternVL2.5 (72B)~\cite{chen2024expanding}&0.472&0.466&0.621&0.627&0.735&0.550&0.433&0.398&0.498&0.591&0.542&0.581&0.571&0.474&0.401&0.662&0.642&0.500&0.500&0.596&0.538 &16\\
$\bigstar$InternVL3 (72B)~\cite{wang2024mpo}&0.509&0.485&0.588&0.610&0.609&0.567&0.586&0.503&0.546&0.586&0.532&0.655&0.584&0.479&0.448&0.571&0.643&0.503&0.518&0.575 &0.544 &13\\
\hdashline
$\triangle$Gemini1.5-pro \cite{Gemini}&0.494&0.418&0.612&0.532&0.585&0.622&0.566&0.430&0.502&0.391&0.497&0.514&0.461&0.485&0.506&0.490&0.515&0.489&0.508&0.516&0.497&21\\
$\triangle$Claude3.5 \cite{Claude3.5}&0.362&0.401&0.516&0.517&0.488&0.462&0.402&0.316&0.496&0.451&0.438&0.473&0.439&0.413&0.317&0.290&0.492&0.414&0.280&0.490&0.427 &24\\
$\triangle$Grok2 Vision \cite{Grok2}&0.488&0.501&0.653&0.649&0.552&0.643&0.629&0.518&0.591&0.506&0.579&0.604&0.579&0.567&0.535&0.557&0.599&0.535&0.591&0.632&0.563 &11\\
$\triangle$ChatGPT-4o \cite{GPT4}&0.403&0.455&0.681&0.652&0.682&0.564&0.581&0.523&0.577&0.706&0.473&0.545&0.569&0.419&0.360&0.532&0.507&0.474&0.489&0.548&0.526 &18\\
\hdashline
$\blacklozenge$LOVE (Ours)&0.786&0.745&0.837&0.870&0.847&0.822&0.877&0.804&0.790&0.856&0.796&0.834&0.873&0.763&0.776&0.861&0.759&0.814&0.837&0.792&0.793&1\\

    \Xhline{1px}
  \end{tabular}}
\label{srcc1}
\end{table*}
\begin{table*}[t]
\vspace{-2mm}
\centering
\renewcommand\arraystretch{1.1}
\caption{Performance comparisons of V2T Interpretation Models on \textbf{correspondence} SRCC.
}
   \resizebox{\linewidth}{!}{\begin{tabular}{l||cccccccccccccccccccc>{\columncolor{mycolor_green}}c:>{\columncolor{mycolor_blue}}c}
  \Xhline{1px}
 Models
&$\text{Object}$&$\text{Color}$&$\text{Count.}$&$\text{Texture}$&$\text{Position}$ &$\text{HOI}$&$\text{Face}$&$\text{Emo.}$&$\text{Human}$&$\text{OCR}$&$\text{Scene}$&$\text{Style}$&$\text{Shape}$&$\text{View}$&$\text{Know.}$&$\text{Ling.}$&$\text{Imag.}$&$\text{Motion}$&$\text{Event}$&$\text{Complex}$& $\text{Overall}$ &Rank\\
    \hline
    $\spadesuit$BMPRI \cite{quality:BMPRI}&0.301&0.319&0.290&0.327&0.515&0.371&0.351&0.547&0.421&0.254&0.451&0.248&0.297&0.423&0.496&0.342&0.44&0.368&0.374&0.382&0.362&32\\
    $\spadesuit$BPRI  \cite{min2017blind}&0.183&0.122&0.178&0.258&0.275&0.227&0.257&0.387&0.239&0.093&0.248&0.007&0.119&0.264&0.290&0.113&0.278&0.266&0.201&0.264&0.202&40\\
    $\spadesuit$BRISQUE \cite{mittal2012no}&0.360&0.361&0.310&0.389&0.559&0.395&0.364&0.509&0.424&0.372&0.447&0.285&0.379&0.39&0.448&0.406&0.441&0.373&0.394&0.327&0.381&30\\
    $\spadesuit$HOSA \cite{xu2016blind}&0.410&0.371&0.372&0.421&0.592&0.442&0.445&0.567&0.474&0.339&0.493&0.290&0.364&0.451&0.492&0.429&0.478&0.409&0.437&0.411&0.415&24\\
    $\spadesuit$NIQE \cite{mittal2012making}&0.417&0.393&0.389&0.435&0.613&0.457&0.453&0.575&0.469&0.350&0.525&0.309&0.401&0.473&0.520&0.449&0.496&0.428&0.469&0.425&0.435&22\\
    $\spadesuit$QAC \cite{xue2013learning}&0.449&0.388&0.319&0.413&0.571&0.428&0.378&0.574&0.433&0.259&0.487&0.297&0.282&0.411&0.516&0.454&0.452&0.412&0.422&0.423&0.395&27\\
    
    \hdashline
    $\diamondsuit$V-Aesthetic Quality~\cite{huang2024vbench}&0.403&0.312&0.318&0.604&0.402&0.402&0.486&0.379&0.421&0.420&0.414&0.285&0.422&0.405&0.477&0.425&0.445&0.316&0.276&0.333&0.403&26\\
    $\diamondsuit$V-Imaging Quality~\cite{huang2024vbench}&0.159&0.185&0.241&0.194&0.243&0.179&0.259&0.202&0.189&0.195&0.218&0.186&0.288&0.154&0.331&0.217&0.285&0.121&0.146&0.208&0.195&43\\
    $\diamondsuit$V-Overall Consistency~\cite{huang2024vbench}&0.172&0.236&0.439&0.436&0.370&0.288&0.141&0.166&0.271&0.608&0.329&0.250&0.315&0.304&0.225&0.033&0.403&0.257&0.188&0.384&0.308&35\\
    $\diamondsuit$V-Subject Consistency~\cite{huang2024vbench}&0.161&0.191&0.196&0.266&0.313&0.251&0.313&0.101&0.244&0.146&0.138&0.181&0.252&0.163&0.322&0.135&0.265&0.049&0.096&0.244&0.165&45\\
    $\diamondsuit$V-Temporal Flickering~\cite{huang2024vbench}&0.211&0.231&0.159&0.276&0.428&0.218&0.349&0.371&0.179&0.045&0.255&0.152&0.248&0.269&0.362&0.145&0.277&0.127&0.266&0.302&0.196&42\\

    \hdashline
    $\clubsuit$VSFA \cite{VSFA}&0.256&0.264&0.268&0.269&0.428&0.226&0.204&0.246&0.167&0.329&0.331&0.118&0.229&0.203&0.392&0.275&0.301&0.233&0.233&0.252&0.244&37\\
    $\clubsuit$BVQA \cite{li2022blindly}&0.211&0.275&0.198&0.133&0.145&0.289&0.108&0.340&0.319&0.047&0.310&0.086&0.090&0.334&0.190&0.077&0.268&0.262&0.256&0.219&0.238&38\\
    $\clubsuit$SimpleVQA \cite{sun2022deep}&0.349&0.316&0.314&0.487&0.503&0.349&0.496&0.427&0.335&0.277&0.402&0.281&0.453&0.423&0.422&0.400&0.428&0.194&0.328&0.445&0.347&33\\
    $\clubsuit$FAST-VQA \cite{wu2022fast}&0.370&0.358&0.342&0.451&0.585&0.390&0.523&0.537&0.375&0.283&0.448&0.252&0.463&0.471&0.497&0.383&0.482&0.334&0.370&0.409&0.392&29\\
    $\clubsuit$DOVER \cite{wu2023dover}&0.355&0.413&0.307&0.550&0.557&0.345&0.591&0.503&0.306&0.355&0.429&0.226&0.498&0.486&0.483&0.379&0.417&0.258&0.321&0.478&0.376&31\\

    \hdashline
    $\heartsuit$CLIPScore~\cite{hessel2021clipscore}&0.130&0.240&0.292&0.299&0.273&0.112&0.302&0.074&0.175&0.574&0.229&0.204&0.272&0.158&0.160&0.048&0.392&0.128&0.137&0.233&0.229&39\\
    $\heartsuit$BLIPScore~\cite{li2022blip}&0.193&0.319&0.428&0.454&0.449&0.398&0.245&0.160&0.375&0.622&0.317&0.307&0.407&0.321&0.352&0.083&0.314&0.177&0.178&0.362&0.316&34\\
    
    $\heartsuit$AestheticScore~\cite{schuhmann2022laion}&0.32&0.295&0.330&0.539&0.523&0.440&0.449&0.499&0.442&0.369&0.393&0.322&0.436&0.401&0.448&0.292&0.394&0.333&0.294&0.335&0.393&28\\

    $\heartsuit$ImageReward~\cite{xu2023imagereward}&0.441&0.553&0.596&0.632&0.690&0.625&0.460&0.402&0.518&0.715&0.520&0.501&0.598&0.495&0.489&0.231&0.487&0.365&0.272&0.526&0.508&16\\
    $\heartsuit$PickScore~\cite{kirstain2023pick}&0.529&0.413&0.517&0.563&0.616&0.479&0.555&0.213&0.359&0.676&0.424&0.515&0.593&0.551&0.407&0.247&0.432&0.259&0.472&0.444&0.414&25\\
    $\heartsuit$HPSv2~\cite{wu2023human}&0.470&0.524&0.548&0.643&0.657&0.582&0.572&0.393&0.504&0.652&0.517&0.450&0.654&0.582&0.532&0.300&0.489&0.369&0.500&0.550&0.499&17\\

    $\heartsuit$VQAScore~\cite{li2024evaluating}&0.150&0.222&0.172&0.063&0.202&0.131&0.058&0.073&0.218&0.513&0.291&0.031&0.192&0.187&0.134&0.086&0.202&0.255&0.165&0.249&0.176&44\\

    $\heartsuit$FGA-BLIP2~\cite{han2024evalmuse40kreliablefinegrainedbenchmark}&0.665&0.667&0.718&0.770&0.771&0.720&0.614&0.516&0.612&0.750&0.508&0.551&0.697&0.602&0.508&0.302&0.593&0.356&0.400&0.600&0.596&10\\

\hdashline

$\bigstar$DeepseekVL2 (1B)~\cite{wu2024deepseekvl2mixtureofexpertsvisionlanguagemodels}&0.027&0.025&0.004&0.003&0.015&0.029&0.075&0.112&0.098&0.087&0.055&0.004&0.078&0.045&0.119&0.046&0.001&0.041&0.076&0.059&0.017&48\\
$\bigstar$VideoLlava (7B)~\cite{lin2023video}&0.198&0.381&0.163&0.310&0.383&0.338&0.360&0.244&0.200&0.057&0.159&0.404&0.243&0.228&0.279&0.169&0.200&0.098&0.046&0.123&0.201&41\\
$\bigstar$mPLUG-Owl3 (7B)~\cite{ye2024mplug}&0.503&0.535&0.640&0.666&0.720&0.590&0.539&0.478&0.483&0.675&0.407&0.531&0.577&0.549&0.393&0.795&0.456&0.341&0.251&0.476&0.548&14\\
$\bigstar$VideoLlama3 (8B)~\cite{damonlpsg2025videollama3}&0.381&0.488&0.466&0.394&0.573&0.509&0.389&0.330&0.397&0.400 &0.291&0.448&0.463&0.327&0.254&0.755&0.398&0.253&0.296&0.349&0.423&23\\
$\bigstar$Qwen2-VL (7B) \cite{wang2024qwen2}&0.478&0.505&0.492&0.426&0.467&0.516&0.484&0.360&0.435&0.449&0.414&0.592&0.382&0.382&0.468&0.666&0.441&0.303&0.314&0.411&0.450&21\\
$\bigstar$Qwen2.5-VL (7B)~\cite{Qwen2.5-VL}&0.497&0.538&0.455&0.717&0.721&0.492&0.475&0.232&0.442&0.640&0.407&0.495&0.624&0.412&0.402&0.639&0.476&0.378&0.279&0.472&0.511&15\\
$\bigstar$Llama3.2-Vision (11B)~\cite{meta2024llama}&0.055&0.181&0.110&0.164&0.020&0.028&0.003&0.090&0.009&0.117&0.038&0.156&0.154&0.037&0.214&0.007&0.046&0.054&0.033&0.098&0.080&47\\
$\bigstar$CogAgent (18B)~\cite{hong2024cogagentvisuallanguagemodel}&0.13&0.185&0.165&0.169&0.127&0.108&0.157&0.079&0.122&0.125&0.080&0.074&0.174&0.130&0.075&0.074&0.116&0.004&0.174&0.070&0.119&46\\
$\bigstar$InternVL2.5 (8B)~\cite{chen2024expanding}&0.448&0.543&0.507&0.590&0.628&0.452&0.276&0.329&0.459&0.561&0.398&0.469&0.481&0.340&0.388&0.603&0.449&0.357&0.346&0.403&0.486&19\\
$\bigstar$InternVL3 (9B)~\cite{wang2024mpo}&0.467&0.441&0.536&0.566&0.632&0.482&0.359&0.448&0.454&0.552&0.380&0.395&0.458&0.391&0.381&0.642&0.440&0.362&0.382&0.390&0.478&20\\
$\bigstar$InternVideo2.5 (8B)~\cite{wang2025internvideo}&0.708&0.713&0.736&0.748&0.856&0.692&0.744&0.569&0.648&0.864&0.578&0.647&0.550&0.644&0.499&0.697&0.608&0.421&0.544&0.534&0.498&18\\
$\bigstar$LLaVA-NeXT (8B)~\cite{liu2024llavanext}&0.357&0.380&0.313&0.353&0.423&0.357&0.376&0.136&0.325&0.120&0.352&0.260&0.255&0.272&0.148&0.160&0.296&0.211&0.119&0.235&0.285&36\\
$\bigstar$InternVL2.5 (38B)~\cite{chen2024expanding}&0.685&0.633&0.679&0.740&0.816&0.628&0.510&0.546&0.621&0.649&0.556&0.687&0.616&0.500&0.449&0.798&0.646&0.439&0.525&0.637&0.647&5\\
$\bigstar$InternVL3 (38B)~\cite{wang2024mpo}&0.612&0.624&0.629&0.606&0.760&0.618&0.536&0.448&0.548&0.700&0.537&0.610&0.453&0.503&0.487&0.757&0.591&0.484&0.499&0.526&0.600&9\\
$\bigstar$Qwen2-VL (72B) \cite{wang2024qwen2}&0.513&0.600&0.611&0.642&0.725&0.547&0.460&0.444&0.474&0.608&0.449&0.554&0.572&0.487&0.502&0.716&0.543&0.414&0.353&0.511&0.560&13\\
$\bigstar$Qwen2.5-VL (72B)~\cite{Qwen2.5-VL}&0.582&0.675&0.626&0.683&0.799&0.595&0.618&0.414&0.532&0.693&0.512&0.652&0.613&0.517&0.538&0.827&0.594&0.471&0.401&0.577&0.627&7\\
$\bigstar$Llava-one-vision (72B)~\cite{xiong2024llavaovchat}&0.556&0.564&0.660&0.631&0.679&0.573&0.442&0.367&0.590&0.594&0.498&0.514&0.598&0.387&0.480&0.772&0.566&0.398&0.344&0.493&0.570&12\\
$\bigstar$InternVL2.5 (72B)~\cite{chen2024expanding}&0.702&0.682&0.703&0.728&0.780&0.654&0.513&0.465&0.580&0.723&0.585&0.611&0.665&0.560&0.502&0.808&0.663&0.536&0.499&0.627&0.661&4\\
$\bigstar$InternVL3 (72B)~\cite{wang2024mpo}&0.635&0.628&0.570&0.690&0.760&0.655&0.586&0.444&0.621&0.702&0.583&0.578&0.610&0.541&0.485&0.825&0.606&0.525&0.493&0.592&0.631&6\\

\hdashline
$\triangle$Gemini1.5-pro \cite{Gemini}&0.611&0.630&0.668&0.680&0.763&0.588&0.566&0.532&0.534&0.696&0.422&0.651&0.644&0.579&0.535&0.785&0.596&0.393&0.346&0.523&0.610&8\\
$\triangle$Claude3.5 \cite{Claude3.5}&0.593&0.618&0.576&0.705&0.689&0.607&0.486&0.442&0.519&0.649&0.530&0.621&0.631&0.531&0.474&0.733&0.536&0.448&0.317&0.557&0.583&11\\
$\triangle$Grok2 Vision \cite{Grok2}&0.641&0.697&0.737&0.739&0.793&0.655&0.633&0.577&0.652&0.755&0.571&0.630&0.581&0.562&0.498&0.835&0.613&0.474&0.451&0.589&0.666&2\\
$\triangle$ChatGPT-4o \cite{GPT4}&0.622&0.655&0.757&0.748&0.813&0.653&0.659&0.551&0.648&0.736&0.504&0.584&0.700&0.570&0.550&0.782&0.615&0.571&0.379&0.608&0.664&3\\
\hdashline
$\blacklozenge$LOVE (Ours)&0.747&0.754&0.808&0.845&0.862&0.749&0.773&0.651&0.671&0.806&0.629&0.772&0.813&0.643&0.527&0.850&0.672&0.553&0.574&0.663 &0.747 &1\\

    \Xhline{1px}
  \end{tabular}}
\label{srcc2}
\end{table*}
\begin{table*}[t]
\vspace{-4mm}
\centering
\renewcommand\arraystretch{1.1}
\caption{Performance comparisons of V2T Interpretation Models on \textbf{task-specific accuracy} (\%).
}
   \resizebox{\linewidth}{!}{\begin{tabular}{l||cccccccccccccccccccc>{\columncolor{mycolor_green}}c:>{\columncolor{mycolor_blue}}c}
  \Xhline{1px}
 Models
&$\text{Object}$&$\text{Color}$&$\text{Count.}$&$\text{Texture}$&$\text{Position}$ &$\text{HOI}$&$\text{Face}$&$\text{Emo.}$&$\text{Human}$&$\text{OCR}$&$\text{Scene}$&$\text{Style}$&$\text{Shape}$&$\text{View}$&$\text{Know.}$&$\text{Ling.}$&$\text{Imag.}$&$\text{Motion}$&$\text{Event}$&$\text{Complex}$&$\text{Overall}$ &Rank\\
    \hline
 
$\spadesuit$BMPRI \cite{quality:BMPRI}&27.78&25.00&50.00&36.36&80.00&16.67&41.67&33.33&30.00&91.67&29.63&60.00&63.64&11.11&27.27&80.00&12.50&47.06&81.82&15.79&64.00&30\\
$\spadesuit$BPRI  \cite{min2017blind}&27.78&25.00&50.00&36.36&80.00&16.67&41.67&33.33&30.00&91.67&29.63&60.00&63.64&11.11&27.27&80.00&12.50&47.06&81.82&15.79&63.56&33\\
$\spadesuit$BRISQUE \cite{mittal2012no}&27.78&25.00&50.00&36.36&80.00&16.67&41.67&33.33&30.00&91.67&29.63&60.00&63.64&11.11&27.27&80.00&12.50&47.06&81.82&15.79&64.67&26\\
$\spadesuit$HOSA \cite{xu2016blind}&27.78&25.00&50.00&36.36&80.00&16.67&41.67&33.33&30.00&91.67&29.63&60.00&63.64&11.11&27.27&80.00&12.50&47.06&81.82&15.79&64.34&29\\
$\spadesuit$NIQE \cite{mittal2012making}&27.78&25.00&50.00&36.36&80.00&16.67&41.67&33.33&30.00&91.67&29.63&60.00&63.64&11.11&27.27&80.00&12.50&47.06&81.82&15.79&62.21&39\\
$\spadesuit$QAC \cite{xue2013learning}&27.78&25.00&50.00&36.36&80.00&16.67&41.67&33.33&30.00&91.67&29.63&60.00&63.64&11.11&27.27&80.00&12.50&47.06&81.82&15.79&64.40&28\\
\hdashline

$\diamondsuit$V-Aesthetic Quality~\cite{huang2024vbench}&72.22&65.00&60.00&54.55&20.00&70.00&50.00&44.44&90.00&100.0&62.96&40.00&63.64&55.56&72.73&50.00&83.33&41.18&54.55&68.42&64.54&27\\
$\diamondsuit$V-Imaging Quality~\cite{huang2024vbench}&44.44&45.00&60.00&54.55&20.00&43.33&33.33&55.56&60.00&75.00&48.15&50.00&63.64&55.56&63.64&40.00&41.67&58.82&81.82&47.37&60.60&43\\
$\diamondsuit$V-Overall Consistency~\cite{huang2024vbench}&55.56&60.00&70.00&54.55&80.00&50.00&41.67&88.89&50.00&91.67&77.78&40.00&63.64&66.67&54.55&50.00&75.00&70.59&72.73&57.89&61.96&41\\
$\diamondsuit$V-Subject Consistency~\cite{huang2024vbench}&61.11&60.00&50.00&63.64&20.00&73.33&66.67&55.56&60.00&50.00&70.37&50.00&36.36&77.78&72.73&20.00&87.50&64.71&27.27&73.68&62.52&36\\
$\diamondsuit$V-Temporal Flickering~\cite{huang2024vbench}&72.22&75.00&50.00&63.64&20.00&83.33&58.33&66.67&70.00&8.33&70.37&40.00&36.36&88.89&72.73&20.00&87.50&52.94&18.18&84.21&63.69&32\\
\hdashline

$\clubsuit$VSFA \cite{VSFA}&38.89&45.00&50.00&63.64&40.00&40.00&41.67&77.78&40.00&83.33&33.33&70.00&54.55&22.22&90.91&40.00&33.33&41.18&72.73&36.84&57.09&46\\
$\clubsuit$BVQA \cite{li2022blindly}&33.33&25.00&30.00&45.45&60.00&16.67&50.00&33.33&30.00&83.33&37.04&80.00&45.45&11.11&72.73&40.00&25.00&35.29&81.82&15.79&58.47&44\\
$\clubsuit$SimpleVQA \cite{sun2022deep}&66.67&50.00&30.00&54.55&20.00&43.33&83.33&55.56&50.00&83.33&40.74&40.00&63.64&50.00&72.73&60.00&62.50&47.06&45.45&52.63&60.78&42\\
$\clubsuit$FAST-VQA \cite{wu2022fast}&77.78&65.00&60.00&72.73&20.00&66.67&83.33&55.56&80.00&66.67&62.96&40.00&27.27&77.78&72.73&40.00&75.00&41.18&36.36&63.16&66.37&22\\
$\clubsuit$DOVER \cite{wu2023dover}&72.22&55.00&40.00&72.73&20.00&50.00&83.33&44.44&60.00&83.33&37.04&30.00&45.45&66.67&72.73&60.00&58.33&41.18&45.45&63.16&62.61&35\\
\hdashline

$\heartsuit$CLIPScore~\cite{hessel2021clipscore}&44.44&45.00&50.00&45.45&60.00&46.67&58.33&66.67&70.00&83.33&62.96&40.00&63.64&55.56&54.55&60.00&79.17&52.94&63.64&57.89&58.27&45\\
$\heartsuit$BLIPScore~\cite{li2022blip}&61.11&60.00&70.00&45.45&40.00&76.67&41.67&55.56&70.00&66.67&74.07&60.00&72.73&83.33&72.73&30.00&75.00&47.06&63.64&68.42&63.93&31\\
$\heartsuit$AestheticScore~\cite{schuhmann2022laion}&72.22&60.00&30.00&54.55&30.00&73.33&58.33&55.56&80.00&91.67&62.96&60.00&27.27&66.67&63.64&60.00&87.50&41.18&18.18&63.16&64.87&25\\
$\heartsuit$ImageReward~\cite{xu2023imagereward}&77.78&50.00&70.00&54.55&50.00&66.67&58.33&55.56&80.00&83.33&66.67&50.00&36.36&55.56&72.73&60.00&75.00&58.82&54.55&52.63&68.33&18\\
$\heartsuit$PickScore~\cite{kirstain2023pick}&72.22&55.00&60.00&72.73&30.00&63.33&66.67&55.56&40.00&91.67&51.85&50.00&54.55&55.56&45.45&60.00&58.33&29.41&81.82&52.63&62.29&38\\
$\heartsuit$HPSv2~\cite{wu2023human}&61.11&45.00&70.00&63.64&30.00&43.33&50.00&55.56&80.00&83.33&44.44&60.00&36.36&50.00&63.64&60.00&37.50&41.18&72.73&57.89&67.68&19\\
$\heartsuit$VQAScore~\cite{li2024evaluating}&22.22&35.00&70.00&45.45&80.00&23.33&58.33&66.67&30.00&83.33&48.15&30.00&54.55&16.67&27.27&70.00&58.33&47.06&72.73&42.11&52.97&47\\
$\heartsuit$FGA-BLIP2~\cite{han2024evalmuse40kreliablefinegrainedbenchmark}&55.56&50.00&60.00&90.91&50.00&43.33&83.33&33.33&60.00&91.67&55.56&80.00&54.55&61.11&72.73&60.00&45.83&47.06&72.73&52.63&67.06&20\\

\hdashline
$\bigstar$DeepseekVL2 (1B)~\cite{wu2024deepseekvl2mixtureofexpertsvisionlanguagemodels}&33.33&25.00&50.00&36.36&80.00&20.00&41.67&33.33&40.00&75.00&22.22&50.00&63.64&11.11&27.27&80.00&20.83&47.06&72.73&36.84&39.29&48\\
$\bigstar$VideoLlava (7B)~\cite{lin2023video}&72.22&70.00&60.00&63.64&70.00&86.67&58.33&66.67&90.00&75.00&66.67&50.00&36.36&88.89&72.73&50.00&91.67&52.94&63.64&73.68&68.46&17\\
$\bigstar$mPLUG-Owl3 (7B)~\cite{ye2024mplug}&33.33&18.18&66.67&60.00&100.0&14.29&66.67&50.00&33.33&90.91&50.00&100.0&25.00&28.57&50.00&87.50&25.00&60.00&81.82&12.50&63.02&34\\
$\bigstar$VideoLlama3 (8B)~\cite{damonlpsg2025videollama3}&83.33&70.00&40.00&72.73&70.00&63.33&58.33&66.67&80.00&100.0&74.07&40.00&54.55&83.33&54.55&100.0&79.17&47.06&72.73&78.95&70.16&16\\
$\bigstar$Qwen2-VL (7B) \cite{wang2024qwen2}&77.78&65.00&50.00&90.91&80.00&66.67&58.33&44.44&80.00&58.33&74.07&60.00&54.55&83.33&63.64&100.0&83.33&47.06&54.55&78.95&71.56&13\\
$\bigstar$Qwen2.5-VL (7B)~\cite{Qwen2.5-VL}&72.22&65.00&60.00&81.82&80.00&36.67&58.33&66.67&70.00&91.67&55.56&70.00&90.91&55.56&45.45&100.0&37.50&47.06&81.82&36.84&62.34&37\\
$\bigstar$Llama3.2-Vision (11B)~\cite{meta2024llama}&72.22&55.00&80.00&63.64&90.00&63.33&75.00&44.44&70.00&91.67&70.37&40.00&63.64&61.11&36.36&60.00&54.17&41.18&90.91&42.11&62.19&40\\
$\bigstar$CogAgent (18B)~\cite{hong2024cogagentvisuallanguagemodel}&77.78&70.00&50.00&72.73&50.00&80.00&66.67&66.67&80.00&41.67&62.96&30.00&63.64&66.67&63.64&50.00&83.33&52.94&18.18&78.95&65.32&24\\
$\bigstar$InternVL2.5 (8B)~\cite{chen2024expanding}&83.33&65.00&60.00&63.64&70.00&80.00&66.67&66.67&70.00&91.67&74.07&50.00&27.27&72.22&36.36&80.00&79.17&52.94&90.91&68.42&66.30 &21\\
$\bigstar$InternVL3 (9B)~\cite{wang2024mpo}&83.33&60.00&40.00&90.91&50.00&70.00&66.67&66.67&90.00&33.33&81.48&30.00&72.73&72.22&45.45&80.00&79.17&47.06&27.27&84.21&65.82&23\\
$\bigstar$InternVideo2.5 (8B)~\cite{wang2025internvideo}&83.33&75.00&50.00&63.64&70.00&63.33&66.67&55.56&100.0&91.67&77.78&60.00&45.45&77.78&72.73&90.00&83.33&41.18&72.73&73.68&70.64&14\\
$\bigstar$LLaVA-NeXT (8B)~\cite{liu2024llavanext}&72.22&75.00&60.00&54.55&80.00&73.33&66.67&66.67&70.00&91.67&62.96&60.00&63.64&77.78&72.73&90.00&75.00&47.06&72.73&73.68&70.21&15\\
$\bigstar$InternVL2.5 
(38B)~\cite{chen2024expanding}&83.33&65.00&40.00&72.73&70.00&76.67&66.67&66.67&100.0&83.33&74.07&60.00&72.73&83.33&63.64&80.00&79.17&64.71&72.73&73.68&75.81&3\\
$\bigstar$InternVL3 (38B)~\cite{wang2024mpo}&83.33&60.00&50.00&72.73&70.00&80.00&66.67&66.67&90.00&91.67&74.07&50.00&54.55&83.33&54.55&80.00&83.33&58.82&63.64&78.95&73.89&7\\
$\bigstar$Qwen2-VL (72B) \cite{wang2024qwen2}&77.78&65.00&60.00&72.73&70.00&70.00&58.33&55.56&90.00&58.33&74.07&60.00&63.64&83.33&63.64&100.0&87.50&76.47&72.73&68.42&73.12&12\\
$\bigstar$Qwen2.5-VL (72B)~\cite{Qwen2.5-VL}&77.78&60.00&50.00&100.0&60.00&70.00&58.33&66.67&90.00&91.67&74.07&50.00&63.64&77.78&72.73&100.0&75.00&58.82&81.82&52.63&73.83&8\\

$\bigstar$Llava-one-vision (72B)~\cite{xiong2024llavaovchat}&77.78&65.00&40.00&72.73&60.00&76.67&58.33&55.56&60.00&100.0&74.07&60.00&54.55&77.78&63.64&70.00&83.33&41.18&63.64&73.68&73.31&10\\
$\bigstar$InternVL2.5 (72B)~\cite{chen2024expanding}&83.33&65.00&40.00&72.73&80.00&80.00&66.67&66.67&90.00&91.67&74.07&60.00&72.73&83.33&63.64&90.00&87.50&64.71&72.73&78.95&75.18&4\\
$\bigstar$InternVL3 (72B)~\cite{wang2024mpo}&83.33&65.00&40.00&72.73&70.00&80.00&58.33&66.67&100.0&83.33&74.07&60.00&72.73&83.33&63.64&80.00&87.50&52.94&54.55&78.95&74.59&6\\
\hdashline
$\triangle$Gemini1.5-pro \cite{Gemini}&77.78&70.00&60.00&72.73&80.00&60.00&66.67&66.67&80.00&91.67&66.67&60.00&45.45&83.33&63.64&90.00&62.50&64.71&81.82&57.89&73.38&9\\
$\triangle$Claude3.5 \cite{Claude3.5}&77.78&65.00&60.00&72.73&70.00&83.33&66.67&66.67&80.00&91.67&66.67&60.00&63.64&72.22&63.64&80.00&75.00&64.71&81.82&57.89&73.20&11\\
$\triangle$Grok2 Vision \cite{Grok2}&77.78&70.00&60.00&90.91&80.00&83.33&66.67&55.56&90.00&91.67&77.78&60.00&54.55&77.78&63.64&90.00&83.33&47.06&72.73&47.37&76.51&2\\
$\triangle$ChatGPT-4o \cite{GPT4}&77.78&70.00&70.00&81.82&90.00&73.33&58.33&55.56&80.00&91.67&74.07&50.00&63.64&61.11&63.64&80.00&75.00&64.71&81.82&63.16&74.84&5\\
\hdashline
$\blacklozenge$LOVE (Ours)&77.78&60.00&50.00&72.73&60.00&70.00&75.00&66.67&90.00&91.67&74.07&40.00&45.45&88.89&63.64&100.0&83.33&70.59&81.82&73.68&78.69&1\\

    \Xhline{1px}
  \end{tabular}}\label{srcc3}

\end{table*}
We analyze the performance of various V2T Interpretation models across 20 tasks, focusing on perception SRCC, correspondence SRCC, and task-specific accuracy, as shown in Tables \ref{srcc1}-\ref{srcc3}. Traditional handcrafted models, such as BMPRI~\cite{quality:BMPRI} and BPRI~\cite{min2017blind}, as well as deep learning-based models, such as FAST-VQA~\cite{wu2022fast} and DOVER~\cite {wu2023dover}, rank higher on perception than correspondence. This is because these models primarily extract quality features related to image or video quality. Given that AIGV perception includes factors like clarity and aesthetic, traditional image or user-generated content (UGC) video quality features are partially applicable in AIGV as well, so they perform better on the perception. However, these models perform poorly in correspondence because they lack textual input and cannot incorporate text-based information.

In contrast, Vision-Language pretraining methods rank higher in correspondence than in perception. This is because these models focus on understanding and aligning both visual and textual inputs, enabling them to better handle multimodal relationships. LMM-based models with large parameters tend to perform better overall, with enhanced generalization capabilities compared to models with fewer parameters. Some open-source LMMs such as InternVL2.5 (72B)~\cite{chen2024expanding} and Qwen2.5-VL (72B)~\cite{Qwen2.5-VL}, perform comparably to their closed-source LMMs such as Gemini1.5-pro \cite{Gemini} and Claude3.5 \cite{Claude3.5}, indicating that parameter size and pretraining techniques play significant roles in achieving higher performance across both perception and correspondence tasks.

From Table \ref{srcc3}, it is evident that models generally perform well on tasks such as OCR, Object, and Linguistic Structure, where visual features like object recognition and text extraction play a central role. These tasks primarily involve identifying concrete visual elements or interpreting textual content, which aligns closely with the strengths of current V2T interpretation.
In contrast, models tend to perform less effectively on tasks such as Counting and Style, which demand more fine-grained reasoning and subjective interpretation. Counting requires precise object identification and accurate quantification, which can be particularly challenging in scenarios involving visually similar objects, overlapping elements, or occlusions. Style, on the other hand, is inherently subjective, involving the assessment of artistic or aesthetic quality factors that can vary significantly based on visual context and individual preferences.
These performance disparities highlight current limitations in V2T interpretation, especially in handling abstract tasks that go beyond straightforward visual recognition.


\end{document}